\documentclass[10pt,twocolumn,letterpaper]{article}
\ifdefined\pdfsuppressptexinfo
  \pdftrailerid{}
\fi

\newif\ifshowauthors
\newif\ifshowdraftaids
\newif\ifshowappendix
\newif\ifshowaihighlight
\showauthorstrue
\showdraftaidsfalse
\showappendixtrue
\showaihighlightfalse

\ifshowauthors
  \usepackage[pagenumbers]{cvpr} 
\else
  \usepackage[review]{cvpr}      
\fi
\makeatletter
\@ifpackageloaded{lineno}{%
  \ifdefined\IfFormatAtLeastTF
    \IfFormatAtLeastTF{2025-06-01}{%
      \RemoveFromHook{build/column/before}[lineno]%
      \AtBeginDocument{%
        \def\@LN@makecol{%
          \@LN@orig@makecol
          \setbox\@outputbox\vbox{%
            \boxmaxdepth\@maxdepth
            \protected@write\@auxout{}{\string\def\string\@LN@column{\if@firstcolumn1\else2\fi}}%
            \box\@outputbox}}}%
    }{}%
  \fi
}{}
\makeatother

\usepackage[normalem]{ulem}

\newcommand{\TODO}[1]{\textbf{\color{red}[TODO: #1]}}

\usepackage[T1]{fontenc}   
\usepackage{xspace}
\usepackage{amsmath}
\usepackage{amssymb}
\usepackage{booktabs}
\usepackage{multirow}
\usepackage{makecell}      
\usepackage{graphicx}
\usepackage{xcolor}
\usepackage{microtype}
\usepackage{soul}          
\usepackage{fontawesome5}  

\definecolor{cvprblue}{rgb}{0.21,0.49,0.74}
\usepackage[pagebackref,breaklinks,colorlinks,allcolors=cvprblue]{hyperref}
\makeatletter
\def\cvprsect#1{\cvprsection[#1]{\texorpdfstring{\hskip -1em.~}{}#1}}
\def\cvprsubsect#1{\cvprsubsection[#1]{\texorpdfstring{\hskip -1em.~}{}#1}}
\makeatother
\usepackage{titletoc}
\newcommand{\cdsupptoc}{%
  \startcontents[supp]%
  \begingroup\let\\\space 
  \printcontents[supp]{}{1}[2]{\section*{Contents}}%
  \endgroup\vspace{0.5em}}
\let\cdmaketitlesupp\maketitlesupplementary
\renewcommand{\maketitlesupplementary}{\cdmaketitlesupp\cdsupptoc}

\def\confName{CVPR}
\def\confYear{2027}

\newcommand{\CD}{ContinualDrive\xspace}
\newcommand{\PlanTF}{PlanTF\xspace}
\newcommand{\DiffP}{Diffusion Planner\xspace}
\newcommand{\FullFT}{Full-FT\xspace}         
\newcommand{\AdaFT}{Adapter-FT\xspace}       
\newcommand{\Replay}{Replay\xspace}          
\newcommand{\PEGP}{PEGP-style\xspace}        
\newcommand{\LoRAFT}{LoRA-FT\xspace}       
\newcommand{\LoRAReplay}{LoRA+Replay\xspace} 
\newcommand{\LoRAPEGP}{LoRA+PEGP\xspace}   

\definecolor{CDtealA}{HTML}{17A2A2}\definecolor{CDtealB}{HTML}{0D6B88}
\definecolor{CDambA}{HTML}{EF9A22}\definecolor{CDambB}{HTML}{DC5426}
\newcommand{\CDlogo}{{\textcolor[HTML]{17A2A2}{C}\textcolor[HTML]{169B9F}{o}\textcolor[HTML]{14949C}{n}\textcolor[HTML]{138D98}{t}\textcolor[HTML]{128695}{i}\textcolor[HTML]{118092}{n}\textcolor[HTML]{10798E}{u}\textcolor[HTML]{0E728B}{a}\textcolor[HTML]{0D6B88}{l}\textcolor[HTML]{EF9A22}{D}\textcolor[HTML]{EA8823}{r}\textcolor[HTML]{E67724}{i}\textcolor[HTML]{E16625}{v}\textcolor[HTML]{DC5426}{e}}}

\newcommand{\Czero}{\ensuremath{C_0}\xspace}            
\newcommand{\Tk}[1]{\ensuremath{T_{#1}}\xspace}         
\newcommand{\Qij}{\ensuremath{Q(i,j)}\xspace}           

\newcommand{\CLS}{CLS\xspace}                            
\newcommand{\SR}{SR\textsubscript{v0}\xspace}           
\newcommand{\ADEs}{ADE\textsubscript{8s}\xspace}        
\newcommand{\FDEs}{FDE\textsubscript{8s}\xspace}
\newcommand{\PDMS}{PDMS-style\xspace}                   
\newcommand{\NR}{NR\xspace}                              
\newcommand{\IDMR}{IDM-R\xspace}                         

\newcommand{\Acq}{\ensuremath{\mathrm{Acq}}\xspace}                   
\newcommand{\BWT}{\ensuremath{\mathrm{BWT}}\xspace}                   
\newcommand{\BWTold}{\ensuremath{\mathrm{BWT}_{\mathrm{old5}}}\xspace} 
\newcommand{\Dend}{\ensuremath{\Delta_{\mathrm{end}}}\xspace}         
\newcommand{\Dfin}{\ensuremath{\Delta_{\mathrm{fin}}}\xspace}         
\newcommand{\Fgt}{\ensuremath{F}\xspace}                              

\newcommand{\rob}{\ensuremath{^{\dagger}}}
\newcommand{\ci}[2]{\ensuremath{[#1,\,#2]}}

\newcommand{\gStraight}{\textit{straight-TL}\xspace}      
\newcommand{\gFollow}{\textit{follow-lead}\xspace}        
\newcommand{\gHighLat}{\textit{high-lat-acc}\xspace}      
\newcommand{\gNearMulti}{\textit{near-multi-veh}\xspace}  
\newcommand{\gWaitPed}{\textit{wait-ped}\xspace}          
\newcommand{\gPickup}{\textit{pickup-dropoff}\xspace}     
\newcommand{\gStationary}{\textit{stationary}\xspace}     
\newcommand{\gSentinel}{\textit{high-speed}\xspace}       

\newcommand*{\projecturl}{https://lzrbit.github.io/continualdrive/}
\newcommand*{\codeurl}{https://github.com/continualdrive/benchmark}

\newcommand{\paperlinks}{%
  {\normalsize
   \href{\codeurl}{\textcolor{black}{\faGithub}~\texttt{Code}}\qquad
   \href{\projecturl}{\textcolor{black}{\faGlobe}~\texttt{Website}}}}

\newcommand{\authoraffiliations}{%
  {\normalsize $^{1}$NTU\qquad $^{2}$KAIST\qquad $^{3}$UCLA}}

\sethlcolor{yellow}
\AtBeginDocument{%
  \soulregister{\cite}{7}\soulregister{\cref}{7}\soulregister{\Cref}{7}\soulregister{\ref}{7}%
  \soulregister{\eqref}{7}}

\ifshowdraftaids\else
  \renewcommand{\TODO}[1]{}
  
\fi

\renewcommand{\paragraph}[1]{\par\vspace{.5em}\noindent\textbf{#1}\hspace{0.5em}\ignorespaces}
\title{\texorpdfstring{\CDlogo}{ContinualDrive}: A Deployment-Anchored, Multi-Axis Benchmark\\for Continual Updates of Learned Driving Planners}

\author{%
  Zirui Li$^{1*}$\quad Zhiqi Mao$^{1*}$\quad Yangang Ren$^{1}$\quad Heye Huang$^{2}$\\
  Haohan Yang$^{3}$\quad Qihang Huang$^{1}$\quad Zhan Chen$^{1}$\quad Chen Lv$^{1\dagger}$\\[0.35em]
  \authoraffiliations\\[0.15em]
  {\small $^{*}$Equal contribution\qquad $^{\dagger}$Corresponding author}\\[0.45em]
  \paperlinks%
}

\begin{document}
\twocolumn[{%
  \renewcommand\twocolumn[1][]{#1}%
  \maketitle
  \begin{center}
\captionsetup{type=figure}
\IfFileExists{figures/fig1_overview.pdf}{%
  \includegraphics[width=\textwidth]{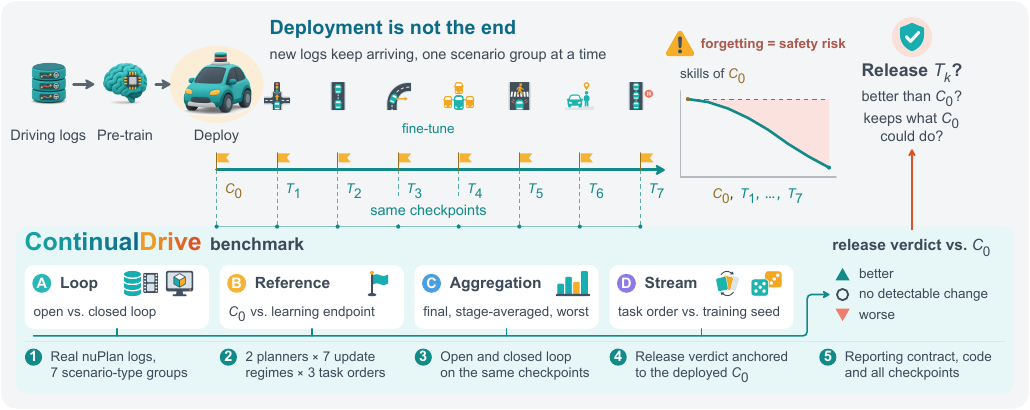}%
}{%
  \fbox{\parbox[c][2.55in][c]{0.985\textwidth}{\centering\sffamily
    Overview figure in preparation}}%
}
\captionof{figure}{\textbf{Post-deployment continual learning with \CD.}
  A deployed planner \Czero is fine-tuned on new logs group by group and can forget what it could do (schematic curve); \CD evaluates its checkpoints along four axes, anchoring the release verdict to \Czero.}
\label{fig:teaser}

  \end{center}
  \vspace{4pt}
}]
\begin{figure*}[t]
  \centering
  \includegraphics[width=\textwidth]{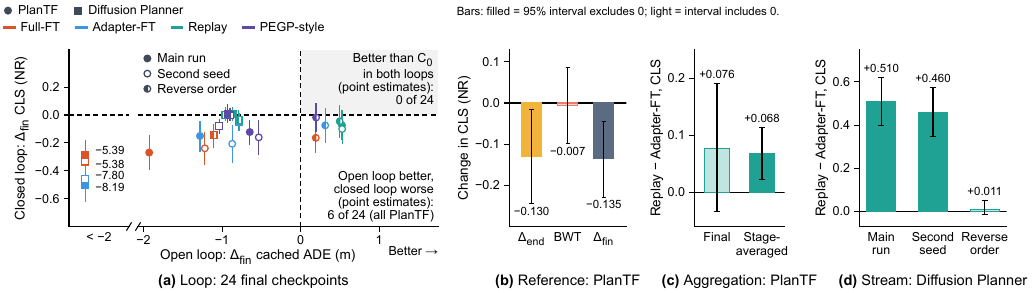}
  \caption{\textbf{Same checkpoints, different verdicts.}
  (a)~Loop: \Dfin (change from \Czero; higher is better) of the 24 core-suite final checkpoints in cached open-loop ADE (m) and non-reactive \CLS (collapsed \DiffP runs left of the axis break); none gains in both loops (shaded), and no \CLS interval lies above zero; quadrant counts are point estimates.
  (b)~Reference: pooled over 12 \PlanTF runs, the \CLS loss is present at the learning endpoint (\Dend) and persists (\Dfin), yet \BWT shows no detectable change.
  (c)~Aggregation: \PlanTF \Replay minus \AdaFT in non-reactive \CLS (main run) excludes zero only when stage-averaged.
  (d)~Stream: \DiffP \Replay leads \AdaFT in final non-reactive \CLS in both forward-order seeds but not detectably in reverse order.
  Whiskers: 95\% intervals.}
  \label{fig:verdicts}
\end{figure*}

\begin{abstract}
Learned driving planners keep being updated after deployment as new logs arrive, yet continual learning evaluations rarely ask whether an update improves on the deployed model, and their verdicts rest on unstated evaluation choices.
We present \CD, a deployment-anchored benchmark for post-deployment continual learning of driving planners.
It updates a deployed planner on real-world nuPlan logs one scenario-type group at a time, evaluates the same checkpoints in open and closed loop, makes four measurement axes explicit (evaluation loop, reference checkpoint, temporal aggregation and stream protocol), and anchors a three-valued release verdict to the deployed checkpoint.
We study two planners (\PlanTF and \DiffP) under full fine-tuning and adapter or LoRA updates, with and without replay or gradient projection, in three task orders.
Verdicts can reverse or vanish when one axis changes: open-loop ADE ranks the final \PlanTF model above the deployed one in half of its core-suite runs, whereas every closed-loop score ranks it below; backward transfer misses this loss, which on average is already present when each group has just been learned; and regime rankings vary far less across training seeds than across task orders.
In the core suite of full and adapter updates, no regime measurably improved on the deployed planner in closed loop; on a panel more than ten times larger every evaluated final checkpoint trails it, and joint training on all groups recovers much of the loss without exceeding its nuPlan closed-loop score.
We distill these findings into a reporting contract and will release the code and all checkpoints.
\end{abstract}

\section{Introduction}
\label{sec:intro}

As learned driving planners are retrained on new logs, including data collected while deployed~\cite{ren2026improving}, a release decision asks two questions: does an update beat the checkpoint in service, and which scenarios that checkpoint passed does it now fail (negative flips~\cite{yan2021pct})?
Continual learning evaluations measure forgetting against the learning endpoint~\cite{lopezpaz2017gem} and rarely ask either question.
\CD targets learned planners that map object-level inputs from real-world nuPlan logs~\cite{caesar2021nuplan,karnchanachari2024nuplan} to ego trajectories, the planning stage of modular stacks; perception and sensor-to-plan training are out of scope~\cite{chen2024e2esurvey,dauner2024navsim,zheng2025diffusionplanner}.
A post-deployment update is an offline update of a deployed checkpoint \Czero with newly logged expert data grouped by scenario type; we simulate such a stream by re-partitioning nuPlan logs, without on-vehicle learning, fleet data or temporal ordering.
\Czero is the checkpoint every update must not regress; for \PlanTF it is one we trained, not a fielded system.

Any verdict on whether an update helped, forgot or beat another regime silently fixes four measurement choices: evaluation loop, reference checkpoint, temporal aggregation and stream protocol.
Each is known to matter: open-loop error can misjudge closed-loop driving~\cite{codevilla2018offline,dauner2023parting}, task order changes continual learning results~\cite{yoon2020apd,bell2022taskorder,liu2023libero}, and metrics differ in reference and averaging~\cite{lopezpaz2017gem,chaudhry2018rwalk,diazrodriguez2018metrics}.
In driving, continual learning spans perception~\cite{verwimp2023clad,sun2022shift}, motion forecasting~\cite{ma2021continual,bao2021lifelong,schischka2026cimf,lin2023otfcl,lin2026h2c,lin2025syrem,li2025dualls} and interactive behaviour learning~\cite{lin2024dgsm,li2023cdbl}; protocols that update driving policies, such as DeLL~\cite{du2026dell} and \mbox{$\infty$-Driver}~\cite{lai2024infdriver}, cover subsets of the four choices (\cref{tab:positioning}).

\CD combines, on real-world nuPlan logs, (i)~an explicit deployed-checkpoint reference, (ii)~open- and closed-loop (non-reactive and IDM-reactive) re-evaluation of the same checkpoints on scenario-paired panels, and (iii)~full versus parameter-efficient updates (adapters and LoRA~\cite{hu2022lora}) with replay and gradient-projection variants across two planners and three task orders; to our knowledge, no prior driving continual learning protocol reports these together.
Four core regimes act as probes: \FullFT (full fine-tuning), \AdaFT (adapters~\cite{houlsby2019adapter} on a frozen backbone), \Replay (\AdaFT with random replay~\cite{chaudhry2019tiny}) and \PEGP (\AdaFT with two-sided gradient projection~\cite{qiao2024pegp}, partly re-implemented).
Each updates \PlanTF~\cite{cheng2024plantf} and \DiffP~\cite{zheng2025diffusionplanner} on seven scenario-type groups (\cref{fig:teaser}) with two forward-order seeds and in reverse order; we score every checkpoint by cached open-loop ADE and by nuPlan's closed-loop score (\CLS) and a pass rate (\SR) on a 64-scenario core panel.
These 24 runs, the core suite, carry all intervals; a third order, a LoRA branch, joint references, equal-data runs and a 720-scenario expanded panel extend and check it (\cref{sec:extensions}).

From the deployment reading (closed loop, \Czero reference, final checkpoint), changing one axis at a time flips core-suite verdicts along every axis (\cref{fig:verdicts}): \PlanTF's cached open-loop ADE beats \Czero in 6 of 12 runs while its closed-loop scores trail \Czero in all 12 (a,~loop); backward transfer (\BWT) misses a closed-loop loss already present at the learning endpoint (b,~reference); stage averages separate regimes that the final checkpoint cannot (c,~aggregation); and a \DiffP \Replay lead over \AdaFT detected in both forward-order seeds is not detectable in reverse order (d,~stream).
Each flip is an existence claim, not a frequency estimate.
Seeds moved final rankings far less than orders, partly because the position of one damaging update decides whether \DiffP \Replay keeps a large final lead (\cref{sec:order,sec:why}); open- and closed-loop changes rank \PlanTF runs similarly and both flag \DiffP's collapses; and one verdict did not flip: in the core suite, no regime measurably improved on \Czero (none of 96 final-checkpoint intervals lies above zero), which the expanded panel and the joint references support as point estimates.
Our contributions are:
\begin{itemize}
\item \textbf{C1.} A deployment-anchored protocol evaluating the same checkpoints on paired open- and closed-loop panels, with documented groups and \Czero exposure (\cref{sec:protocol}).
\item \textbf{C2.} A measurement framework: four axes, three-valued verdicts and one-axis flips, order-comparable statistics, and \Czero-anchored pass-to-fail counts~\cite{yan2021pct,li2026deitp} (\cref{sec:axes}).
\item \textbf{C3.} A 24-run core suite with paired intervals, plus a third order, a LoRA branch, joint references, equal-data runs and an expanded panel: a flip along each axis, the updates that drive them, and one verdict that did not flip (\cref{sec:results}).
\item \textbf{C4.} A contract and seven reporting rules (\cref{sec:contract}).
\end{itemize}
We propose no new method and no leaderboard, study two planners, match update budgets rather than compute, and report order sensitivity, not an effect of chronology.

\begin{table}[t]
  \providecommand{\ding}[1]{{\usefont{U}{pzd}{m}{n}\char#1}} 
  \definecolor{cmYes}{HTML}{2E8B57}\definecolor{cmNo}{HTML}{D1495B}\definecolor{cmPart}{HTML}{E59A1C}
  \providecommand{\cmY}{\textcolor{cmYes}{\ding{52}}}  
  \providecommand{\cmN}{\textcolor{cmNo}{\ding{56}}}   
  \providecommand{\cmP}{\textcolor{cmPart}{\ding{51}}} 
  \providecommand{\rh}[1]{\rotatebox{90}{#1}}
  \caption{\textbf{Comparison with continual learning benchmarks and protocols.}
    \cmY: yes; \cmP: partial; \cmN: no; --: none; blank: not determinable; codes and partial marks are explained in the notes.}
  \label{tab:positioning}
  \centering
  \scriptsize
  \setlength{\tabcolsep}{2pt}
  \begin{tabular}{@{}l ccc cccccc ccc c@{}}
    \toprule
    & \multicolumn{3}{c}{Post-deploy.} & \multicolumn{6}{c}{Multi-axis evaluation} & \multicolumn{3}{c}{Updates} & \\
    \cmidrule(lr){2-4}\cmidrule(lr){5-10}\cmidrule(lr){11-13}
    & \rh{Real logs} & \rh{\Czero ref.} & \rh{Case regr.}
    & \rh{Closed loop} & \rh{Open + closed} & \rh{Reference} & \rh{Aggregation} & \rh{$>$1 order} & \rh{$>$1 seed}
    & \rh{\# models} & \rh{FT \vs PEFT} & \rh{Replay / proj.}
    & \rh{Code + ckpt.} \\
    \midrule
    CLAD~\cite{verwimp2023clad}             & \cmY & \cmN & \cmN & \cmN & \cmN & --    & F,A & \cmN & \cmN & --  & \cmN & R    & \cmP \\
    $\infty$-Driver~\cite{lai2024infdriver} & \cmY & \cmN & \cmN & \cmP & \cmN & B,J   & F,A & \cmN & \cmN & 1   & \cmN & R    & \cmP \\
    DeLL~\cite{du2026dell}                  & \cmN & \cmN & \cmN & \cmY & \cmN & E,B   & F,A & \cmP & \cmN & 1   & \cmN & R    & \cmP \\
    LIBERO~\cite{liu2023libero}             & \cmN & \cmN & \cmN & \cmY & \cmP & E     & A   & \cmY & \cmY & 3   & \cmN & R    & \cmP \\
    TAIL~\cite{liu2024tail}                 & \cmN & \cmN & \cmN & \cmY &      & E     & A   & \cmN & \cmY & 1   & \cmY & R    & \cmN \\
    \midrule
    \CD                                     & \cmY & \cmY & \cmY & \cmY & \cmY & P,E,B,J,0 & F,A,W & \cmY & \cmP & 2 & \cmY & R,P & \cmY$^{\dagger}$ \\
    \bottomrule
  \end{tabular}
  \par\smallskip
  \parbox{\columnwidth}{\scriptsize
    \emph{Codes.} Reference: P previous checkpoint, E learning endpoint, B best checkpoint, J joint-training model, 0 start checkpoint (\Czero). Aggregation: F final, A stage-averaged, W worst stage. R / P: replay / gradient-projection regimes compared.
    \emph{Notes.}
    CLAD: perception (classification, detection); no weights.
    $\infty$-Driver: closed loop only in its supplement, on a separately trained CARLA stream; code partly released.
    DeLL: reverse order only in an appendix of its arXiv v1; code without a checkpoint link.
    LIBERO: imitation loss \vs task success in an appendix; code and data, no checkpoints.
    TAIL: appendix losses, checkpoints not stated.
    \CD: two seeds in forward order only; $^{\dagger}$code and all checkpoints by camera-ready.}
\end{table}

\section{Related Work}
\label{sec:related}

\paragraph{Learned planning and evaluation fidelity.}
nuPlan~\cite{caesar2021nuplan,karnchanachari2024nuplan} tests learned planners such as \PlanTF~\cite{cheng2024plantf} and \DiffP~\cite{zheng2025diffusionplanner} in closed loop on real-world logs; nuPlan-R replaces its Intelligent Driver Model (IDM) agents~\cite{treiber2000idm} with learned ones~\cite{peng2026nuplanr}.
Open-loop scores can misjudge driving: offline error correlates weakly with driving quality~\cite{codevilla2018offline}, nuPlan's open- and closed-loop scores are misaligned~\cite{dauner2023parting}, ego status alone matches sensor-based planners on open-loop nuScenes metrics~\cite{zhai2023rethinking,li2024egostatus}, and imitation loss can mislead about task success in robot learning~\cite{liu2023libero}.
NAVSIM~\cite{dauner2024navsim,cao2025pseudosim}, interPlan~\cite{hallgarten2024interplan} and Bench2Drive~\cite{jia2024bench2drive} add non-reactive, long-tail and multi-ability tests, and closed-loop results vary across training seeds~\cite{codevilla2019limitations}; these studies score fixed checkpoints, whereas \CD scores update streams.

\paragraph{Continual learning in driving.}
Driving continual learning covers perception~\cite{verwimp2023clad,sun2022shift,wang2022cotta} and trajectory prediction~\cite{ma2021continual,bao2021lifelong,schischka2026cimf}, where case-level forgetting tracks cases lost while averages look stable~\cite{li2026deitp}.
Policy-level work learns from takeovers, curricula or growing experts~\cite{hu2026sage,niu2024clic,cui2025mope}, studies forgetting in fine-tuned driving vision-language models~\cite{mao2026blindspot,tang2026coltdrive}, or adds memory to language-model planners~\cite{yao2025lilodriver,yao2026lunaad}.
Closest are protocols that update driving policies: $\infty$-Driver~\cite{lai2024infdriver} measures a self-trained vision policy's retention in open loop, and DeLL~\cite{du2026dell} runs a CARLA closed-loop stream with one sensor-based backbone and, in its main results, one task order, scoring retention against past-task success.
Each covers some columns of \cref{tab:positioning}; to our knowledge, no prior driving protocol reports all three elements of \cref{sec:intro}.\footnote{Contemporaneous work (public on or after September 15, 2026) post-trains driving policies in closed loop or from failures~\cite{dacol2026opted,wei2026counterplay,sun2026refinedrive,lin2026roxdrive}, adapts frozen policies~\cite{yang2026evoad,zhao2026plansnotexec}, studies retention in vision-language-action models and 3D detection~\cite{shrivastava2026sambar,khan2026cl3dod}, or proposes nuPlan planners and benchmarks~\cite{distelzweig2026interact,kim2026odyssey}.}

\paragraph{Continual learning methods and evaluation.}
Retention methods replay samples~\cite{chaudhry2019tiny,buzzega2020der}, regularize weights or outputs~\cite{kirkpatrick2017ewc,li2018lwf}, isolate parameters~\cite{mallya2018packnet}, or project gradients away from earlier tasks' subspaces~\cite{farajtabar2020ogd,saha2021gpm}, also for prompts and adapters~\cite{qiao2024pgp,qiao2024pegp}; Adam-NSCL projects the optimizer's update instead~\cite{wang2021adamnscl}.
Low-rank adapters~\cite{hu2022lora} learn and forget less than full fine-tuning in language models~\cite{biderman2024lora}; TAIL~\cite{liu2024tail} compares parameter-efficient modules with full fine-tuning for continual imitation, but with one adapter per task suite rather than one shared, deployable model.
GEM's backward transfer~\cite{lopezpaz2017gem}, RWalk's forgetting~\cite{chaudhry2018rwalk} and learning accuracy~\cite{riemer2019mer} use different references, other metrics count memory and compute~\cite{diazrodriguez2018metrics}, and results depend on task order~\cite{yoon2020apd,bell2022taskorder,lin2023theory,evron2022catastrophic}, which LIBERO and Continual World vary~\cite{liu2023libero,wolczyk2021continualworld}, and on compute budgets~\cite{prabhu2023budgeted,ghunaim2023realtime}.
Negative flips~\cite{yan2021pct} and forgetting events~\cite{toneva2019forgetting} count individual regressions, and Cao \etal require that new data never degrade a driving policy~\cite{cao2023dcarl}; \CD adds the deployed checkpoint, which backward transfer omits, as their reference.

\section{The \texorpdfstring{\CD}{ContinualDrive} Protocol}
\label{sec:protocol}

\subsection{Update stream and splits}
\label{sec:stream}

The stream visits seven scenario-type groups, named by historical nuPlan scenario-type IDs, in forward (main) order \gStraight, \gFollow, \gHighLat, \gNearMulti, \gWaitPed, \gPickup, \gStationary (\cref{fig:teaser}) or in reverse order; stage~$k$ is position~$k$, and \Tk{k} is the checkpoint after it.
Groups hold 11,282 to 100,000 training samples (413,939 from 2,142 logs; 900--7,820 updates per group) and 640 development samples each; an eighth group, \gSentinel, is evaluated but never trained in the stream (the \emph{sentinel}), and the 5,120 development samples (448 logs) share no sample or log with training.
The IDs are historical labels, not behavior categories: in five groups, under 5\% of the training samples carry the named type. For the \gPickup group, training uses only car-park scenes, whereas evaluation uses mostly pickup/drop-off scenes (supplement).
The \emph{core suite} runs each planner--regime pair in three stream settings (24 runs): the \emph{main run} (forward order, training seed 20261014), a \emph{second seed} (forward, 20261020) and the \emph{reverse order} (seed 20261014).

\subsection{Two deployed checkpoints}
\label{sec:deployment}

On the target groups, \PlanTF's deployed checkpoint scores lower and \DiffP's higher (\cref{tab:reportcard}). This comparison does not assess pretraining quality or model architecture.
\PlanTF~\cite{cheng2024plantf} starts from our 30-epoch pretraining on 300,000 samples of \gStraight, \gStationary and \gSentinel, so its sentinel measures pretraining-group retention (\gStraight and \gStationary hold 48.3\% of the stream's training samples); \DiffP~\cite{zheng2025diffusionplanner} starts from the official EMA weights.
Neither our 300,000 \PlanTF pretraining samples nor the official \DiffP training-log list shares a log with evaluation (supplement).
We train and evaluate raw weights without EMA and sample one trajectory with the official 10-step DPM solver, without guidance, with fixed seeds.

\subsection{Update regimes as probes}
\label{sec:regimes}

\FullFT updates all weights.\footnote{\PlanTF \FullFT keeps training-mode BatchNorm, whose statistics under gradient accumulation differ from those of a physical batch of 256.}
The other three regimes train width-16 bottleneck adapters~\cite{houlsby2019adapter} with zero-initialized up-projections on a frozen backbone: \AdaFT alone, \Replay with random replay~\cite{chaudhry2019tiny} of 512 stored samples per completed group (loss weight 0.25), and \PEGP, our partial re-implementation of Qiao \etal~\cite{qiao2024pegp}.
\PEGP projects each adapter's down- and up-projection gradients onto the complements of their input subspaces before AdamW, using up to 1,024 current-group samples (energy threshold 0.9, cumulative rank cap 16 per layer and side); it omits the official code's frozen-adapter warm-up, feature alignment and weight fusion.
Replay and projection start at stage~2; no regime stores \Czero data or a \Czero basis.
The cap binds early; as projection acts on gradients, not optimizer updates~\cite{wang2021adamnscl}, about 10\% of the down-projection update leaks into the protected subspace (traced \DiffP run; supplement).

\subsection{Budget and determinism}
\label{sec:budget}

Sequential runs train 20 epochs per group with AdamW (peak learning rate $10^{-4}$, per-group warm-up and cosine decay, 4-step accumulation of 64-sample micro-batches) and keep the last-epoch checkpoint without validation-based selection: 32,420 optimizer updates per full-pool run (update-matched budgets).
\Replay adds 19.0\% sample visits (24.3\% in the third order), and \PEGP adds basis collection.
The adapter regimes share one \Tk{1} per planner and stream setting; adapter training is deterministic on our hardware, \FullFT training is not, and its run-to-run spread is not quantified.

\subsection{Evaluation loops and panel}
\label{sec:evaluation}

\emph{Open loop.}
After every stage of every run, we compute ADE and FDE of the planner's single output trajectory on the 5,120 cached development samples.
\emph{Closed loop.}
The \emph{core panel} holds 64 twenty-second development scenarios (8 per group, 54 logs).
In nuPlan simulation~\cite{karnchanachari2024nuplan}, other agents replay their logs (non-reactive, \NR) or, for vehicles only, follow the IDM~\cite{treiber2000idm} (IDM-reactive, \IDMR).
We report nuPlan's closed-loop score \CLS and the success rate \SR. \CLS summarizes driving performance. \SR is the share of scenarios whose 20-s simulations complete without at-fault collisions, satisfy drivable-area and driving-direction constraints, and achieve an ego-progress ratio of at least 0.2.
Under \NR, the panel also scores the initial plan in open loop: \ADEs and \FDEs on an 8-s grid, and \PDMS, the NAVSIM v1.1 scorer~\cite{dauner2024navsim} on its first 4\,s with a non-reactive background.
Full \Czero--\Tk{7} \NR matrices exist only for the eight main runs; elsewhere, and for \IDMR, we evaluate \Czero, each group before and after its stage, and \Tk{7}.

\subsection{Extensions and controls}
\label{sec:extensions}

Five extensions reuse the deployed checkpoints, groups and development set; sequential ones use the main run's seed and are evaluated like the second-seed runs, and all are reported as point estimates.
\emph{Third order:} the core regimes in the non-mirrored order \gHighLat, \gNearMulti, \gFollow, \gStationary, \gWaitPed, \gStraight, \gPickup (8 runs).
\emph{LoRA branch:} three regimes with rank-16 LoRA~\cite{hu2022lora} on attention query and value projections of the frozen backbone: unprotected (\LoRAFT), with replay (\LoRAReplay) or with \PEGP projection of both factors (\LoRAPEGP), in all three orders (18 runs); \PlanTF \LoRAReplay in reverse order uses batch 2,048 and is not update-matched.
\emph{Joint references:} \FullFT and \AdaFT trained from each \Czero on the shuffled union of all groups for 20 epochs, with a sequential run's sample visits but one schedule (4 runs).
\emph{Equal-data runs:} the core regimes in the forward order with 10,000 samples per group (8 runs; 800 updates per group).
\emph{Expanded panel:} the core panel plus 82 random development scenarios per group (720 scenarios, 630 targets), evaluated at \Czero and \Tk{7} of the eight main runs; it is not an independent test split.

\begin{table*}[t]
  \providecommand{\ding}[1]{{\usefont{U}{pzd}{m}{n}\char#1}} 
  \definecolor{sgnUp}{HTML}{138A8A}\definecolor{sgnDown}{HTML}{F07C6C}
  \providecommand{\sgnP}{\textcolor{sgnUp}{\ding{115}}}   
  \providecommand{\sgnN}{\textcolor{sgnDown}{\ding{116}}} 
  \providecommand{\sgns}[3]{\if#1+\sgnP\else\sgnN\fi\,\if#2+\sgnP\else\sgnN\fi\,\if#3+\sgnP\else\sgnN\fi}
  \centering
  \caption{\textbf{Deployment report card of the core suite.}
    \NR closed loop and main run unless marked (s1/s2/rev: main run, second seed, reverse order).
    $\Delta$: final minus \Czero; \rob: the interval also excludes zero under a log-cluster bootstrap and a sign-flip test.
    No interval lies above \Czero.}
  \label{tab:reportcard}
  \scriptsize
  \setlength{\tabcolsep}{2.0pt}
  \begin{tabular}{@{}l r r c c c c c c c c@{}}
    \toprule
    & \multicolumn{2}{c}{Cost} & \multicolumn{5}{c}{Closed loop, \NR} & \IDMR & \multicolumn{2}{c}{Open loop, cached ADE} \\
    \cmidrule(lr){2-3}\cmidrule(lr){4-8}\cmidrule(lr){9-9}\cmidrule(l){10-11}
    Regime & Params & Visits
      & \makecell{Final \CLS\\s1/s2/rev $\uparrow$}
      & \makecell{$\Delta$\CLS sign\\s1/s2/rev}
      & \makecell{$\Delta$\CLS [95\% CI]\\main run $\uparrow$}
      & \makecell{\SR\\lost/gained}
      & \makecell{Stage-avg.\,/\,worst\\\CLS (stage) $\uparrow$}
      & \makecell{$\Delta$\CLS\\$\uparrow$}
      & \makecell{Final (m)\\s1/s2/rev $\downarrow$}
      & \makecell{Forgetting\\(m) $\downarrow$} \\
    \midrule
    \multicolumn{11}{@{}l}{\textit{\PlanTF}} \\
    \Czero & 2,023,073 & -- & 0.528 & -- & -- & 32/56 pass & -- & 0.557 & 3.561 & -- \\
    \FullFT & 100\% & 8.28M & 0.258/0.289/0.364 & \sgns{-}{-}{-} & $-0.270$ \ci{-0.395}{-0.149}\rob & 19/4 & 0.257\,/\,0.166 (4) & $-0.247$\rob & 5.489/4.780/3.367 & 2.279 \\
    \AdaFT & 1.62\% & 8.28M & 0.378/0.321/0.454 & \sgns{-}{-}{-} & $-0.150$ \ci{-0.263}{-0.039}\rob & 12/4 & 0.368\,/\,0.198 (6) & $-0.057$\phantom{\rob} & 4.842/4.432/3.248 & 1.906 \\
    +\Replay & 1.62\% & 9.85M & 0.454/0.428/0.482 & \sgns{-}{-}{-} & $-0.074$ \ci{-0.182}{+0.036}\phantom{\rob} & 9/5 & 0.436\,/\,0.362 (4) & $-0.009$\phantom{\rob} & 3.034/3.031/3.063 & 0.052 \\
    +\PEGP & 1.62\% & 8.28M\textsuperscript{a} & 0.406/0.367/0.510 & \sgns{-}{-}{-} & $-0.122$ \ci{-0.216}{-0.035}\rob & 9/3 & 0.406\,/\,0.289 (4) & $-0.050$\phantom{\rob} & 4.208/4.093/3.364 & 1.207 \\
    \midrule
    \multicolumn{11}{@{}l}{\textit{\DiffP}} \\
    \Czero & 6,042,628 & -- & 0.823 & -- & -- & 48/56 pass & -- & 0.829 & 2.978 & -- \\
    \FullFT & 100\% & 8.28M & 0.538/0.488/0.679 & \sgns{-}{-}{-} & $-0.286$ \ci{-0.395}{-0.179}\rob & 14/3 & 0.673\,/\,0.538 (7) & $-0.390$\rob & 8.372/8.360/4.086 & 5.324 \\
    \AdaFT & 0.61\% & 8.28M & 0.318/0.364/0.776 & \sgns{-}{-}{-} & $-0.505$ \ci{-0.623}{-0.378}\rob & 28/2 & 0.726\,/\,0.318 (7) & $-0.521$\rob & 11.165/10.776/3.762 & 8.758 \\
    +\Replay & 0.61\% & 9.85M & 0.828/0.824/0.786 & \sgns{+}{+}{-} & $+0.005$ \ci{-0.042}{+0.062}\phantom{\rob} & 1/2 & 0.807\,/\,0.764 (4) & $-0.044$\phantom{\rob} & 3.842/3.935/3.765 & 0.319 \\
    +\PEGP & 0.61\% & 8.28M\textsuperscript{a} & 0.832/0.745/0.821 & \sgns{+}{-}{-} & $+0.009$ \ci{-0.054}{+0.074}\phantom{\rob} & 1/3 & 0.810\,/\,0.775 (1) & $-0.030$\phantom{\rob} & 3.907/4.014/3.877 & 0.402 \\
    \bottomrule
  \end{tabular}

  \vspace{2pt}
  \begin{minipage}{\textwidth}
    \scriptsize
    $\uparrow$/$\downarrow$: higher/lower is better; \sgnP/\sgnN: final \CLS above/below \Czero; \Czero rows give \Czero levels.
    Visits per run (32,420 updates); \Replay adds visits from at most 3,072 stored samples; \textsuperscript{a}plus a 73,728\,B (\PlanTF) or 79,872\,B (\DiffP) projection basis.
    Intervals: paired scenario bootstrap (panel uncertainty, not training variance).
    \SR: gate pass rate, not route completion; lost/gained: \Czero-passed scenarios failing at the end, and vice versa.
    \CLS: 20-s segments, not leaderboard-comparable.
    Forgetting: \Fgt over the six old groups.
  \end{minipage}
\end{table*}

\section{Measurement Axes, Verdicts, and Statistics}
\label{sec:axes}

Let \Qij be the score of checkpoint \Tk{i} on the group learned at stage~$j$ ($\Tk{0}=\Czero$), with ADE and FDE negated so that higher is better.
The macro $\bar Q(k)=\frac17\sum_{j}Q(k,j)$ of \Tk{k} weights the seven target groups (56 core-panel \emph{targets}) equally; the sentinel is reported separately.
With acquisition $\Acq(j)=Q(j,j)-Q(j{-}1,j)$ and forgetting $\Fgt(j)=\nobreak\max_{i\ge j}Q(i,j)-Q(7,j)$,
\begin{align}
\BWT &= \tfrac16\textstyle\sum_{j=1}^{6}\,[Q(7,j)-Q(j,j)], \nonumber\\
\Dend &= \tfrac17\textstyle\sum_{j=1}^{7}\,[Q(j,j)-Q(0,j)], \nonumber\\
\Dfin &= \bar Q(7)-\bar Q(0) = \Dend+\tfrac67\,\BWT, \label{eq:identity}\\
\bar Q_{\mathrm{avg}} &= \tfrac17\textstyle\sum_{k=1}^{7}\bar Q(k), \quad \bar Q_{\mathrm{worst}}=\textstyle\min_{k\ge1}\bar Q(k), \nonumber
\end{align}
where macro \Acq averages over seven groups and macro \Fgt averages over six. \BWTold averages the \BWT terms of stages 2--6, which hold the same five groups in the forward and reverse orders.
\BWT is GEM's backward transfer~\cite{lopezpaz2017gem}; unlike RWalk's forgetting~\cite{chaudhry2018rwalk}, \Fgt's maximum includes the final checkpoint, so $\Fgt\ge0$; across orders we compare only fixed-set statistics.

\paragraph{Axes.}
Every verdict fixes four choices (\cref{fig:teaser}): (A)~loop; (B)~reference: the previous checkpoint (\Acq), the learning endpoint (\BWT, \BWTold), its later best (\Fgt) or \Czero (\Dend, \Dfin, $\bar Q(k)-\bar Q(0)$, \SR pass-to-fail and fail-to-pass counts, sentinel retention); (C)~aggregation: $\bar Q(7)$, $\bar Q_{\mathrm{avg}}$ or $\bar Q_{\mathrm{worst}}$; (D)~stream: task order or training seed.
The deployment reading fixes closed loop (\NR, with \IDMR as a check), the \Czero reference and the final checkpoint (\cref{tab:reportcard}).

\paragraph{Verdicts and flips.}
A verdict (per run or pooled) is better, no detectable change, or worse, as the 95\% interval lies above, straddles or lies below zero; straddling does not establish equality.
Changing one axis flips a verdict by a \emph{sign reversal} (better$\leftrightarrow$worse, $A{>}B\leftrightarrow B{>}A$) or a \emph{detection change} (no detectable change$\leftrightarrow$better or worse), which alone does not show two estimates differ.
In \cref{fig:verdicts}, flip (a) holds in sign in all three runs (\PlanTF \Replay), and (b)--(d) are detection changes supported by intervals.

\paragraph{Uncertainty.}
Intervals are 95\% percentile intervals of a paired, task-stratified scenario bootstrap ($B=10{,}000$; figure whiskers use $B=100{,}000$), reflecting evaluation-panel uncertainty for one training run, not training variance; statistics pooled over runs use a scenario-cluster bootstrap.
Intervals marked \rob{} also exclude zero under a log-cluster bootstrap and have a sign-flip $p<0.05$ (Holm-adjusted when regimes are compared); \SR transitions use exact McNemar tests.
Sign counts over runs are not independent tests (runs share \Czero, the panel and, for adapters, \Tk{1}); we qualify results as holding \emph{in all three runs}, \emph{in both forward-order seeds} or \emph{in the main run}, or mark them \emph{(descriptive)}.

\section{Results}
\label{sec:results}

\subsection{The deployment verdict}
\label{sec:deploy}

\paragraph{An invariant in the core suite: no regime measurably improved on \Czero.}
No \Czero-referenced closed-loop macro interval lies above zero: not the 96 \Dfin intervals (24 runs~$\times$ \{\CLS, \SR{}\}~$\times$ \{\NR, \IDMR{}\}), the 96 \Dend intervals, the 52 distinct main-run checkpoints, or the main-run stage-averaged and worst-stage readings.
The closest are two \SR cells at \ci{0.000}{0.089} (two scenarios gained, none lost; exact McNemar $p=0.5$) and, picked in hindsight, the largest main-run \CLS-\NR checkpoint gain, $+0.032$ \ci{-0.079}{+0.144} (\PlanTF \PEGP, stage~2).
\PlanTF is the headroom case: its lower-scoring \Czero leaves room, and its open-loop acquisition is positive in all 12 runs, yet all 48 of its closed-loop \Dfin point estimates are negative, with more pass-to-fail than fail-to-pass transitions in every run$\times$mode cell (146 \vs 56 under \NR).
\DiffP is the ceiling case (\SR 48/56 at \Czero): \Replay shows no detectable change in all three runs and \PEGP in two of three (second seed: $-0.078$ \ci{-0.137}{-0.028}).
The same verdict in both cases argues against a pure ceiling effect.

\paragraph{Scope and power.}
The invariant concerns aggregate readings referenced to \Czero: single updates can measurably beat the previous checkpoint (\PlanTF \AdaFT's last update: $+0.179$ \ci{+0.096}{+0.273}\rob, main run), and single-group readings (8 scenarios each) stay descriptive (supplement).
Main-run \NR intervals exclude \CLS gains above $+0.074$ for all regimes, but the panel resolves only about $\pm0.115$ (\PlanTF) and $\pm0.070$ (\DiffP; median half-widths): a paired test on 56 targets has 80\% power only for changes of about 0.17 and 0.14 \CLS, and $+0.05$ would need about 90 and 63 scenarios per group.

\paragraph{Expanded panel and joint references.}
On the 630 targets of the expanded panel (90 per group), every main-run final checkpoint scores below \Czero in \CLS and \SR under both modes, including \DiffP \Replay and \PEGP (\CLS-\NR 0.820 and 0.829 \vs 0.851), which show no detectable change on the core panel.
Joint training recovers much of the sequential loss in final \CLS-\NR (\PlanTF \FullFT 0.480 \vs 0.258 sequentially; \DiffP \AdaFT 0.822 \vs 0.318), yet no joint reference exceeds its \Czero in \CLS: sequential training accounts for much of the regression but not for the absence of a gain, and joint runs are references, not upper bounds (point estimates; supplement).

\subsection{Axis A: evaluation loop}
\label{sec:loop}

\paragraph{Signal without calibration.}
Open-loop error, a known weak proxy for closed-loop driving~\cite{codevilla2018offline,dauner2023parting}, carries signal but no calibration against \Czero.
It ranks \PlanTF runs, none of which collapses, similarly to closed loop (Spearman $\rho=+0.77$ between cached-ADE and \CLS-\NR changes; descriptive) and flags \DiffP's four collapsed runs, which drive its $\rho=+0.78$ ($+0.29$ without them; \cref{fig:verdicts}a).
For \PlanTF, cached ADE rates the final checkpoint better than \Czero in 6 of 12 runs, while all four closed-loop measures rate it lower in all 12.
The six are the four reverse-order runs, which end on the \PlanTF pretraining group \gStraight, and \Replay in both forward-order seeds.
In the main run, \PlanTF \Replay improves \ADEs by 1.526\,m \ci{+0.30}{+3.06}\rob{} while \CLS-\NR falls from 0.528 to 0.454.
As point estimates, \PlanTF's flip recurs for \Replay and \PEGP in the third order, which ends on \gPickup, for \Replay on the expanded panel (\ADEs 2.904 \vs 3.735\,m) and, without a sequential stream, for both joint references (cached ADE 3.233 and 2.946 \vs 3.561\,m), each below \Czero in every closed-loop reading.
Within one update, the loops can also disagree the other way: \PlanTF \AdaFT's last update (\gStationary, main run) worsens \ADEs on the 48 other targets ($-1.56$\,m in $Q$, \ci{-2.88}{-0.46}\rob) but raises their \CLS-\NR ($+0.226$ \ci{+0.131}{+0.327}\rob), and \SR loses none and gains 13 of them.
Among \Czero-passed run--target pairs, \PlanTF pass-to-fail transitions are not detectably rarer when \ADEs improves (0.34 \vs 0.42, difference \ci{-0.07}{+0.23}); for \DiffP, they are likelier when \ADEs worsens (all pairs: \cref{fig:loopref}b).

\paragraph{Open-loop release gates approve regressions.}
Point-estimate gates on cached ADE and \ADEs each approve 6 of the 12 \PlanTF runs (7 distinct runs in total), all below \Czero in every closed-loop point estimate (three detectably); an interval gate approves 2 on \ADEs (both $-0.074$ in \CLS-\NR, no detectable change) and none on \PDMS: open loop ranks runs and flags collapse but does not certify gains over \Czero.

\subsection{Axis B: reference checkpoint}
\label{sec:reference}

\begin{figure*}[t]
  \centering
  \begin{minipage}[c]{4.64in}
    \includegraphics[width=4.64in]{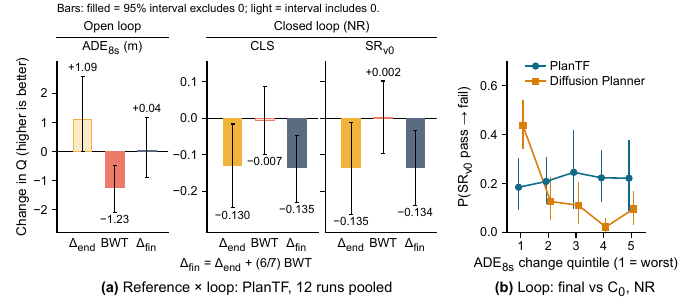}
  \end{minipage}\hfill
  \begin{minipage}[c]{2.10in}
    \caption{\textbf{Loop and reference on the same checkpoints.}
      (a)~\PlanTF, pooled over 12 runs: \Dend, \BWT and \Dfin for \ADEs (negated, m), \CLS-\NR and \SR-\NR (scenario-cluster 95\% intervals); each triple obeys \cref{eq:identity}, and the closed-loop loss sits in \Dend, which \BWT excludes.
      (b)~Share of all run--target pairs (12 runs~$\times$ 56 targets per planner) that pass \SR at \Czero and fail at the final checkpoint (\NR), by quintile of the scenario's \ADEs change (1 = largest increase; scenario-cluster intervals).}
    \label{fig:loopref}
  \end{minipage}
\end{figure*}

\paragraph{A choice of question.}
Pooled over the 12 \PlanTF runs (scenario-cluster bootstrap; \cref{fig:loopref}a), non-reactive closed loop shows no detectable forgetting against the learning endpoint, yet the same checkpoints are worse than \Czero.
Closed loop is already worse at the learning endpoint and changes little afterwards (\CLS-\NR: \Dend $-0.130$ \ci{-0.245}{-0.015}, \BWT $-0.007$ \ci{-0.096}{+0.086}, \Dfin $-0.135$ \ci{-0.228}{-0.047}), whereas open loop improves there by the point estimate and then forgets (\ADEs: \Dend $+1.09$\,m \ci{-0.02}{+2.56}, \BWT $-1.23$\,m \ci{-2.09}{-0.49}): the closed-loop loss sits in \Dend, which \BWT excludes by construction.
In the main run, the first update, on the \Czero pretraining group \gStraight, carries 44--93\% of the \Czero-to-endpoint \CLS-\NR loss (\FullFT: $-0.217$ \ci{-0.334}{-0.099}, possibly affected by BatchNorm); measured from \Tk{1} instead, no \PlanTF regime shows a detectable \NR closed-loop change at the learning endpoint or the final checkpoint.

\paragraph{A flat \BWT hides churn.}
In the main run, a \PlanTF update flips \SR on 26--29\% of bystander scenarios (targets outside the learned group) on average, with a net \SR change of only $-0.006$ to $-0.060$. Over the 12 \PlanTF runs, the 576 old-group pairs behind \BWT lose 116 passes and gain 117; as two seeds of one regime disagree on 4--13 of 56 targets at \Tk{7}, churn needs a seed-to-seed floor.

\paragraph{Sentinel retention.}
\DiffP \Replay and \PEGP forget least on the target groups (0.319 and 0.402\,m \vs 8.758\,m for \AdaFT; main run, cached ADE), yet their cached ADE on the evaluation-only sentinel rises from 4.696 to 7.916 and 7.249\,m, 55--69\% of it already at the shared \Tk{1}, where it is confounded with the switch from EMA to raw weights; endpoint references cannot register this loss.

\begin{figure*}[t]
  \centering
  \includegraphics[width=\linewidth]{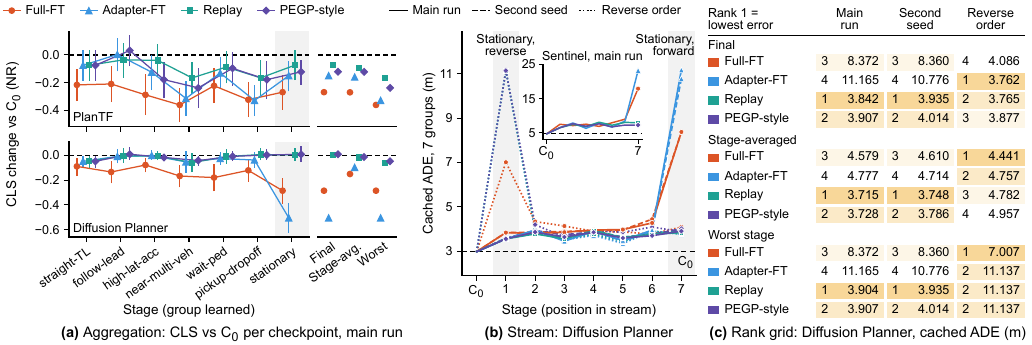}
  \caption{\textbf{Aggregation and order on the same checkpoints.}
    (a)~Change in non-reactive \CLS from \Czero at each main-run checkpoint (56 targets; 95\% intervals), with final, stage-averaged and worst-stage values at the right; shaded: \gStationary stage.
    (b)~\DiffP 7-group cached ADE (m) by stage: main run (solid), second seed (dashed), reverse order (dotted; adapters share \Tk{1}); black dashed: \Czero; inset: sentinel, main run.
    (c)~\DiffP regime ranks in cached ADE per summary and run (reverse worst-stage tie). (b), (c): descriptive.}
  \label{fig:aggorder}
\end{figure*}

\subsection{Axis C: temporal aggregation}
\label{sec:aggregation}

Final, stage-averaged and worst-stage summaries of the same checkpoints can order regimes differently (\cref{fig:aggorder}a).
In the main run, final \CLS-\NR cannot separate \PlanTF \Replay from \AdaFT (\Replay minus \AdaFT: $+0.076$ \ci{-0.033}{+0.186}), but the stage average can ($+0.068$ \ci{+0.022}{+0.114}\rob).
Pooling seven checkpoints narrows the interval and registers mid-stream losses of \AdaFT (worst stage 0.198) that its last update partly undoes ($+0.179$).
For \DiffP (main run), \FullFT minus \AdaFT reverses from $+0.220$ \ci{+0.124}{+0.316}\rob{} at the final checkpoint to $-0.054$ \ci{-0.101}{-0.008} stage-averaged (borderline: Holm-adjusted sign-flip $p=0.08$), as \AdaFT tracks \Replay and \PEGP until its \gStationary update at stage~7 ($-0.468$ \ci{-0.587}{-0.344}\rob) while \FullFT degrades early.
What does not flip (descriptive): \PlanTF's cached ADE ranks \Replay first and \FullFT last under every summary, in all three runs and in the third order.
Aggregation also interacts with order: \DiffP \Replay and \PEGP lead \FullFT in final cached ADE in all three runs, but trail it in reverse order under the other two summaries (\cref{fig:aggorder}c; descriptive).

\subsection{Axis D: stream protocol}
\label{sec:order}

Final rankings change far more across task orders than across the two training seeds, which we compare only in the forward order.
Over nine open- and closed-loop metric families, the mean Kendall $\tau_b$ of final regime rankings is 0.95 (\PlanTF) and 0.84 (\DiffP) between the seeds, but 0.41 and $-0.03$ between the forward and reverse orders, 0.65 and 0.29 between the forward and third, and 0.36 and 0.28 between the reverse and third (descriptive); the four closed-loop families alone give 0.89 and 0.73 between the seeds but 0.27 and 0.52 between the forward and reverse orders.
\DiffP \Replay minus \AdaFT in final \CLS-\NR is $+0.510$ \ci{+0.398}{+0.619} and $+0.460$ \ci{+0.347}{+0.572} in the forward-order seeds but $+0.011$ \ci{-0.013}{+0.051} in reverse order, an interval that excludes both forward gaps but does not show equality.
Seed is weaker than order but not inert: 3 (\DiffP) and 1 (\PlanTF) of 54 pairwise final comparisons per planner change sign only between seeds, against 26 and 15 only under reversal, and only 9 of 83 tested regime orderings hold under both a seed change and reversal (criteria in the supplement).
Order comparisons also change preceding states and random streams, so they show order sensitivity, not an effect of position alone.

\paragraph{A third order.}
Moving \gStationary to stage~4 changes when its damage occurs, not whether: its update again collapses \DiffP \AdaFT ($+9.90$\,m cached ADE on the other six groups) but not \Replay or \PEGP, and the next update undoes it.
\Replay thus ends only 0.046 \CLS-\NR above \AdaFT (0.510 in the main run), while their worst stages in cached ADE differ threefold (12.59 \vs 4.09\,m): whether protection shows at the final checkpoint depends on where the damaging update sits, whereas cached-ADE worst-stage and stage-averaged readings show it.
Margins need not shrink (\PlanTF \PEGP leads \AdaFT by 0.144 \CLS-\NR, 0.028 in the main run), and no third-order final checkpoint exceeds \Czero in closed loop (point estimates).

\subsection{What drives the flips}
\label{sec:why}

For \DiffP, one update dominates (\cref{fig:aggorder}b): training on \gStationary worsens \AdaFT's mean cached ADE on the other six target groups by 8.22 and 7.86\,m when it comes last, 9.32\,m when first and 9.90\,m when fourth, and on the sentinel by up to 18.6\,m, producing 96\% (cached ADE) and 92\% (\CLS-\NR) of \AdaFT's \BWT in the main run.
As these contrasts change position, preceding state and random streams together, they show that the damage followed the group, not its position; the position decided only whether later updates could undo it before the final checkpoint.
\Replay and \PEGP act only from stage~2, so they share \AdaFT's \Tk{1} and collapse when \gStationary opens the stream; when it comes last, they remove 98.8\% and 97.1\% of its macro shock (main run).
Adapters alone gave no protection against this update: \AdaFT suffered 1.75, 1.75 and 2.02 times the collateral damage of \FullFT in the three runs (cached ADE).
In the equal-data runs, this update still adds 3.94\,m to \AdaFT's error on the other six groups, half the full-pool 8.22\,m yet far more than any other update there, and \Replay and \PEGP again remove it ($-0.04$ and $+0.05$\,m): the damage tracks the group's content, not only its size, although these runs also shrink every group and the budget (point estimates).

As we train without EMA (\cref{sec:deployment}), the collapse may reflect the recipe, but not one that fails on every group: one group triggers it, the next update largely undoes it in the reverse and third orders, and \Replay and \PEGP avoid it at identical hyperparameters unless it comes first; a \gStationary-specific failure without EMA remains possible, as we did not test EMA during updates.

\PlanTF differs: its closed-loop loss already appears at the learning endpoint (\cref{sec:reference}), and in both forward-order seeds its last update gives only 27--41\% of \FullFT's and \AdaFT's cached-ADE \BWT on groups learned at stages 1--5.

\paragraph{LoRA branch.}
In the forward order, where \gStationary comes last, LoRA repeats the adapter pattern: \DiffP \LoRAFT collapses, though less than \AdaFT (final \CLS-\NR 0.527 \vs 0.318; \Czero 0.823), and \LoRAReplay and \LoRAPEGP end at 0.762, below their adapter counterparts (0.828 and 0.832).
In the reverse and third orders, \LoRAFT ends within 0.03 \CLS-\NR of \Czero for both planners and replay or projection adds at most $+0.013$ in \CLS-\NR; several of these runs end above \Czero in some closed-loop readings (four in \CLS-\NR, by at most $+0.022$; under \IDMR, up to $+0.035$ in \CLS and 51 \vs 48 of 56 in \SR), within the panel's resolution (\cref{sec:deploy}); we do not extend the core-suite verdict to this branch (point estimates; supplement).

\section{Benchmark Contract and Reporting Rules}
\label{sec:contract}

\paragraph{Contract.}
A \CD v1 result must
\begin{enumerate*}[label=(\arabic*)]
\item start from the specified \Czero checkpoints and report their scores;
\item run forward, reverse and non-mirrored orders and at least two forward-order seeds;
\item use 32,420 updates per run or state the budget;
\item evaluate the cached set at every stage and closed loop (\NR, \IDMR) at \Czero, before and after each group and at \Tk{7}, with full \NR stage matrices behind stage-averaged or worst-stage claims;
\item report closed-loop \Dfin and \Dend with paired intervals, \SR pass-to-fail and fail-to-\allowbreak pass counts, final, stage-averaged and worst-stage values, fixed-set \BWT across orders and sentinel retention, with open loop as a diagnostic;
\item select no hyperparameter or checkpoint on development data, and report each run and seed separately.
\end{enumerate*}
With the third order, our four core regimes meet (1)--(4); only the core suite has the paired intervals of (5), and the development split informed our design (6).

\paragraph{Reporting rules.}
\textbf{R1.}~Let closed loop decide and open loop detect collapse (\cref{sec:loop}).
\textbf{R2.}~Anchor verdicts to \Czero and count \SR pass-to-fail transitions~\cite{yan2021pct,li2026deitp} against a seed-to-seed floor, next to \CLS (\cref{sec:reference}).
\textbf{R3.}~Report final, stage-averaged and worst-stage values (\cref{sec:aggregation}).
\textbf{R4.}~Run several seeds and orders, one non-mirrored, and compare orders only with fixed-set statistics (\cref{sec:order}).
\textbf{R5.}~Keep an evaluation-only sentinel group (\cref{sec:reference}).
\textbf{R6.}~Report visits, trainable parameters, memory, projection rank and cap per stage, and whether constraints act on gradients or optimizer updates (\cref{sec:protocol}).
\textbf{R7.}~Disclose group composition, \Czero exposure, and sampling and EMA settings (\cref{sec:protocol}).

\section{Conclusion}

\label{sec:conclusion}
\CD shows that verdicts on the same checkpoints of two planners can reverse or vanish when one of four axes changes: task orders moved final rankings far more than training seeds, and \DiffP's replay and projection kept a large final-checkpoint lead over \AdaFT only when the most damaging group came last.
Yet no aggregate closed-loop reading of the 24-run core suite improved measurably on \Czero under our update-matched recipe, every evaluated final checkpoint on the expanded panel trails it, and joint training does not exceed its \CLS.\label{end-of-main}

\newpage
\hyphenation{OpenAI}
\section*{LLM Usage Statement}
We used Claude (Anthropic) to edit, restructure
and condense the text of each section, to search for and summarize related work, and to write scripts that lay out and render figures from our experimental outputs. The authors designed the method and experiments, conducted all experiments and analyses, and verified every statement, number, citation and figure in the paper. No LLM was used to construct the benchmark data or to form the research ideas, hypotheses or conclusions. All AI-assisted code was reviewed and tested by the authors. The authors take full responsibility for the accuracy, integrity, and final content of this work.

\par
{
    \small
    \bibliographystyle{ieeenat_fullname}
    \bibliography{main}

@inproceedings{du2026dell,
  title     = {Deconfounded Lifelong Learning for Autonomous Driving via Dynamic Knowledge Spaces},
  author    = {Jiayuan Du and Yuebing Song and Yiming Zhao and Xianghui Pan and Jiawei Lian and Yuchu Lu and Liuyi Wang and Chengju Liu and Qijun Chen},
  booktitle = {European Conference on Computer Vision (ECCV)},
  year      = {2026},
  eprint    = {2603.14354},
  archiveprefix = {arXiv}
}

@inproceedings{lai2024infdriver,
  title     = {Uncertainty-Guided Never-Ending Learning to Drive},
  author    = {Lei Lai and Eshed Ohn-Bar and Sanjay Arora and John Seon Keun Yi},
  booktitle = {IEEE/CVF Conference on Computer Vision and Pattern Recognition (CVPR)},
  pages     = {15088--15098},
  year      = {2024},
  doi       = {10.1109/CVPR52733.2024.01429}
}

@article{verwimp2023clad,
  title     = {{CLAD}: A Realistic Continual Learning Benchmark for Autonomous Driving},
  author    = {Eli Verwimp and Kuo Yang and Sarah Parisot and Lanqing Hong and Steven McDonagh and Eduardo P{\'e}rez-Pellitero and Matthias {De Lange} and Tinne Tuytelaars},
  journal   = {Neural Networks},
  volume    = {161},
  pages     = {659--669},
  year      = {2023},
  doi       = {10.1016/j.neunet.2023.02.001},
  eprint    = {2210.03482},
  archiveprefix = {arXiv}
}

@article{yao2025lilodriver,
  title     = {{LiloDriver}: A Lifelong Learning Framework for Closed-loop Motion Planning in Long-tail Autonomous Driving Scenarios},
  author    = {Huaiyuan Yao and Pengfei Li and Bu Jin and Yupeng Zheng and An Liu and Lisen Mu and Qing Su and Qian Zhang and Yilun Chen and Peng Li},
  journal   = {arXiv preprint arXiv:2505.17209},
  year      = {2025}
}

@inproceedings{mao2026blindspot,
  title     = {The Blind Spot of Adaptation: Quantifying and Mitigating Forgetting in Fine-tuned Driving Models},
  author    = {Runhao Mao and Hanshi Wang and Yixiang Yang and Qianli Ma and Jingmeng Zhou and Zhipeng Zhang},
  booktitle = {IEEE/CVF Conference on Computer Vision and Pattern Recognition (CVPR)},
  year      = {2026},
  eprint    = {2604.04857},
  archiveprefix = {arXiv}
}

@article{ma2021continual,
  title     = {Continual Multi-Agent Interaction Behavior Prediction With Conditional Generative Memory},
  author    = {Hengbo Ma and Yaofeng Sun and Jiachen Li and Masayoshi Tomizuka and Chiho Choi},
  journal   = {IEEE Robotics and Automation Letters},
  volume    = {6},
  number    = {4},
  pages     = {8410--8417},
  year      = {2021},
  doi       = {10.1109/LRA.2021.3104334}
}

@article{lin2024dgsm,
  title     = {Continual Interactive Behavior Learning With Traffic Divergence Measurement: A Dynamic Gradient Scenario Memory Approach},
  author    = {Yunlong Lin and Zirui Li and Cheng Gong and Chao Lu and Xinwei Wang and Jianwei Gong},
  journal   = {IEEE Transactions on Intelligent Transportation Systems},
  volume    = {25},
  number    = {3},
  pages     = {2355--2372},
  year      = {2024},
  doi       = {10.1109/TITS.2023.3324983},
  eprint    = {2212.11167},
  archiveprefix = {arXiv},
  primaryclass  = {cs.RO}
}

@article{lin2026h2c,
  title     = {{H2C}: Hippocampal Circuit-Inspired Continual Learning for Lifelong Trajectory Prediction in Autonomous Driving},
  author    = {Yunlong Lin and Zirui Li and Guodong Du and Xiaocong Zhao and Cheng Gong and Xinwei Wang and Chao Lu and Jianwei Gong},
  journal   = {IEEE Transactions on Intelligent Transportation Systems},
  volume    = {27},
  number    = {8},
  pages     = {9834--9851},
  year      = {2026},
  doi       = {10.1109/TITS.2026.3679635},
  eprint    = {2508.01158},
  archiveprefix = {arXiv},
  primaryclass  = {cs.AI}
}

@article{li2026deitp,
  title     = {Toward Zero-Forget Continual Learning for Interactive Trajectory Prediction: A Dynamically Expandable Approach},
  author    = {Huiqian Li and Xiaozhou Wu and Jin Huang and Zhihua Zhong},
  journal   = {Communications in Transportation Research},
  volume    = {6},
  number    = {1},
  pages     = {9640015},
  year      = {2026},
  doi       = {10.26599/COMMTR.2026.9640015}
}

@article{lin2025syrem,
  title     = {Escaping Stability-Plasticity Dilemma in Online Continual Learning for Motion Forecasting via Synergetic Memory Rehearsal},
  author    = {Yunlong Lin and Chao Lu and Tongshuai Wu and Xiaocong Zhao and Guodong Du and Yanwei Sun and Zirui Li and Jianwei Gong},
  journal   = {IEEE Transactions on Neural Networks and Learning Systems},
  volume    = {37},
  number    = {11},
  pages     = {6580--6594},
  year      = {2026},
  doi       = {10.1109/TNNLS.2026.3700190},
  eprint    = {2508.19571},
  archiveprefix = {arXiv},
  primaryclass  = {cs.LG}
}

@article{schischka2026cimf,
  title     = {Class-Incremental Motion Forecasting},
  author    = {Nicolas Schischka and Nikhil Gosala and B. Ravi Kiran and Senthil Yogamani and Abhinav Valada},
  journal   = {arXiv preprint arXiv:2603.09420},
  year      = {2026}
}

@article{bao2021lifelong,
  title     = {Lifelong Vehicle Trajectory Prediction Framework Based on Generative Replay},
  author    = {Peng Bao and Zonghai Chen and Jikai Wang and Deyun Dai and Hao Zhao},
  journal   = {arXiv preprint arXiv:2111.07511},
  year      = {2021}
}

@article{cao2023dcarl,
  title     = {Continuous improvement of self-driving cars using dynamic confidence-aware reinforcement learning},
  author    = {Zhong Cao and Kun Jiang and Weitao Zhou and Shaobing Xu and Huei Peng and Diange Yang},
  journal   = {Nature Machine Intelligence},
  volume    = {5},
  number    = {2},
  pages     = {145--158},
  year      = {2023},
  doi       = {10.1038/s42256-023-00610-y}
}

@inproceedings{niu2024clic,
  title     = {Continual Driving Policy Optimization with Closed-Loop Individualized Curricula},
  author    = {Haoyi Niu and Yizhou Xu and Xingjian Jiang and Jianming Hu},
  booktitle = {IEEE International Conference on Robotics and Automation (ICRA)},
  pages     = {6850--6857},
  year      = {2024},
  doi       = {10.1109/ICRA57147.2024.10611578},
  eprint    = {2309.14209},
  archiveprefix = {arXiv}
}

@article{cui2025mope,
  title     = {Continual Adaptation for Autonomous Driving with the Mixture of Progressive Experts Network},
  author    = {Yixin Cui and Shuo Yang and Chi Wan and Xincheng Li and Jiaming Xing and Yuanjian Zhang and Yanjun Huang and Hong Chen},
  journal   = {arXiv preprint arXiv:2502.05943},
  year      = {2025}
}

@article{yao2026lunaad,
  title     = {{LUNA-AD}: Lightweight Uncertainty-Aware Language Model with Lifelong Learning for Autonomous Driving},
  author    = {Ruoyu Yao and Pei Liu and Ruiguo Zhong and Mingxing Peng and Rui Yang and Jun Ma},
  journal   = {arXiv preprint arXiv:2606.08470},
  year      = {2026}
}

@article{hu2026sage,
  title     = {Self-Aware Active Learning Enables Continual Improvement in Autonomous Driving},
  author    = {Dong Hu and Chao Huang and Carman K. M. Lee and Dimitrios Kanoulas},
  journal   = {arXiv preprint arXiv:2608.29772},
  year      = {2026}
}

@inproceedings{tang2026coltdrive,
  title     = {{CoLT-Drive}: Counterfactual Long-Tail Benchmarking and Knowledge-Preserving Adaptation for Driving Affordance Prediction},
  author    = {Zhengxu Tang and Guofeng Cui and Ziyu Gong and Xiaozhou Zhang and Ruifeng Deng and Chengzhi Qi and Ke Chen and Sachin Patil and Tianjun Xiao and Langechuan Liu and Pichao Wang},
  booktitle = {Conference on Empirical Methods in Natural Language Processing (EMNLP)},
  year      = {2026},
  eprint    = {2609.00242},
  archiveprefix = {arXiv}
}

@inproceedings{sun2022shift,
  title     = {{SHIFT}: A Synthetic Driving Dataset for Continuous Multi-Task Domain Adaptation},
  author    = {Tao Sun and Mattia Segu and Janis Postels and Yuxuan Wang and Luc {Van Gool} and Bernt Schiele and Federico Tombari and Fisher Yu},
  booktitle = {IEEE/CVF Conference on Computer Vision and Pattern Recognition (CVPR)},
  year      = {2022}
}

@inproceedings{wang2022cotta,
  title     = {Continual Test-Time Domain Adaptation},
  author    = {Qin Wang and Olga Fink and Luc {Van Gool} and Dengxin Dai},
  booktitle = {IEEE/CVF Conference on Computer Vision and Pattern Recognition (CVPR)},
  year      = {2022}
}

@article{caesar2021nuplan,
  title     = {{nuPlan}: A Closed-Loop {ML}-based Planning Benchmark for Autonomous Vehicles},
  author    = {Holger Caesar and Juraj Kabzan and Kok Seang Tan and Whye Kit Fong and Eric Wolff and Alex Lang and Luke Fletcher and Oscar Beijbom and Sammy Omari},
  journal   = {arXiv preprint arXiv:2106.11810},
  year      = {2021}
}

@inproceedings{karnchanachari2024nuplan,
  title     = {Towards Learning-Based Planning: The {nuPlan} Benchmark for Real-World Autonomous Driving},
  author    = {Napat Karnchanachari and Dimitris Geromichalos and Kok Seang Tan and Nanxiang Li and Christopher Eriksen and Shakiba Yaghoubi and Noushin Mehdipour and Gianmarco Bernasconi and Whye Kit Fong and Yiluan Guo and Holger Caesar},
  booktitle = {IEEE International Conference on Robotics and Automation (ICRA)},
  pages     = {629--636},
  year      = {2024},
  doi       = {10.1109/ICRA57147.2024.10610077}
}

@inproceedings{peng2026nuplanr,
  title     = {{nuPlan-R}: A Closed-Loop Planning Benchmark for Autonomous Driving via Reactive Multi-Agent Simulation},
  author    = {Mingxing Peng and Ruoyu Yao and Xusen Guo and Jun Ma},
  booktitle = {IEEE Intelligent Vehicles Symposium (IV)},
  pages     = {702--709},
  year      = {2026},
  doi       = {10.1109/IV66570.2026.11623929}
}

@article{treiber2000idm,
  title     = {Congested Traffic States in Empirical Observations and Microscopic Simulations},
  author    = {Martin Treiber and Ansgar Hennecke and Dirk Helbing},
  journal   = {Physical Review E},
  volume    = {62},
  pages     = {1805--1824},
  year      = {2000},
  doi       = {10.1103/PhysRevE.62.1805}
}

@inproceedings{dauner2023parting,
  title     = {Parting with Misconceptions about Learning-based Vehicle Motion Planning},
  author    = {Daniel Dauner and Marcel Hallgarten and Andreas Geiger and Kashyap Chitta},
  booktitle = {Conference on Robot Learning (CoRL)},
  series    = {Proceedings of Machine Learning Research},
  volume    = {229},
  pages     = {1268--1281},
  year      = {2023}
}

@inproceedings{cheng2024plantf,
  title     = {Rethinking Imitation-based Planners for Autonomous Driving},
  author    = {Jie Cheng and Yingbing Chen and Xiaodong Mei and Bowen Yang and Bo Li and Ming Liu},
  booktitle = {IEEE International Conference on Robotics and Automation (ICRA)},
  pages     = {14123--14130},
  year      = {2024},
  doi       = {10.1109/ICRA57147.2024.10611364}
}

@inproceedings{zheng2025diffusionplanner,
  title     = {Diffusion-Based Planning for Autonomous Driving with Flexible Guidance},
  author    = {Yinan Zheng and Ruiming Liang and Kexin Zheng and Jinliang Zheng and Liyuan Mao and Jianxiong Li and Weihao Gu and Rui Ai and Shengbo Eben Li and Xianyuan Zhan and Jingjing Liu},
  booktitle = {International Conference on Learning Representations (ICLR)},
  year      = {2025}
}

@inproceedings{dauner2024navsim,
  title     = {{NAVSIM}: Data-Driven Non-Reactive Autonomous Vehicle Simulation and Benchmarking},
  author    = {Daniel Dauner and Marcel Hallgarten and Tianyu Li and Xinshuo Weng and Zhiyu Huang and Zetong Yang and Hongyang Li and Igor Gilitschenski and Boris Ivanovic and Marco Pavone and Andreas Geiger and Kashyap Chitta},
  booktitle = {Advances in Neural Information Processing Systems (NeurIPS) Datasets and Benchmarks Track},
  year      = {2024}
}

@inproceedings{cao2025pseudosim,
  title     = {Pseudo-Simulation for Autonomous Driving},
  author    = {Wei Cao and Marcel Hallgarten and Tianyu Li and Daniel Dauner and Xunjiang Gu and Caojun Wang and Yakov Miron and Marco Aiello and Hongyang Li and Igor Gilitschenski and Boris Ivanovic and Marco Pavone and Andreas Geiger and Kashyap Chitta},
  booktitle = {Conference on Robot Learning (CoRL)},
  year      = {2025}
}

@inproceedings{hallgarten2024interplan,
  title     = {Can Vehicle Motion Planning Generalize to Realistic Long-tail Scenarios?},
  author    = {Marcel Hallgarten and Julian Zapata and Martin Stoll and Katrin Renz and Andreas Zell},
  booktitle = {IEEE/RSJ International Conference on Intelligent Robots and Systems (IROS)},
  year      = {2024}
}

@inproceedings{jia2024bench2drive,
  title     = {{Bench2Drive}: Towards Multi-Ability Benchmarking of Closed-Loop End-To-End Autonomous Driving},
  author    = {Xiaosong Jia and Zhenjie Yang and Qifeng Li and Zhiyuan Zhang and Junchi Yan},
  booktitle = {Advances in Neural Information Processing Systems (NeurIPS) Datasets and Benchmarks Track},
  year      = {2024}
}

@inproceedings{codevilla2018offline,
  title     = {On Offline Evaluation of Vision-based Driving Models},
  author    = {Felipe Codevilla and Antonio M. L{\'o}pez and Vladlen Koltun and Alexey Dosovitskiy},
  booktitle = {European Conference on Computer Vision (ECCV)},
  year      = {2018},
  doi       = {10.1007/978-3-030-01267-0_15}
}

@article{zhai2023rethinking,
  title     = {Rethinking the Open-Loop Evaluation of End-to-End Autonomous Driving in {nuScenes}},
  author    = {Jiang-Tian Zhai and Ze Feng and Jinhao Du and Yongqiang Mao and Jiang-Jiang Liu and Zichang Tan and Yifu Zhang and Xiaoqing Ye and Jingdong Wang},
  journal   = {arXiv preprint arXiv:2305.10430},
  year      = {2023}
}

@inproceedings{li2024egostatus,
  title     = {Is Ego Status All You Need for Open-Loop End-to-End Autonomous Driving?},
  author    = {Zhiqi Li and Zhiding Yu and Shiyi Lan and Jiahan Li and Jan Kautz and Tong Lu and Jose M. Alvarez},
  booktitle = {IEEE/CVF Conference on Computer Vision and Pattern Recognition (CVPR)},
  year      = {2024}
}

@inproceedings{codevilla2019limitations,
  title     = {Exploring the Limitations of Behavior Cloning for Autonomous Driving},
  author    = {Felipe Codevilla and Eder Santana and Antonio M. L{\'o}pez and Adrien Gaidon},
  booktitle = {IEEE/CVF International Conference on Computer Vision (ICCV)},
  pages     = {9329--9338},
  year      = {2019}
}

@article{chen2024e2esurvey,
  title     = {End-to-End Autonomous Driving: Challenges and Frontiers},
  author    = {Li Chen and Penghao Wu and Kashyap Chitta and Bernhard Jaeger and Andreas Geiger and Hongyang Li},
  journal   = {IEEE Transactions on Pattern Analysis and Machine Intelligence},
  volume    = {46},
  number    = {12},
  pages     = {10164--10183},
  year      = {2024},
  doi       = {10.1109/TPAMI.2024.3435937}
}

@inproceedings{lopezpaz2017gem,
  title     = {Gradient Episodic Memory for Continual Learning},
  author    = {David Lopez-Paz and Marc'Aurelio Ranzato},
  booktitle = {Advances in Neural Information Processing Systems (NeurIPS)},
  year      = {2017}
}

@inproceedings{chaudhry2018rwalk,
  title     = {Riemannian Walk for Incremental Learning: Understanding Forgetting and Intransigence},
  author    = {Arslan Chaudhry and Puneet K. Dokania and Thalaiyasingam Ajanthan and Philip H. S. Torr},
  booktitle = {European Conference on Computer Vision (ECCV)},
  year      = {2018},
  doi       = {10.1007/978-3-030-01252-6_33}
}

@article{chaudhry2019tiny,
  title     = {On Tiny Episodic Memories in Continual Learning},
  author    = {Arslan Chaudhry and Marcus Rohrbach and Mohamed Elhoseiny and Thalaiyasingam Ajanthan and Puneet K. Dokania and Philip H. S. Torr and Marc'Aurelio Ranzato},
  journal   = {arXiv preprint arXiv:1902.10486},
  year      = {2019}
}

@inproceedings{buzzega2020der,
  title     = {Dark Experience for General Continual Learning: a Strong, Simple Baseline},
  author    = {Pietro Buzzega and Matteo Boschini and Angelo Porrello and Davide Abati and Simone Calderara},
  booktitle = {Advances in Neural Information Processing Systems (NeurIPS)},
  year      = {2020}
}

@article{kirkpatrick2017ewc,
  title     = {Overcoming catastrophic forgetting in neural networks},
  author    = {James Kirkpatrick and Razvan Pascanu and Neil Rabinowitz and Joel Veness and Guillaume Desjardins and Andrei A. Rusu and Kieran Milan and John Quan and Tiago Ramalho and Agnieszka Grabska-Barwinska and Demis Hassabis and Claudia Clopath and Dharshan Kumaran and Raia Hadsell},
  journal   = {Proceedings of the National Academy of Sciences},
  volume    = {114},
  number    = {13},
  pages     = {3521--3526},
  year      = {2017},
  doi       = {10.1073/pnas.1611835114}
}

@article{li2018lwf,
  title     = {Learning without Forgetting},
  author    = {Zhizhong Li and Derek Hoiem},
  journal   = {IEEE Transactions on Pattern Analysis and Machine Intelligence},
  volume    = {40},
  number    = {12},
  pages     = {2935--2947},
  year      = {2018},
  doi       = {10.1109/TPAMI.2017.2773081}
}

@inproceedings{saha2021gpm,
  title     = {Gradient Projection Memory for Continual Learning},
  author    = {Gobinda Saha and Isha Garg and Kaushik Roy},
  booktitle = {International Conference on Learning Representations (ICLR)},
  year      = {2021}
}

@inproceedings{wang2021adamnscl,
  title     = {Training Networks in Null Space of Feature Covariance for Continual Learning},
  author    = {Shipeng Wang and Xiaorong Li and Jian Sun and Zongben Xu},
  booktitle = {IEEE/CVF Conference on Computer Vision and Pattern Recognition (CVPR)},
  year      = {2021}
}

@inproceedings{mallya2018packnet,
  title     = {{PackNet}: Adding Multiple Tasks to a Single Network by Iterative Pruning},
  author    = {Arun Mallya and Svetlana Lazebnik},
  booktitle = {IEEE/CVF Conference on Computer Vision and Pattern Recognition (CVPR)},
  year      = {2018},
  doi       = {10.1109/CVPR.2018.00810}
}

@inproceedings{farajtabar2020ogd,
  title     = {Orthogonal Gradient Descent for Continual Learning},
  author    = {Mehrdad Farajtabar and Navid Azizan and Alex Mott and Ang Li},
  booktitle = {International Conference on Artificial Intelligence and Statistics (AISTATS)},
  series    = {Proceedings of Machine Learning Research},
  volume    = {108},
  pages     = {3762--3773},
  year      = {2020}
}

@inproceedings{riemer2019mer,
  title     = {Learning to Learn without Forgetting by Maximizing Transfer and Minimizing Interference},
  author    = {Matthew Riemer and Ignacio Cases and Robert Ajemian and Miao Liu and Irina Rish and Yuhai Tu and Gerald Tesauro},
  booktitle = {International Conference on Learning Representations (ICLR)},
  year      = {2019}
}

@article{diazrodriguez2018metrics,
  title     = {Don't forget, there is more than forgetting: new metrics for Continual Learning},
  author    = {Natalia D{\'\i}az-Rodr{\'\i}guez and Vincenzo Lomonaco and David Filliat and Davide Maltoni},
  journal   = {arXiv preprint arXiv:1810.13166},
  year      = {2018}
}

@inproceedings{toneva2019forgetting,
  title     = {An Empirical Study of Example Forgetting during Deep Neural Network Learning},
  author    = {Mariya Toneva and Alessandro Sordoni and Remi {Tachet des Combes} and Adam Trischler and Yoshua Bengio and Geoffrey J. Gordon},
  booktitle = {International Conference on Learning Representations (ICLR)},
  year      = {2019}
}

@inproceedings{yan2021pct,
  title     = {Positive-Congruent Training: Towards Regression-Free Model Updates},
  author    = {Sijie Yan and Yuanjun Xiong and Kaustav Kundu and Shuo Yang and Siqi Deng and Meng Wang and Wei Xia and Stefano Soatto},
  booktitle = {IEEE/CVF Conference on Computer Vision and Pattern Recognition (CVPR)},
  year      = {2021}
}

@inproceedings{yoon2020apd,
  title     = {Scalable and Order-robust Continual Learning with Additive Parameter Decomposition},
  author    = {Jaehong Yoon and Saehoon Kim and Eunho Yang and Sung Ju Hwang},
  booktitle = {International Conference on Learning Representations (ICLR)},
  year      = {2020}
}

@article{bell2022taskorder,
  title     = {The Effect of Task Ordering in Continual Learning},
  author    = {Samuel J. Bell and Neil D. Lawrence},
  journal   = {arXiv preprint arXiv:2205.13323},
  year      = {2022}
}

@inproceedings{lin2023theory,
  title     = {Theory on Forgetting and Generalization of Continual Learning},
  author    = {Sen Lin and Peizhong Ju and Yingbin Liang and Ness Shroff},
  booktitle = {International Conference on Machine Learning (ICML)},
  series    = {Proceedings of Machine Learning Research},
  volume    = {202},
  pages     = {21078--21100},
  year      = {2023}
}

@inproceedings{evron2022catastrophic,
  title     = {How catastrophic can catastrophic forgetting be in linear regression?},
  author    = {Itay Evron and Edward Moroshko and Rachel Ward and Nati Srebro and Daniel Soudry},
  booktitle = {Conference on Learning Theory (COLT)},
  series    = {Proceedings of Machine Learning Research},
  volume    = {178},
  pages     = {4028--4079},
  year      = {2022}
}

@inproceedings{prabhu2023budgeted,
  title     = {Computationally Budgeted Continual Learning: What Does Matter?},
  author    = {Ameya Prabhu and Hasan Abed Al Kader Hammoud and Puneet K. Dokania and Philip H. S. Torr and Ser-Nam Lim and Bernard Ghanem and Adel Bibi},
  booktitle = {IEEE/CVF Conference on Computer Vision and Pattern Recognition (CVPR)},
  pages     = {3698--3707},
  year      = {2023}
}

@inproceedings{ghunaim2023realtime,
  title     = {Real-Time Evaluation in Online Continual Learning: A New Hope},
  author    = {Yasir Ghunaim and Adel Bibi and Kumail Alhamoud and Motasem Alfarra and Hasan Abed Al Kader Hammoud and Ameya Prabhu and Philip H. S. Torr and Bernard Ghanem},
  booktitle = {IEEE/CVF Conference on Computer Vision and Pattern Recognition (CVPR)},
  pages     = {11888--11897},
  year      = {2023}
}

@inproceedings{qiao2024pgp,
  title     = {Prompt Gradient Projection for Continual Learning},
  author    = {Jingyang Qiao and Zhizhong Zhang and Xin Tan and Chengwei Chen and Yanyun Qu and Yong Peng and Yuan Xie},
  booktitle = {International Conference on Learning Representations (ICLR)},
  year      = {2024}
}

@article{qiao2024pegp,
  title     = {Gradient Projection For Continual Parameter-Efficient Tuning},
  author    = {Jingyang Qiao and Zhizhong Zhang and Xin Tan and Yanyun Qu and Wensheng Zhang and Zhi Han and Yuan Xie},
  journal   = {arXiv preprint arXiv:2405.13383},
  year      = {2024}
}

@article{biderman2024lora,
  title     = {{LoRA} Learns Less and Forgets Less},
  author    = {Dan Biderman and Jacob Portes and Jose Javier {Gonzalez Ortiz} and Mansheej Paul and Philip Greengard and Connor Jennings and Daniel King and Sam Havens and Vitaliy Chiley and Jonathan Frankle and Cody Blakeney and John P. Cunningham},
  journal   = {Transactions on Machine Learning Research (TMLR)},
  year      = {2024}
}

@inproceedings{houlsby2019adapter,
  title     = {Parameter-Efficient Transfer Learning for {NLP}},
  author    = {Neil Houlsby and Andrei Giurgiu and Stanislaw Jastrzebski and Bruna Morrone and Quentin de Laroussilhe and Andrea Gesmundo and Mona Attariyan and Sylvain Gelly},
  booktitle = {International Conference on Machine Learning (ICML)},
  series    = {Proceedings of Machine Learning Research},
  volume    = {97},
  pages     = {2790--2799},
  year      = {2019}
}

@inproceedings{hu2022lora,
  title     = {{LoRA}: Low-Rank Adaptation of Large Language Models},
  author    = {Edward J. Hu and Yelong Shen and Phillip Wallis and Zeyuan Allen-Zhu and Yuanzhi Li and Shean Wang and Lu Wang and Weizhu Chen},
  booktitle = {International Conference on Learning Representations (ICLR)},
  year      = {2022}
}

@inproceedings{liu2024tail,
  title     = {{TAIL}: Task-specific Adapters for Imitation Learning with Large Pretrained Models},
  author    = {Zuxin Liu and Jesse Zhang and Kavosh Asadi and Yao Liu and Ding Zhao and Shoham Sabach and Rasool Fakoor},
  booktitle = {International Conference on Learning Representations (ICLR)},
  year      = {2024}
}

@inproceedings{wolczyk2021continualworld,
  title     = {{Continual World}: A Robotic Benchmark For Continual Reinforcement Learning},
  author    = {Maciej Wo{\l}czyk and Micha{\l} Zaj{\k{a}}c and Razvan Pascanu and {\L}ukasz Kuci{\'n}ski and Piotr Mi{\l}o{\'s}},
  booktitle = {Advances in Neural Information Processing Systems (NeurIPS)},
  year      = {2021}
}

@inproceedings{liu2023libero,
  title     = {{LIBERO}: Benchmarking Knowledge Transfer for Lifelong Robot Learning},
  author    = {Bo Liu and Yifeng Zhu and Chongkai Gao and Yihao Feng and Qiang Liu and Yuke Zhu and Peter Stone},
  booktitle = {Advances in Neural Information Processing Systems (NeurIPS) Datasets and Benchmarks Track},
  year      = {2023}
}

@article{yang2026evoad,
  title     = {{EvoAD}: Self-Evolving Autonomous Driving via Latent World-Model Guided Post-hoc Reflection},
  author    = {Junjie Yang and Qiming Zhang and Yuyan Li and Shuqi Shen and Yusong Huang and Yanggang Sheng and Wenpeng Xu and Ziyi Shi and Tianshun Li and Yijie Chen and Yujia Wang and Jun Ma and Xinhu Zheng and Hongliang Lu and Hai Yang},
  journal   = {arXiv preprint arXiv:2610.23886},
  year      = {2026}
}

@article{sun2026refinedrive,
  title     = {{RefineDrive}: Reliable Failure-Guided Learning for Vision-Language-Action Driving},
  author    = {Zhe Sun and Ziyi Luo and Yehao Lu and Lei Zhou and Xi Li},
  journal   = {arXiv preprint arXiv:2609.35078},
  year      = {2026}
}

@article{dacol2026opted,
  title     = {{OPTED}: On-Policy Fine-Tuning for End-to-End Driving using a Render-Free Teacher},
  author    = {Damiano {Da Col} and Maximilian Igl and Peter Karkus and Kashyap Chitta and Boris Ivanovic and Marco Pavone and Konrad Schindler and Christos Sakaridis},
  journal   = {arXiv preprint arXiv:2609.20756},
  year      = {2026}
}

@article{zhao2026plansnotexec,
  title     = {Plans Are Not Executions: Adapting Frozen World-Model Planners to Controller-Induced Execution Shift},
  author    = {Danqi Zhao and Keyu Chen and Wenchao Sun and Minglei Huang and Jianhe Zhang and Yan Zhu and Hao Cheng and Zifan Cao and Sifa Zheng},
  journal   = {arXiv preprint arXiv:2610.22041},
  year      = {2026}
}

@article{shrivastava2026sambar,
  title     = {{SAMBAR}: Selective Anchoring via Method of Multipliers for Balanced Knowledge Acquisition and Retention in Vision-Language-Action Models},
  author    = {Aayushi Shrivastava and Xunlan Zhou and Hongrui Zhao and Ziyu Chen and Negar Mehr},
  journal   = {arXiv preprint arXiv:2609.32108},
  year      = {2026}
}

@article{khan2026cl3dod,
  title     = {{CL3DOD}: Understanding the Stability-Plasticity Dilemma in Outdoor Class-Incremental {3D} Object Detection},
  author    = {Muhammad Ahmed Ullah Khan and Khalil Ahmed and Didier Stricker and Muhammad Zeshan Afzal},
  journal   = {arXiv preprint arXiv:2611.03229},
  year      = {2026}
}

@article{kim2026odyssey,
  title     = {Odyssey: A Closed-Loop Benchmark for Long-Horizon Real-World Driving with Explicit Navigation Routes},
  author    = {Jungho Kim and Hongjae Shin and Seunghoon Yu and Heecheol Yoo and Myeongjun Kim and Jiyong Oh and Donghyuk Kwak and Seunghyeop Nam and Haesung Oh and Hyunju Kim and Hyungchan Cho and Jaehyun Park and Soo Won Seo and Jun Won Choi},
  journal   = {arXiv preprint arXiv:2610.06469},
  year      = {2026}
}

@article{lin2026roxdrive,
  title     = {{RoXDrive}: Closed-Loop Reinforcement Learning for End-to-End Autonomous Driving via Action-Faithful Rollouts},
  author    = {Hongbin Lin and Chaoda Zheng and Yiming Yang and Xiangyu Li and Shijia Chen and Jinhao Deng and Kangjie Chen and Dongbin Zhang and Jie Feng and Yu Zhang and Xianming Liu and Shuguang Cui and Boyang Wang and Zhen Li},
  journal   = {arXiv preprint arXiv:2609.36851},
  year      = {2026}
}

@article{wei2026counterplay,
  title     = {{CounterPlay}: Counterfactual Post-Training for Self-Play Driving Policies},
  author    = {Jiarong Wei and Yin Wu and Runkai He and Abhinav Valada},
  journal   = {arXiv preprint arXiv:2609.21617},
  year      = {2026}
}

@article{distelzweig2026interact,
  title     = {{INTERACT}: Interactive Planning for Autonomous Driving via Anchor-Conditioned Prediction and Trust-Region Refinement},
  author    = {Aron Distelzweig and Andreas Look and Faris Janjo{\v{s}} and Steffen Hagedorn and Luigi Palmieri and Joschka Boedecker},
  journal   = {arXiv preprint arXiv:2609.31137},
  year      = {2026}
}

@article{li2025dualls,
  title     = {Complementary Learning System Empowers Online Continual Learning of Vehicle Motion Forecasting in Smart Cities},
  author    = {Zirui Li and Yunlong Lin and Guodong Du and Xiaocong Zhao and Cheng Gong and Chen Lv and Chao Lu and Jianwei Gong},
  journal   = {arXiv preprint arXiv:2508.19597},
  year      = {2025},
  eprint    = {2508.19597},
  archiveprefix = {arXiv},
  primaryclass  = {cs.LG}
}

@inproceedings{lin2023otfcl,
  title     = {Rethinking Trajectory Prediction in Real-World Applications: An Online Task-Free Continual Learning Perspective},
  author    = {Yunlong Lin and Zirui Li and Cheng Gong and Qi Liu and Chao Lu and Jianwei Gong},
  booktitle = {IEEE 26th International Conference on Intelligent Transportation Systems (ITSC)},
  pages     = {5020--5026},
  year      = {2023},
  doi       = {10.1109/ITSC57777.2023.10421951}
}

@article{li2023cdbl,
  title     = {Continual Driver Behaviour Learning for Connected Vehicles and Intelligent Transportation Systems: Framework, Survey and Challenges},
  author    = {Zirui Li and Cheng Gong and Yunlong Lin and Guopeng Li and Xinwei Wang and Chao Lu and Miao Wang and Shanzhi Chen and Jianwei Gong},
  journal   = {Green Energy and Intelligent Transportation},
  volume    = {2},
  number    = {4},
  pages     = {100103},
  year      = {2023},
  doi       = {10.1016/j.geits.2023.100103}
}

@article{ren2026improving,
  title     = {Improving While Deployed: Learning Proactive End-to-End Driving from Imperfect Human Takeover Data},
  author    = {Yangang Ren and Di Zeng and Jun Li and Zirui Li and Chengjiang Zhang and Xiaoqing Ye and Ji Tao and Chen Lv},
  journal   = {arXiv preprint arXiv:2611.23719},
  year      = {2026},
  eprint    = {2611.23719},
  archiveprefix = {arXiv},
  primaryclass  = {cs.RO}
}
}

\ifshowappendix
  \ifshowdraftaids\IfFileExists{sec/app_pending.tex}{
}{}\fi
  \clearpage
  \setcounter{page}{1}\renewcommand{\thepage}{S\arabic{page}}
  \appendix\raggedbottom\clubpenalty=10000 \csname @clubpenalty\endcsname=10000 \widowpenalty=10000
\maketitlesupplementary
\IfFileExists{sec/app_a_formulation.tex}{
\section{Problem Formulation:\texorpdfstring{\protect\\}{ }Post-Deployment Continual Learning for Autonomous Driving}
\label{app:formulation}
\setcounter{figure}{0}\renewcommand{\thefigure}{\thesection\arabic{figure}}\renewcommand{\theHfigure}{\thesection.\arabic{figure}}
\setcounter{equation}{0}\renewcommand{\theequation}{\thesection\arabic{equation}}\renewcommand{\theHequation}{\thesection.\arabic{equation}}
\begin{figure*}[t]
  \centering
  \IfFileExists{figures/figA1_formulation.pdf}{%
    \includegraphics[width=\textwidth]{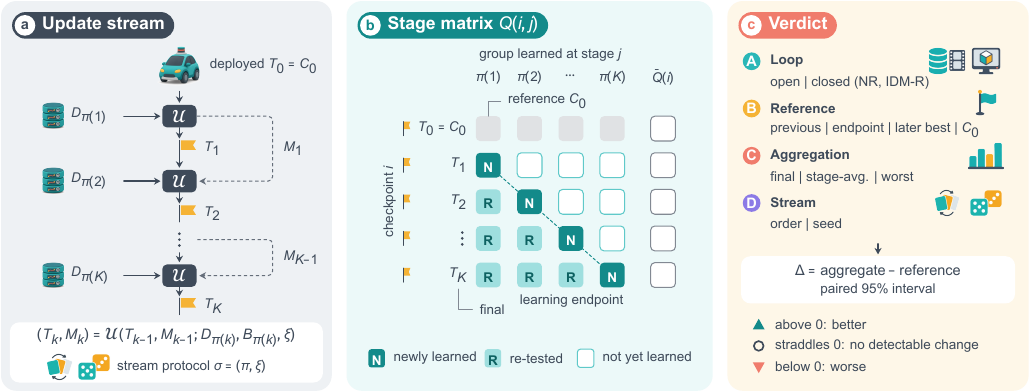}%
  }{%
    \fbox{\parbox[c][2.7in][c]{0.985\textwidth}{\centering\sffamily Formulation figure in preparation}}%
  }
  \caption{\textbf{Post-deployment continual learning, formally.}
    (a)~The deployed checkpoint $T_0=\Czero$ is updated on the groups $D_{\pi(1)},\dots,D_{\pi(K)}$ by an update regime $\mathcal{U}$ that carries a state $M_k$ (\cref{eq:update}); the stream protocol $\sigma=(\pi,\xi)$ fixes order and seed.
    (b)~Each evaluation loop yields a stage matrix $Q(i,j)$: row $T_0$ is the \Czero reference, the diagonal holds the learning endpoints (N), the cells below it re-test earlier groups (R), and row $T_K$ is the final checkpoint; $\bar Q(i)$ is the macro score of $T_i$.
    (c)~A verdict compares an aggregate with a reference through a paired 95\% interval; the four choices are the measurement axes A--D.}
  \label{fig:formulation}
\end{figure*}

\paragraph{Planner and deployed checkpoint.}
A learned planner $f_T$ maps a driving scene $x$ (tracked agents, vectorized map, route) to an ego trajectory $\hat y = f_T(x)$, where $T$ denotes a checkpoint, \ie, the planner's weights.
Before deployment, $T_0$ is trained on data $D_0$ and put into service; we call it the deployed checkpoint, $T_0 = \Czero$.
For \PlanTF, $D_0$ is our pretraining set of 300,000 samples (\cref{sec:deployment}); for \DiffP, \Czero is the official checkpoint and $D_0$ its training data.

\paragraph{Update stream.}
After deployment, newly logged expert data arrive and are grouped by scenario type into $K$ groups with training sets $D_1,\dots,D_K$ ($K=7$ here).
Each group $g$ also has a disjoint development set $S_g$ (here 640 cached samples) and closed-loop targets $P_g$ drawn from it (here 8 scenarios in the core panel and 90 in the expanded panel); a sentinel group is evaluated but never trained in the stream.
A \emph{stream protocol} $\sigma=(\pi,\xi)$ fixes the order $\pi$, a permutation of $\{1,\dots,K\}$, and the training seed $\xi$, which sets data order, dropout, diffusion noise and replay draws (here three orders, with two seeds in forward order).
An \emph{update regime} $\mathcal{U}$ carries a state $M$ (a replay memory or projection bases; $M_0=\varnothing$) and maps
\begin{equation}
  (T_k, M_k) = \mathcal{U}\bigl(T_{k-1}, M_{k-1};\, D_{\pi(k)},\, B_{\pi(k)},\, \xi\bigr)
  \label{eq:update}
\end{equation}
for $k = 1,\dots,K$, where $B_g$ is the number of optimizer updates spent on group $g$; budgets are matched across regimes (here 32,420 updates per run on the full training pools and 5,600 in the equal-data runs), except for one run of the LoRA branch that uses a larger batch (\cref{app:lora}).
Each update is an offline training cycle on logged data, not on-vehicle learning, and every $T_k$ is a candidate for release.
A joint reference instead trains $T_0$ once on $D_1\cup\dots\cup D_K$ and yields a final checkpoint without intermediate stages.
As in our runs, no regime stores data from $D_0$ or a basis computed on it (\cref{sec:regimes}).

\paragraph{Evaluation and the stage matrix.}
An evaluation loop $e$ is either open loop (single-shot prediction scored against the logged trajectory) or closed loop (a 20-s simulation in which other agents replay their logs or, for vehicles, follow the IDM).
For each loop, scoring every checkpoint on every group gives the $(K{+}1)\times K$ stage matrix
\begin{equation}
  Q_e(i,j) = \mathrm{score}_e\bigl(T_i;\, \pi(j)\bigr)
\end{equation}
for $i = 0,\dots,K$ and $j = 1,\dots,K$: the score of $T_i$ on the development set (open loop) or the closed-loop targets of the group learned at stage $j$, oriented so that higher is better (displacement errors are negated).
Row $i=0$ scores \Czero, the diagonal entries $Q(j,j)$ are the learning endpoints, and row $i=K$ scores the final checkpoint (\cref{fig:formulation}); $\bar Q(i)=\frac1K\sum_j Q(i,j)$ is the macro score of $T_i$.

\paragraph{Measurement axes.}
A verdict is a function $v(e, r, a, \sigma)$ of four choices, the axes of \cref{sec:axes}:
(A)~the loop $e$;
(B)~the reference $r$: the previous checkpoint $Q(j{-}1,j)$ (acquisition), the learning endpoint $Q(j,j)$ (backward transfer), the later best $\max_{i\ge j}Q(i,j)$ (forgetting) or the deployed checkpoint $Q(0,j)$;
(C)~the aggregation $a$ over checkpoints: final ($i=K$), stage-averaged ($\bar Q_{\mathrm{avg}}=\frac1K\sum_{k\ge1}\bar Q(k)$) or worst stage ($\bar Q_{\mathrm{worst}}=\min_{k\ge1}\bar Q(k)$);
(D)~the stream protocol $\sigma$.
For general $K$,
\begin{align}
  \BWT &= \tfrac{1}{K-1}\textstyle\sum_{j=1}^{K-1}\bigl[Q(K,j)-Q(j,j)\bigr], \nonumber\\
  \Dend &= \tfrac1K\textstyle\sum_{j=1}^{K}\bigl[Q(j,j)-Q(0,j)\bigr], \nonumber\\
  \Dfin &= \bar Q(K)-\bar Q(0) = \Dend + \tfrac{K-1}{K}\,\BWT .
  \label{eq:identityK}
\end{align}
The identity follows by adding and subtracting $Q(j,j)$ inside $\bar Q(K)-\bar Q(0)$, since the $j=K$ term of $\sum_j[Q(K,j)-Q(j,j)]$ vanishes.
It makes the reference axis concrete: a loss that is already present when a group has just been learned enters \Dend and \Dfin, but never \BWT.

\paragraph{Verdicts and the release question.}
For a contrast $\Delta$ (a difference of aggregated scores, against a reference or between two regimes) with a paired 95\% interval $I$, the verdict is better if $I$ lies above zero, worse if it lies below zero, and no detectable change otherwise; a flip is a change of verdict when one axis changes (\cref{sec:axes}).
The question a deployment team asks of a candidate $T_k$ fixes the loop to closed loop and the reference to \Czero: is $\bar Q(k)-\bar Q(0)$ better, and which closed-loop targets that \Czero passed does $T_k$ now fail?
The second part counts pass-to-fail and fail-to-pass transitions between \Czero and $T_k$, tested with McNemar's exact test.
Post-deployment continual learning seeks update regimes $\mathcal{U}$ whose candidates answer this question favorably across stream protocols $\sigma$.
A benchmark for it must state $(e, r, a, \sigma)$ for every reported verdict because, as \cref{sec:results} shows, the same checkpoints can be judged differently when one of them changes.

\paragraph{Relation to standard continual learning evaluation.}
Continual learning evaluations usually report accuracy after the last task, averaged over tasks, and forgetting against the learning endpoint or the best earlier checkpoint~\cite{lopezpaz2017gem,chaudhry2018rwalk}.
These say whether a learner kept what it learned, which a deployment team also needs, but not whether the model in service improved.
The deployment question changes the reference to \Czero and asks for closed-loop scores, summaries over the sequence of releases and a stated stream protocol; driving and robot-learning protocols report some of these (\cref{tab:positioning}).
The formulation keeps the usual quantities and makes all four choices explicit.

}{}
\IfFileExists{sec/app_c_data.tex}{
\section{Data and Closed-Loop Panels}
\label{app:data}

\makeatletter
\edef\cdCsavedtable{\the\c@table}%
\edef\cdCsavedfigure{\the\c@figure}%
\let\cdCsavedthetable\thetable
\let\cdCsavedthefigure\thefigure
\providecommand*{\theHtable}{\arabic{table}}%
\providecommand*{\theHfigure}{\arabic{figure}}%
\let\cdCsavedtheHtable\theHtable
\let\cdCsavedtheHfigure\theHfigure
\makeatother
\setcounter{table}{0}%
\setcounter{figure}{0}%
\renewcommand{\thetable}{\thesection\arabic{table}}%
\renewcommand{\thefigure}{\thesection\arabic{figure}}%
\renewcommand{\theHtable}{\thesection.\arabic{table}}%
\renewcommand{\theHfigure}{\thesection.\arabic{figure}}%

\begin{table}[t]
  \caption{\textbf{Update stream and group composition} in forward order; the reverse order runs from $k=7$ to $k=1$, and the third order visits $k=3,4,2,7,5,1,6$.
  Upd.: optimizer updates on the full training pools (equal-data runs: 10,000 samples and 800 updates per group).
  Lit.: share of samples whose raw nuPlan scenario type equals the group's historical ID, train\,/\,dev (\%).
  TVD: train--dev total-variation distance of raw types.
  --: no stage, training share or TVD, as the stream never trains the sentinel.
  Training (2,142 logs) and development (448 logs) share no sample ID or log name.}
  \label{tab:composition}
  \centering
  \footnotesize
  \setlength{\tabcolsep}{3.2pt}
  \begin{tabular}{@{}clrrrcc@{}}
    \toprule
    $k$ & Group & Train & Upd. & Dev & Lit.\ tr\,/\,dev & TVD \\
    \midrule
    1 & \gStraight{}$^{s}$   & 100,000 & 7,820 & 640 & 0.001\,/\,0.00 & 0.222 \\
    2 & \gFollow{}$^{a}$     &  58,199 & 4,560 & 640 & 0.65\,/\,2.34   & 0.266 \\
    3 & \gHighLat            &  11,282 &   900 & 640 & 99.16\,/\,99.22 & 0.001 \\
    4 & \gNearMulti{}$^{b}$  &  38,146 & 3,000 & 640 & 74.15\,/\,51.25 & 0.248 \\
    5 & \gWaitPed            &  72,811 & 5,700 & 640 & 3.69\,/\,2.66   & 0.330 \\
    6 & \gPickup{}$^{c}$     &  33,501 & 2,620 & 640 & 0.00\,/\,96.88  & 0.969 \\
    7 & \gStationary{}$^{s}$ & 100,000 & 7,820 & 640 & 4.00\,/\,20.62  & 0.299 \\
    -- & \gSentinel{}$^{s}$  & \multicolumn{2}{c}{evaluation only} & 640 & -- & -- \\
    \midrule
       & Total               & 413,939 & 32,420 & 5,120 & & \\
    \bottomrule
  \end{tabular}
  \par\smallskip
  \begin{minipage}{\linewidth}
    \scriptsize
    $^{s}$\,\PlanTF-\Czero pretraining group.
    $^{a}$\,92.8\% of training samples are raw \textit{following\_\allowbreak lane\_\allowbreak without\_\allowbreak lead}.
    $^{b}$\,Raw \textit{near\_\allowbreak high\_\allowbreak speed\_\allowbreak vehicle}: 23.2\% of training, 48.0\% of dev.
    $^{c}$\,Training is 100\% raw \textit{on\_\allowbreak carpark}: the group measures transfer.
    Historical IDs: (1)~\textit{starting\_\allowbreak straight\_\allowbreak traffic\_\allowbreak light\_\allowbreak intersection\_\allowbreak traversal},
    (2)~\textit{following\_\allowbreak lane\_\allowbreak with\_\allowbreak lead},
    (3)~\textit{high\_\allowbreak lateral\_\allowbreak acceleration},
    (4)~\textit{near\_\allowbreak multiple\_\allowbreak vehicles},
    (5)~\textit{waiting\_\allowbreak for\_\allowbreak pedestrian\_\allowbreak to\_\allowbreak cross},
    (6)~\textit{traversing\_\allowbreak pickup\_\allowbreak dropoff},
    (7)~\textit{stationary\_\allowbreak in\_\allowbreak traffic},
    sentinel~\textit{high\_\allowbreak magnitude\_\allowbreak speed}.
    Closed-loop core panel: 64 scenarios, 8 per group, from 54 logs; expanded panel: 720 scenarios, 90 per group, from 271 logs.
  \end{minipage}
\end{table}

\subsection{Update stream and splits}
\label{app:data-composition}

\begin{figure}[t]
  \centering
  \includegraphics[width=\columnwidth]{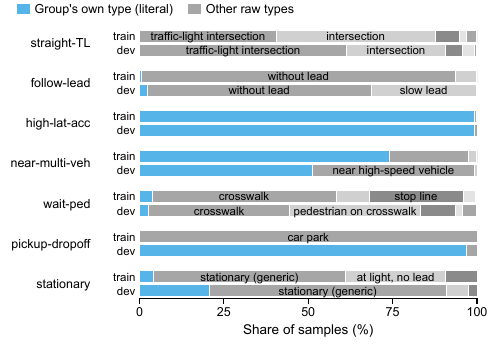}
  \caption{\textbf{Raw scenario types within each group} in training (upper bar) and development (lower bar; 640 samples per group).
    Blue: the group's own historical type; grey: other raw types (a type keeps its shade in both bars), labelled where space allows.
    \gPickup trains only on car-park scenes, and its own type occurs in no training group.}
  \label{fig:C1}
\end{figure}

\Cref{tab:composition} lists the seven groups in stream order, and \cref{fig:C1} shows the raw nuPlan scenario types they contain.

\paragraph{Groups are labels, not skills.}
The group IDs come from a historical mapping that assigns each raw scenario type of the training data to exactly one group, and in five of the seven groups under 5\% of the training samples carry the literal type, \ie the raw type named by the group ID. 
\gStraight, for example, mixes intersection traversals with and without traffic lights, and \gFollow trains mostly without a lead (92.8\% of its training samples, against 66\% of its development samples). 
Only \gHighLat is nearly pure (99.2\% literal in both splits, TVD 0.001); in the five other groups, \gPickup aside, the two splits differ by a TVD of 0.222 to 0.330. 
\gPickup is the extreme case: it trains only on car-park scenes, whereas 620 of its 640 development samples are pickup/drop-off scenes, a raw type absent from every training group (TVD 0.969). 
The group thus measures transfer from car-park driving to pickup/drop-off scenes, not in-distribution learning.
The sentinel's development samples all carry its literal type, which occurs in no training group. 
Group-level results are therefore results on these mixtures, not on single driving skills.

\paragraph{Split isolation.}
Training (413,939 samples, 2,142 logs) and development (5,120 samples, 448 logs) share no sample and no log, no sample belongs to two groups, and both planners use the same samples and group labels. 
The equal-data runs train on a fixed subset of 10,000 samples per group, the same for both planners and all regimes (\cref{app:equal-data}). 
Disjoint logs can still come from nearby locations, the same routes or close recording times.

\subsection{Closed-loop panels}
\label{app:data-panel}

\begin{table}[t]
  \caption{\textbf{Closed-loop core panel by group} (8 scenarios each).
    Logs: distinct logs; Max: most scenarios from one log; Lit.: scenarios of the group's literal type; IDM: median number of IDM vehicles at the start of \IDMR; None: scenarios without IDM vehicles.
    The last row counts distinct logs; --: not computed.}
  \label{tab:C2}
  \centering
  \footnotesize
  \setlength{\tabcolsep}{6pt}
  \begin{tabular}{@{}lrrrrr@{}}
    \toprule
    Group & Logs & Max & Lit. & IDM & None \\
    \midrule
    \gStraight   & 8 & 1 & 0 & 36.5 & 0 \\ 
    \gFollow     & 6 & 3 & 1 &  0.5 & 4 \\ 
    \gHighLat    & 8 & 1 & 8 & 15.0 & 1 \\ 
    \gNearMulti  & 7 & 2 & 5 & 41.5 & 0 \\ 
    \gWaitPed    & 8 & 1 & 0 &  6.0 & 0 \\ 
    \gPickup     & 8 & 1 & 7 & 15.0 & 0 \\ 
    \gStationary & 8 & 1 & 2 & 23.0 & 0 \\ 
    \gSentinel{} (sentinel) & 8 & 1 & 8 & 19.0 & 1 \\ 
    \midrule
    All 64       & 54 & 3 & 31 & -- & 6 \\ 
    \bottomrule
  \end{tabular}
\end{table}

\paragraph{Core panel.}
The core panel holds 64 twenty-second scenarios, 8 per group, from 54 logs (\cref{tab:C2}), drawn from the 5,120 cached development samples rather than from a separate evaluation population.
Each group contributes one anchor scenario, already used in earlier technical validation, and seven scenarios drawn at random without replacement from its other development samples; because of the anchors, the core panel is not a blind selection. 

\paragraph{Expanded panel.}
The expanded panel keeps the 64 core scenarios and adds 82 per group, drawn at random (selection seed 20261111) from the remaining development samples with 2\,s of log history and 20\,s of future log: 720 scenarios, 90 per group (630 targets and 90 sentinel scenarios), from 271 logs.
Selection used no model score, and no scenario was replaced after evaluation.
We drew the panel after the core results were known, so it is a larger development panel, not a test set; it evaluates \Czero and \Tk{7} of the eight main runs (\cref{app:expanded-endpoints}).
The composition, dependence and IDM statistics below describe the core panel.

\paragraph{Composition.}
The core-panel scenarios come from 16 vehicles and from logs dated on 9 days between \mbox{2021-06-07} and \mbox{2021-10-05}; 28 of the 64 come from logs dated \mbox{2021-07-24}. 
As in training, group labels are impure: only 31 of the 64 scenarios carry their group's literal type, none in \gStraight and \gWaitPed, and six of the eight \gFollow scenarios have no lead. 

\paragraph{Dependence.}
Eight logs contribute more than one scenario (12 same-log pairs), seven of them to two groups, and \gFollow takes 3 of its 8 scenarios from one log. 
Two same-log pairs overlap in time: one joins two \gNearMulti scenarios, the other a \gNearMulti and a sentinel scenario from the only log that the sentinel shares with the targets, so the sentinel is not fully disjoint from them. %
Scenarios are therefore not independent: the default intervals resample scenarios, and only the log-cluster bootstrap behind \rob{} resamples whole logs (\cref{sec:axes}).

\paragraph{IDM vehicles.}
Six scenarios contain no IDM vehicle: four of \gFollow, one of \gHighLat and one of the sentinel. 
\IDMR adds no vehicle interaction there, yet these scenarios count in both modes, so not every \NR--\IDMR difference reflects vehicle interaction.

\subsection{Pretraining exposure}
\label{app:data-overlap}

Both panels are subsets of the development set, so they share no sample or log with the stream's training data (\cref{sec:stream,sec:evaluation}).
For \PlanTF, the 300,000 training samples of our pretraining share no sample or log with the 5,120 development samples (448 logs), and hence none with either panel; this statement covers the pretraining's training set, not its 30,000 validation samples, which took no gradient updates. 
Separately, 23,298 of the 413,939 stream training samples (5.6\%) also occur in this pretraining set, so for \PlanTF part of the stream repeats data that \Czero was trained on; this overlap lies between training sets and does not reach evaluation. 
For \DiffP, the public training-log list of the official checkpoint shares no log with our 448 development logs; we do not have the checkpoint's per-sample training list, so this is a log-level statement. 
The two \Czero checkpoints thus differ in source data, training recipe and exposure, and their scores do not isolate architecture effects (\cref{sec:deployment}).

\setcounter{table}{\cdCsavedtable}%
\setcounter{figure}{\cdCsavedfigure}%
\let\thetable\cdCsavedthetable
\let\thefigure\cdCsavedthefigure
\let\theHtable\cdCsavedtheHtable
\let\theHfigure\cdCsavedtheHfigure
}{}
\IfFileExists{sec/app_d_statistics.tex}{
\section{Metric Definitions and Statistical Procedures}
\label{app:statistics}
\makeatletter
\edef\cdDsavedtable{\the\c@table}%
\let\cdDsavedthetable\thetable
\providecommand*{\theHtable}{\arabic{table}}%
\let\cdDsavedtheHtable\theHtable
\makeatother
\setcounter{table}{0}%
\renewcommand{\thetable}{\thesection\arabic{table}}%
\renewcommand{\theHtable}{\thesection.\arabic{table}}%

The driving metrics and their evaluation settings are defined in \cref{sec:evaluation}; notation follows \cref{sec:axes} and \cref{app:formulation}.

\begin{table*}[t]
  \centering
  \caption{\textbf{Continual learning statistics} ($K=7$; $Q$ is higher-is-better; \cref{app:stat-cl}).
    --: no reference, as the macro score is a level.
    Coverage ``all'': needs only \Czero, each group before and after its stage, and \Tk{7}, which every sequential run has; ``full'': needs the complete stage matrix.
    Scenario counts refer to the core panel; the expanded panel (90 per group) has only $k=0$ and $k=7$ of the main runs.
    Across orders: whether we compare the statistic between task orders; $^{a}$same group set in every order, but the values still depend on order (\cref{sec:order}); $^{b}$forward and reverse order only.}
  \label{tab:D2}
  \scriptsize
  \setlength{\tabcolsep}{3pt}
  \renewcommand{\arraystretch}{1.18}
  \begin{tabular}{@{}>{\raggedright\arraybackslash}p{0.14\textwidth}>{\raggedright\arraybackslash}p{0.29\textwidth}>{\raggedright\arraybackslash}p{0.10\textwidth}>{\raggedright\arraybackslash}p{0.20\textwidth}>{\raggedright\arraybackslash}p{0.13\textwidth}c@{}}
    \toprule
    Statistic & Definition & Reference & Averaging set (panel scenarios) & Coverage & Across orders \\
    \midrule
    Macro score & $\bar Q(k)=\frac17\sum_{j=1}^{7}Q(k,j)$ & -- & 7 target groups (56) & $k=0,7$: all; else full & yes \\  %
    Acquisition & $\Acq=\frac17\sum_{j}[Q(j,j)-Q(j{-}1,j)]$ & previous checkpoint & 7 groups (56) & all & yes$^{a}$ \\  %
    Backward transfer & $\BWT=\frac16\sum_{j=1}^{6}[Q(7,j)-Q(j,j)]$ & learning endpoint & groups at positions 1--6 (48) & all & no \\  %
    Common-old \BWT & $\BWTold=\frac15\sum_{j=2}^{6}[Q(7,j)-Q(j,j)]$ & learning endpoint & positions 2--6, the same 5 groups in forward and reverse order (40) & all & yes$^{b}$ \\  %
    Common-old \BWT, three orders & $\mathrm{BWT}_{\cap}=\frac14\sum_{\pi(j)\in G_{\cap}}[Q(7,j)-Q(j,j)]$ & learning endpoint & $G_{\cap}$: \gFollow, \gHighLat, \gNearMulti, \gWaitPed, old in all three orders (32) & all & yes \\  %
    Forgetting ($\ge0$) & $\Fgt=\frac16\sum_{j=1}^{6}[\max_{i\ge j}Q(i,j)-Q(7,j)]$ & later best & positions 1--6 (48) & full & no \\  %
    Endpoint gain & $\Dend=\frac17\sum_{j}[Q(j,j)-Q(0,j)]$ & \Czero & 7 groups (56) & all & yes \\  %
    Final gain & $\Dfin=\bar Q(7)-\bar Q(0)=\Dend+\frac67\BWT$ & \Czero & 7 groups (56) & all & yes \\  %
    Checkpoint gain & $\bar Q(k)-\bar Q(0)$, $0<k<7$ & \Czero & 7 groups (56) & full & yes \\  %
    Stage average, worst stage & $\bar Q_{\mathrm{avg}}=\frac17\sum_{k=1}^{7}\bar Q(k)$, $\bar Q_{\mathrm{worst}}=\min_{k\ge1}\bar Q(k)$ & level or \Czero & 7 groups (56) & full & yes \\  %
    \SR transitions & pass-to-fail and fail-to-pass counts, \Czero \vs \Tk{k} & \Czero & 56 targets & $k=7$ or endpoint: all; else full & yes \\  %
    Sentinel retention & $Q(k,\text{sentinel})-Q(0,\text{sentinel})$ & \Czero & 640 samples or 8 scenarios & $k=7$: all; else full & yes \\  %
    \bottomrule
  \end{tabular}
\end{table*}

\subsection{Continual learning statistics}
\label{app:stat-cl}

\paragraph{Stage matrix.}
$Q(i,j)$ is defined in \cref{sec:axes} for each metric and loop.
Cached entries average a group's 640 development samples, and panel entries its 8 core-panel scenarios. %
The cached stage matrices are complete in every sequential run: the core suite, the third order, the LoRA branch and the equal-data runs. %
In closed loop, each of these runs has \Czero, each group before and after its stage, and \Tk{7}; the complete matrix exists only for \NR in the eight main runs, so the closed-loop statistics that need it (coverage ``full'' in \cref{tab:D2}) are main-run statistics. 
The joint references have only \Czero and their final checkpoint, and the expanded panel only \Czero and \Tk{7} of the eight main runs, so both support final-checkpoint comparisons with \Czero but no stage statistics.
\Cref{tab:D2} lists each statistic with its reference, averaging set and coverage, and states whether we compare it across task orders.

\paragraph{Relation to standard metrics.}
\BWT is GEM's backward transfer~\cite{lopezpaz2017gem}.
RWalk's forgetting~\cite{chaudhry2018rwalk} takes its maximum over the checkpoints before the final one and can be negative; the maximum in \Fgt also includes \Tk{7}, so $\Fgt\ge0$, and $\Fgt=0$ exactly when every old group is at its post-learning best at \Tk{7}.
\Dend is the learning accuracy of MER~\cite{riemer2019mer} measured from \Czero, and \Acq measures the same endpoints from the previous checkpoint.

\paragraph{Comparisons across orders.}
We compare task orders only with statistics over a fixed group set (last column of \cref{tab:D2}).
Averaging over the groups seen so far, $A_{\mathrm{seen}}=\frac17\sum_{k=1}^{7}\frac1k\sum_{j\le k}Q(k,j)$, lets the group set change with the stage and the order: for \DiffP \AdaFT (cached ADE, lower is better), it is 5.41\,m in the main run and 2.90\,m in reverse order, whereas the fixed-set stage average is 4.78 and 4.76\,m. %
A fixed set removes the change of group set but not order dependence, as \Acq shows: on the same seven groups, the cached \Acq of \DiffP \AdaFT ($Q=-$ADE) is $-0.301$\,m in the main run and $+0.662$\,m in reverse order, mostly recovery from the stage-1 collapse (\cref{tab:E2}).
The five groups of \BWTold are old in the forward and reverse orders but not in the third order, which ends on \gPickup; across the three orders we therefore use $\mathrm{BWT}_{\cap}$, whose four groups are old in each (\cref{app:third-order-stage}).

\subsection{Verdicts, intervals and tests}
\label{app:stat-inference}

Verdicts follow \cref{sec:axes}; a paired contrast compares the same scenarios under two checkpoints or two regimes.
Intervals are percentile intervals of a task-stratified scenario bootstrap ($B=10{,}000$; figure whiskers use $B=100{,}000$) that resamples the per-scenario differences within each group (7, 6 or 5 groups of 8 scenarios for the macro, \BWT and \BWTold) and averages the group means with equal weight. %
Closed-loop evaluation repeats exactly (re-evaluating \Czero in 2,048 repeated cases changed no metric), so an interval reflects which scenarios the panel holds, not simulator noise, and describes one training run, not training variance. %
Statistics pooled over runs use a scenario-cluster bootstrap that resamples the 56 targets with their values in every pooled run. %
A contrast marked \rob{} also excludes zero under a log-cluster bootstrap, which resamples the 47 target logs without strata, and has a sign-flip $p<0.05$ (100,000 random sign patterns of the per-scenario differences; two-sided, on their mean); for regime contrasts, $p$ is Holm-adjusted within the six regime pairs of one planner, metric, statistic or summary, and run. %
Single-group cells ($n=8$) use the exact sign-flip test over all 256 sign patterns, whose two-sided $p$ cannot fall below $2/256$, and \SR transitions the exact McNemar test. %
A zero gap between two regimes counts as neither sign.
Intervals and verdicts are computed for the 24 runs of the core suite on the core panel.
A result is \emph{(descriptive)} when no interval or test supports it: cached statistics (group means only), single-group closed-loop values, counts or correlations over runs, and the results of the third order, the LoRA branch, the joint references, the equal-data runs and the expanded panel, which we report as point estimates.
\emph{In all three runs} means the same point-estimate sign in the three core-suite runs of a planner--regime pair: the main run, the second seed and the reverse-order run; such sign counts are not independent tests (\cref{sec:axes}).
All analyses are exploratory, since the development split informed design decisions and no frozen test split exists; multiplicity is adjusted only within the Holm families, and counts over many cells are descriptive.
The power figures of \cref{sec:deploy} pool the within-group variances of the main-run \Czero-to-final \CLS-\NR changes over the four regimes; per regime, the minimum detectable effect (two-sided $\alpha=0.05$, power 0.8) is 0.08 and 0.10 \CLS for \DiffP \Replay and \PEGP, and $+0.05$ would need 21 and 32 scenarios per group; both regimes barely move the planner, and a regime that changed more scenarios would have a larger MDE. %

\subsection{Robustness of regime orderings}
\label{app:stat-robust}

\Cref{sec:order} reports that only 9 of 83 tested regime orderings hold under both seed and order.
These 83 cells pair 16 orderings (9 for \DiffP, 7 for \PlanTF), each stating that regime A beats regime B on the final value or on \BWTold, with four to seven metric families each. %
For a cell, let $g_{\mathrm{main}}$, $g_{\mathrm{s2}}$ and $g_{\mathrm{rev}}$ be the gap A$-$B in $Q$ units in the main run, the second seed and the reverse-order run, and let $s$ be the larger of the two regimes' seed differences, $|Q_{\mathrm{s2}}-Q_{\mathrm{main}}|$ of the same statistic.
Our robustness criterion requires (i)~all three gaps to be positive, (ii)~each to exceed $s$, and (iii)~for panel metrics, the paired task-stratified interval of A$-$B to lie above zero in all three runs; cached cells have no interval and are judged by (i) and (ii).
Nine cells meet it.
Of the others, 43 are positive in all three runs but fail (ii) or (iii) (sign-only), 18 are positive in both forward-order seeds but not in reverse order (order-fragile), 13 are not positive in the main run, and none is positive in the main run but not in the second seed. %
The nine are \DiffP \Replay and \PEGP over \FullFT in final cached ADE and final \CLS-\NR, \DiffP \PEGP over \FullFT also in final \SR-\NR, \DiffP \Replay over \FullFT in cached-ADE \BWTold, \PlanTF \Replay over \FullFT in final \ADEs, and \PlanTF \Replay over \PEGP in final cached ADE and FDE; the four panel cells carry intervals, and the five cached cells are task means. %
Every robust gap is smaller in reverse order than in either forward-order seed; for \DiffP \Replay over \FullFT in final \CLS-\NR, the gaps are 0.291, 0.336 and 0.107 in the main run, the second seed and reverse order ($s=0.049$; interval lower bounds 0.197, 0.237 and 0.028), and 0.119 in the third order. %
Applied to the third order, which has point estimates only, criteria (i) and (ii) hold for seven of the nine cells (with $\mathrm{BWT}_{\cap}$ in place of \BWTold): all nine gaps stay positive, but the gap of \DiffP \PEGP over \FullFT falls below $s$ in final \CLS-\NR (0.082 \vs $s=0.086$) and equals it in final \SR-\NR (both 5 of 56 scenarios). 

\setcounter{table}{\cdDsavedtable}%
\let\thetable\cdDsavedthetable
\let\theHtable\cdDsavedtheHtable
}{}
\IfFileExists{sec/app_e_results.tex}{
\section{Complete Results of the Core Suite}
\label{app:results}

\setcounter{table}{0}
\setcounter{figure}{0}
\renewcommand{\thetable}{\thesection\arabic{table}}
\renewcommand{\thefigure}{\thesection\arabic{figure}}
\providecommand*{\theHtable}{}\providecommand*{\theHfigure}{}
\renewcommand{\theHtable}{\thesection.\arabic{table}}
\renewcommand{\theHfigure}{\thesection.\arabic{figure}}

This section reports the 24 runs of the core suite; \cref{app:extensions} reports the extensions.
Notation, intervals and the mark \rob{} follow \cref{sec:axes}; s1, s2 and rev denote the main run, the second seed and the reverse-order run, and values without an interval are point estimates (descriptive).

\subsection{Complete per-run results}
\label{app:complete}

\paragraph{Closed loop, referenced to \texorpdfstring{\Czero}{C0}.}
\Cref{tab:E1} gives the final score, \Dfin and \Dend of \CLS and \SR under \NR and \IDMR for all 24 runs, the 192 cells behind the invariant of \cref{sec:deploy}. 
All 48 \Dfin and all 48 \Dend point estimates of \PlanTF are negative, and the ten positive \Dfin point estimates all belong to \DiffP \Replay or \PEGP. 
Intervals below zero (verdict worse) appear in 19 of \PlanTF's and 19 of \DiffP's 48 \Dfin cells, and in 18 and 24 of their 48 \Dend cells. 
Split by group into 1,344 run$\times$group cells ($n=8$ each), only one has a positive mean with exact sign-flip $p<0.05$ (\PlanTF \PEGP, reverse order, \IDMR \CLS at the learning endpoint of \gHighLat, $+0.378$, $p=0.031$), which among this many cells is no evidence of a gain; group-level values stay descriptive. 

\begin{table*}[!tp]
  \centering
  \caption{\textbf{Closed loop referenced to \Czero, all 24 runs.}
    Final score at \Tk{7}, \Dfin (\Tk{7} minus \Czero) and \Dend (learning endpoint minus \Czero) of \CLS (a) and \SR (b) under \NR and \IDMR, 56 targets.
    For \SR, Pass counts passing targets at \Tk{7}, L/G the targets lost and gained against \Czero at \Tk{7}, and \Dfin, \Dend are fractions of 56.
    Paired, task-stratified scenario-bootstrap 95\% intervals; \rob{} (main run only) as in \cref{sec:axes}.
    s1/s2/rev: main run, second seed, reverse order; --: not applicable (\Czero is the reference).}
  \label{tab:E1}
  \scriptsize
  \setlength{\tabcolsep}{2pt}
  \begin{tabular}{@{}ll ccc ccc@{}}
    \multicolumn{8}{@{}l}{\textbf{(a)} \CLS} \\
    \toprule
    & & \multicolumn{3}{c}{\NR} & \multicolumn{3}{c}{\IDMR} \\
    \cmidrule(lr){3-5}\cmidrule(l){6-8}
    Regime & Run & Final & \Dfin [95\% CI] & \Dend [95\% CI] & Final & \Dfin [95\% CI] & \Dend [95\% CI] \\
    \midrule
    \multicolumn{8}{@{}l}{\textit{\PlanTF}} \\
    \Czero & & $0.528$ & -- & -- & $0.557$ & -- & -- \\ 
    \FullFT & s1 & $0.258$ & $-0.270$ \ci{-0.395}{-0.149}\rob & $-0.234$ \ci{-0.365}{-0.110}\rob & $0.310$ & $-0.247$ \ci{-0.379}{-0.114}\rob & $-0.207$ \ci{-0.339}{-0.078}\rob \\ 
     & s2 & $0.289$ & $-0.239$ \ci{-0.358}{-0.121} & $-0.168$ \ci{-0.311}{-0.028} & $0.271$ & $-0.287$ \ci{-0.396}{-0.178} & $-0.116$ \ci{-0.241}{+0.008} \\ 
     & rev & $0.364$ & $-0.164$ \ci{-0.271}{-0.059} & $-0.197$ \ci{-0.322}{-0.074} & $0.506$ & $-0.052$ \ci{-0.199}{+0.096} & $-0.156$ \ci{-0.298}{-0.013} \\ 
    \addlinespace[1pt]
    \AdaFT & s1 & $0.378$ & $-0.150$ \ci{-0.263}{-0.039}\rob & $-0.172$ \ci{-0.300}{-0.044}\rob & $0.501$ & $-0.057$ \ci{-0.157}{+0.049} & $-0.135$ \ci{-0.251}{-0.014}\rob \\ 
     & s2 & $0.321$ & $-0.207$ \ci{-0.340}{-0.076} & $-0.174$ \ci{-0.302}{-0.053} & $0.428$ & $-0.129$ \ci{-0.247}{-0.015} & $-0.189$ \ci{-0.316}{-0.057} \\ 
     & rev & $0.454$ & $-0.074$ \ci{-0.199}{+0.049} & $-0.125$ \ci{-0.277}{+0.029} & $0.471$ & $-0.086$ \ci{-0.214}{+0.039} & $-0.048$ \ci{-0.183}{+0.088} \\ 
    \addlinespace[1pt]
    \Replay & s1 & $0.454$ & $-0.074$ \ci{-0.182}{+0.036} & $-0.098$ \ci{-0.222}{+0.027} & $0.549$ & $-0.009$ \ci{-0.126}{+0.105} & $-0.093$ \ci{-0.200}{+0.011} \\ 
     & s2 & $0.428$ & $-0.100$ \ci{-0.206}{+0.006} & $-0.113$ \ci{-0.241}{+0.011} & $0.524$ & $-0.033$ \ci{-0.164}{+0.093} & $-0.104$ \ci{-0.219}{+0.009} \\ 
     & rev & $0.482$ & $-0.046$ \ci{-0.162}{+0.068} & $-0.057$ \ci{-0.195}{+0.077} & $0.486$ & $-0.071$ \ci{-0.186}{+0.042} & $-0.087$ \ci{-0.230}{+0.057} \\ 
    \addlinespace[1pt]
    \PEGP & s1 & $0.406$ & $-0.122$ \ci{-0.216}{-0.035}\rob & $-0.086$ \ci{-0.224}{+0.050} & $0.508$ & $-0.050$ \ci{-0.146}{+0.051} & $-0.060$ \ci{-0.182}{+0.062} \\ 
     & s2 & $0.367$ & $-0.161$ \ci{-0.286}{-0.034} & $-0.102$ \ci{-0.254}{+0.048} & $0.493$ & $-0.065$ \ci{-0.189}{+0.060} & $-0.087$ \ci{-0.208}{+0.036} \\ 
     & rev & $0.510$ & $-0.018$ \ci{-0.120}{+0.083} & $-0.031$ \ci{-0.183}{+0.119} & $0.442$ & $-0.115$ \ci{-0.225}{-0.008} & $-0.114$ \ci{-0.249}{+0.022} \\ 
    \midrule
    \multicolumn{8}{@{}l}{\textit{\DiffP}} \\
    \Czero & & $0.823$ & -- & -- & $0.829$ & -- & -- \\ 
    \FullFT & s1 & $0.538$ & $-0.286$ \ci{-0.395}{-0.179}\rob & $-0.092$ \ci{-0.163}{-0.027}\rob & $0.438$ & $-0.390$ \ci{-0.490}{-0.289}\rob & $-0.116$ \ci{-0.214}{-0.020}\rob \\ 
     & s2 & $0.488$ & $-0.335$ \ci{-0.447}{-0.222} & $-0.093$ \ci{-0.163}{-0.029} & $0.473$ & $-0.356$ \ci{-0.457}{-0.254} & $-0.097$ \ci{-0.186}{-0.005} \\ 
     & rev & $0.679$ & $-0.144$ \ci{-0.235}{-0.064} & $-0.141$ \ci{-0.233}{-0.056} & $0.770$ & $-0.059$ \ci{-0.141}{+0.021} & $-0.130$ \ci{-0.217}{-0.049} \\ 
    \addlinespace[1pt]
    \AdaFT & s1 & $0.318$ & $-0.505$ \ci{-0.623}{-0.378}\rob & $-0.062$ \ci{-0.134}{+0.013} & $0.308$ & $-0.521$ \ci{-0.629}{-0.409}\rob & $-0.101$ \ci{-0.169}{-0.036}\rob \\ 
     & s2 & $0.364$ & $-0.459$ \ci{-0.583}{-0.333} & $-0.088$ \ci{-0.170}{-0.006} & $0.321$ & $-0.508$ \ci{-0.619}{-0.389} & $-0.104$ \ci{-0.172}{-0.037} \\ 
     & rev & $0.776$ & $-0.048$ \ci{-0.112}{+0.013} & $-0.106$ \ci{-0.196}{-0.020} & $0.809$ & $-0.020$ \ci{-0.079}{+0.036} & $-0.121$ \ci{-0.199}{-0.047} \\ 
    \addlinespace[1pt]
    \Replay & s1 & $0.828$ & $+0.005$ \ci{-0.042}{+0.062} & $+0.008$ \ci{-0.036}{+0.062} & $0.784$ & $-0.044$ \ci{-0.111}{+0.019} & $-0.055$ \ci{-0.115}{-0.003} \\ 
     & s2 & $0.824$ & $+0.001$ \ci{-0.043}{+0.057} & $-0.019$ \ci{-0.083}{+0.047} & $0.801$ & $-0.027$ \ci{-0.085}{+0.030} & $-0.050$ \ci{-0.100}{-0.006} \\ 
     & rev & $0.786$ & $-0.037$ \ci{-0.109}{+0.036} & $-0.096$ \ci{-0.179}{-0.012} & $0.797$ & $-0.031$ \ci{-0.090}{+0.023} & $-0.122$ \ci{-0.197}{-0.054} \\ 
    \addlinespace[1pt]
    \PEGP & s1 & $0.832$ & $+0.009$ \ci{-0.054}{+0.074} & $+0.003$ \ci{-0.032}{+0.045} & $0.798$ & $-0.030$ \ci{-0.100}{+0.036} & $-0.032$ \ci{-0.099}{+0.032} \\ 
     & s2 & $0.745$ & $-0.078$ \ci{-0.137}{-0.028} & $+0.002$ \ci{-0.032}{+0.045} & $0.775$ & $-0.054$ \ci{-0.128}{+0.021} & $-0.022$ \ci{-0.079}{+0.034} \\ 
     & rev & $0.821$ & $-0.002$ \ci{-0.043}{+0.045} & $-0.093$ \ci{-0.167}{-0.021} & $0.839$ & $+0.010$ \ci{-0.020}{+0.050} & $-0.078$ \ci{-0.140}{-0.014} \\ 
    \bottomrule
  \end{tabular}

  \vspace{6pt}
  \begin{tabular}{@{}ll cccc cccc@{}}
    \multicolumn{10}{@{}l}{\textbf{(b)} \SR} \\
    \toprule
    & & \multicolumn{4}{c}{\NR} & \multicolumn{4}{c}{\IDMR} \\
    \cmidrule(lr){3-6}\cmidrule(l){7-10}
    Regime & Run & Pass & L/G & \Dfin [95\% CI] & \Dend [95\% CI] & Pass & L/G & \Dfin [95\% CI] & \Dend [95\% CI] \\
    \midrule
    \multicolumn{10}{@{}l}{\textit{\PlanTF}} \\
    \Czero & & 32 & -- & -- & -- & 34 & -- & -- & -- \\ 
    \FullFT & s1 & 17 & 19/4 & $-0.268$ \ci{-0.411}{-0.125}\rob & $-0.250$ \ci{-0.393}{-0.107}\rob & 20 & 19/5 & $-0.250$ \ci{-0.411}{-0.089}\rob & $-0.232$ \ci{-0.375}{-0.089}\rob \\ 
     & s2 & 19 & 17/4 & $-0.232$ \ci{-0.375}{-0.089} & $-0.179$ \ci{-0.321}{-0.036} & 17 & 18/1 & $-0.304$ \ci{-0.429}{-0.179} & $-0.125$ \ci{-0.268}{+0.018} \\ 
     & rev & 22 & 12/2 & $-0.179$ \ci{-0.286}{-0.071} & $-0.214$ \ci{-0.339}{-0.089} & 31 & 13/10 & $-0.054$ \ci{-0.214}{+0.107} & $-0.179$ \ci{-0.321}{-0.018} \\ 
    \addlinespace[1pt]
    \AdaFT & s1 & 24 & 12/4 & $-0.143$ \ci{-0.268}{-0.018} & $-0.179$ \ci{-0.321}{-0.036} & 33 & 7/6 & $-0.018$ \ci{-0.143}{+0.107} & $-0.161$ \ci{-0.286}{-0.036}\rob \\ 
     & s2 & 20 & 17/5 & $-0.214$ \ci{-0.357}{-0.071} & $-0.196$ \ci{-0.321}{-0.071} & 28 & 11/5 & $-0.107$ \ci{-0.250}{+0.018} & $-0.196$ \ci{-0.339}{-0.054} \\ 
     & rev & 27 & 10/5 & $-0.089$ \ci{-0.214}{+0.036} & $-0.125$ \ci{-0.304}{+0.054} & 29 & 10/5 & $-0.089$ \ci{-0.232}{+0.036} & $-0.054$ \ci{-0.214}{+0.107} \\ 
    \addlinespace[1pt]
    \Replay & s1 & 28 & 9/5 & $-0.071$ \ci{-0.196}{+0.054} & $-0.089$ \ci{-0.232}{+0.054} & 33 & 7/6 & $-0.018$ \ci{-0.143}{+0.107} & $-0.089$ \ci{-0.196}{+0.018} \\ 
     & s2 & 26 & 10/4 & $-0.107$ \ci{-0.232}{+0.018} & $-0.125$ \ci{-0.268}{+0.018} & 32 & 9/7 & $-0.036$ \ci{-0.179}{+0.089} & $-0.107$ \ci{-0.232}{+0.018} \\ 
     & rev & 31 & 9/8 & $-0.018$ \ci{-0.161}{+0.125} & $-0.036$ \ci{-0.196}{+0.125} & 30 & 10/6 & $-0.071$ \ci{-0.196}{+0.054} & $-0.089$ \ci{-0.250}{+0.071} \\ 
    \addlinespace[1pt]
    \PEGP & s1 & 26 & 9/3 & $-0.107$ \ci{-0.214}{0.000} & $-0.089$ \ci{-0.232}{+0.054} & 32 & 6/4 & $-0.036$ \ci{-0.143}{+0.071} & $-0.071$ \ci{-0.196}{+0.054} \\ 
     & s2 & 23 & 15/6 & $-0.161$ \ci{-0.304}{-0.018} & $-0.107$ \ci{-0.268}{+0.054} & 31 & 10/7 & $-0.054$ \ci{-0.196}{+0.089} & $-0.089$ \ci{-0.214}{+0.036} \\ 
     & rev & 31 & 7/6 & $-0.018$ \ci{-0.143}{+0.107} & $-0.036$ \ci{-0.196}{+0.125} & 27 & 10/3 & $-0.125$ \ci{-0.250}{0.000} & $-0.107$ \ci{-0.250}{+0.036} \\ 
    \midrule
    \multicolumn{10}{@{}l}{\textit{\DiffP}} \\
    \Czero & & 48 & -- & -- & -- & 48 & -- & -- & -- \\ 
    \FullFT & s1 & 37 & 14/3 & $-0.196$ \ci{-0.321}{-0.071}\rob & $-0.071$ \ci{-0.161}{0.000} & 29 & 22/3 & $-0.339$ \ci{-0.464}{-0.214}\rob & $-0.107$ \ci{-0.214}{0.000} \\ 
     & s2 & 33 & 18/3 & $-0.268$ \ci{-0.411}{-0.125} & $-0.071$ \ci{-0.161}{0.000} & 32 & 19/3 & $-0.286$ \ci{-0.411}{-0.161} & $-0.089$ \ci{-0.196}{+0.018} \\ 
     & rev & 41 & 8/1 & $-0.125$ \ci{-0.232}{-0.036} & $-0.125$ \ci{-0.232}{-0.036} & 47 & 5/4 & $-0.018$ \ci{-0.107}{+0.071} & $-0.107$ \ci{-0.214}{-0.018} \\ 
    \addlinespace[1pt]
    \AdaFT & s1 & 22 & 28/2 & $-0.464$ \ci{-0.607}{-0.321}\rob & $-0.054$ \ci{-0.143}{+0.018} & 20 & 30/2 & $-0.500$ \ci{-0.625}{-0.375}\rob & $-0.089$ \ci{-0.161}{-0.018} \\ 
     & s2 & 25 & 25/2 & $-0.411$ \ci{-0.554}{-0.268} & $-0.089$ \ci{-0.179}{0.000} & 21 & 29/2 & $-0.482$ \ci{-0.607}{-0.357} & $-0.089$ \ci{-0.161}{-0.018} \\ 
     & rev & 45 & 4/1 & $-0.054$ \ci{-0.125}{+0.018} & $-0.107$ \ci{-0.196}{-0.018} & 48 & 2/2 & $0.000$ \ci{-0.071}{+0.054} & $-0.107$ \ci{-0.196}{-0.018} \\ 
    \addlinespace[1pt]
    \Replay & s1 & 49 & 1/2 & $+0.018$ \ci{-0.036}{+0.071} & $+0.018$ \ci{-0.036}{+0.071} & 48 & 3/3 & $0.000$ \ci{-0.071}{+0.071} & $-0.036$ \ci{-0.107}{+0.018} \\ 
     & s2 & 50 & 0/2 & $+0.036$ \ci{0.000}{+0.089} & $-0.018$ \ci{-0.089}{+0.054} & 49 & 2/3 & $+0.018$ \ci{-0.054}{+0.089} & $-0.018$ \ci{-0.071}{+0.036} \\ 
     & rev & 46 & 4/2 & $-0.036$ \ci{-0.125}{+0.036} & $-0.089$ \ci{-0.179}{0.000} & 48 & 2/2 & $0.000$ \ci{-0.054}{+0.054} & $-0.107$ \ci{-0.196}{-0.018} \\ 
    \addlinespace[1pt]
    \PEGP & s1 & 50 & 1/3 & $+0.036$ \ci{-0.036}{+0.107} & $0.000$ \ci{-0.054}{+0.054} & 48 & 3/3 & $0.000$ \ci{-0.071}{+0.071} & $0.000$ \ci{-0.071}{+0.071} \\ 
     & s2 & 45 & 3/0 & $-0.054$ \ci{-0.107}{0.000} & $0.000$ \ci{-0.054}{+0.054} & 47 & 4/3 & $-0.018$ \ci{-0.107}{+0.071} & $+0.018$ \ci{-0.054}{+0.089} \\ 
     & rev & 49 & 1/2 & $+0.018$ \ci{-0.036}{+0.071} & $-0.071$ \ci{-0.143}{+0.018} & 50 & 0/2 & $+0.036$ \ci{0.000}{+0.089} & $-0.054$ \ci{-0.125}{+0.018} \\ 
    \bottomrule
  \end{tabular}
\end{table*}

\paragraph{Closed loop, referenced to stages.}
\Cref{tab:E1b} gives \Acq, \BWT and \BWTold of \CLS and \SR under both modes and the forward transfer FWT of \CLS.
Closed-loop \Fgt needs full stage matrices and is not tabulated (\cref{sec:evaluation}).
\CLS forward transfer is negative in all 24 runs under both \NR and \IDMR, with intervals below zero in 12 runs under \NR and 16 under \IDMR and above zero in none; \SR forward transfer is negative in 23 (\NR) and 22 (\IDMR) runs. 
Updates thus lower closed-loop scores on groups not yet trained, and learning a group mostly does not bring its score back to the \Czero level: \CLS \Dend is negative in 21 of 24 runs under \NR and in all 24 under \IDMR. 
Cached ADE registers this in every run only for \DiffP: its forward transfer is negative in all 12 \DiffP runs ($-1.984$ to $-0.637$\,m) but non-negative in 6 of 12 \PlanTF runs ($-0.648$ to $+0.281$\,m; descriptive). 

\begin{table*}[!t]
  \centering
  \caption{\textbf{Closed loop referenced to stages, all 24 runs.}
    \Acq (against the previous checkpoint; 7 groups), \BWT (final against the learning endpoint; 6 old groups) and \BWTold (5 groups old in both orders) of \CLS and \SR under \NR and \IDMR, and the forward transfer of \CLS, $\mathrm{FWT}=\frac16\sum_{j=2}^{7}[Q(j{-}1,j)-Q(0,j)]$ (each group just before it is learned, against \Czero; 6 groups).
    Point estimates over 56, 48, 40 and 48 targets (descriptive); \SR in fractions.}
  \label{tab:E1b}
  \scriptsize
  \setlength{\tabcolsep}{2pt}
  \begin{tabular}{@{}ll ccc ccc ccc ccc cc@{}}
    \toprule
    & & \multicolumn{3}{c}{\CLS-\NR} & \multicolumn{3}{c}{\SR-\NR} & \multicolumn{3}{c}{\CLS-\IDMR} & \multicolumn{3}{c}{\SR-\IDMR} & \multicolumn{2}{c}{FWT, \CLS} \\
    \cmidrule(lr){3-5}\cmidrule(lr){6-8}\cmidrule(lr){9-11}\cmidrule(lr){12-14}\cmidrule(l){15-16}
    Regime & Run & \Acq & \BWT & \BWTold & \Acq & \BWT & \BWTold & \Acq & \BWT & \BWTold & \Acq & \BWT & \BWTold & \NR & \IDMR \\
    \midrule
    \multicolumn{16}{@{}l}{\textit{\PlanTF}} \\
    \FullFT & s1 & $+0.075$ & $-0.042$ & $+0.007$ & $+0.089$ & $-0.021$ & $+0.025$ & $+0.024$ & $-0.047$ & $+0.069$ & $+0.018$ & $-0.021$ & $+0.100$ & $-0.360$ & $-0.270$ \\ 
     & s2 & $+0.081$ & $-0.083$ & $-0.025$ & $+0.107$ & $-0.062$ & $0.000$ & $+0.138$ & $-0.199$ & $-0.116$ & $+0.161$ & $-0.208$ & $-0.125$ & $-0.290$ & $-0.296$ \\ 
     & rev & $-0.039$ & $+0.039$ & $+0.025$ & $-0.036$ & $+0.042$ & $+0.025$ & $-0.089$ & $+0.122$ & $+0.124$ & $-0.089$ & $+0.146$ & $+0.150$ & $-0.185$ & $-0.079$ \\ 
    \addlinespace[1pt]
    \AdaFT & s1 & $-0.117$ & $+0.025$ & $+0.041$ & $-0.107$ & $+0.042$ & $+0.050$ & $-0.013$ & $+0.091$ & $+0.126$ & $-0.018$ & $+0.167$ & $+0.200$ & $-0.064$ & $-0.142$ \\ 
     & s2 & $-0.139$ & $-0.038$ & $+0.031$ & $-0.143$ & $-0.021$ & $+0.050$ & $-0.049$ & $+0.069$ & $+0.080$ & $-0.054$ & $+0.104$ & $+0.100$ & $-0.041$ & $-0.163$ \\ 
     & rev & $-0.112$ & $+0.059$ & $+0.014$ & $-0.125$ & $+0.042$ & $0.000$ & $-0.028$ & $-0.044$ & $-0.093$ & $-0.054$ & $-0.042$ & $-0.075$ & $-0.015$ & $-0.024$ \\ 
    \addlinespace[1pt]
    \Replay & s1 & $-0.041$ & $+0.028$ & $+0.040$ & $-0.036$ & $+0.021$ & $+0.025$ & $+0.043$ & $+0.098$ & $+0.073$ & $+0.054$ & $+0.083$ & $+0.050$ & $-0.067$ & $-0.159$ \\ 
     & s2 & $-0.023$ & $+0.015$ & $+0.052$ & $-0.036$ & $+0.021$ & $+0.050$ & $+0.023$ & $+0.082$ & $+0.032$ & $+0.018$ & $+0.083$ & $+0.025$ & $-0.105$ & $-0.148$ \\ 
     & rev & $-0.003$ & $+0.013$ & $+0.007$ & $0.000$ & $+0.021$ & $+0.025$ & $-0.031$ & $+0.019$ & $+0.032$ & $-0.054$ & $+0.021$ & $+0.050$ & $-0.063$ & $-0.065$ \\ 
    \addlinespace[1pt]
    \PEGP & s1 & $+0.022$ & $-0.042$ & $-0.013$ & $+0.018$ & $-0.021$ & $0.000$ & $+0.051$ & $+0.012$ & $+0.052$ & $+0.036$ & $+0.042$ & $+0.075$ & $-0.126$ & $-0.129$ \\ 
     & s2 & $+0.083$ & $-0.068$ & $-0.023$ & $+0.107$ & $-0.062$ & $-0.025$ & $+0.024$ & $+0.026$ & $+0.022$ & $0.000$ & $+0.042$ & $+0.025$ & $-0.216$ & $-0.129$ \\ 
     & rev & $-0.002$ & $+0.015$ & $-0.005$ & $-0.018$ & $+0.021$ & $0.000$ & $-0.101$ & $-0.001$ & $-0.023$ & $-0.107$ & $-0.021$ & $-0.025$ & $-0.034$ & $-0.015$ \\ 
    \midrule
    \multicolumn{16}{@{}l}{\textit{\DiffP}} \\
    \FullFT & s1 & $+0.029$ & $-0.226$ & $-0.165$ & $+0.054$ & $-0.146$ & $-0.100$ & $+0.012$ & $-0.320$ & $-0.276$ & $0.000$ & $-0.271$ & $-0.225$ & $-0.141$ & $-0.149$ \\ 
     & s2 & $+0.029$ & $-0.283$ & $-0.204$ & $+0.054$ & $-0.229$ & $-0.150$ & $+0.015$ & $-0.302$ & $-0.245$ & $0.000$ & $-0.229$ & $-0.175$ & $-0.142$ & $-0.130$ \\ 
     & rev & $-0.038$ & $-0.004$ & $-0.058$ & $-0.036$ & $0.000$ & $-0.050$ & $+0.013$ & $+0.083$ & $+0.048$ & $+0.018$ & $+0.104$ & $+0.075$ & $-0.120$ & $-0.167$ \\ 
    \addlinespace[1pt]
    \AdaFT & s1 & $+0.007$ & $-0.517$ & $-0.458$ & $+0.018$ & $-0.479$ & $-0.425$ & $-0.017$ & $-0.489$ & $-0.420$ & $-0.018$ & $-0.479$ & $-0.400$ & $-0.080$ & $-0.099$ \\ 
     & s2 & $-0.052$ & $-0.433$ & $-0.399$ & $-0.054$ & $-0.375$ & $-0.350$ & $-0.035$ & $-0.471$ & $-0.415$ & $-0.036$ & $-0.458$ & $-0.400$ & $-0.042$ & $-0.081$ \\ 
     & rev & $+0.003$ & $+0.068$ & $-0.017$ & $-0.018$ & $+0.062$ & $-0.025$ & $+0.015$ & $+0.118$ & $+0.073$ & $+0.018$ & $+0.125$ & $+0.075$ & $-0.127$ & $-0.159$ \\ 
    \addlinespace[1pt]
    \Replay & s1 & $+0.047$ & $-0.004$ & $+0.000$ & $+0.054$ & $0.000$ & $0.000$ & $+0.022$ & $+0.012$ & $+0.018$ & $+0.018$ & $+0.042$ & $+0.050$ & $-0.045$ & $-0.090$ \\ 
     & s2 & $+0.001$ & $+0.023$ & $+0.040$ & $0.000$ & $+0.062$ & $+0.075$ & $+0.011$ & $+0.027$ & $+0.036$ & $+0.018$ & $+0.042$ & $+0.050$ & $-0.023$ & $-0.072$ \\ 
     & rev & $-0.010$ & $+0.068$ & $-0.016$ & $-0.036$ & $+0.062$ & $-0.025$ & $-0.009$ & $+0.106$ & $+0.038$ & $-0.018$ & $+0.125$ & $+0.050$ & $-0.100$ & $-0.131$ \\ 
    \addlinespace[1pt]
    \PEGP & s1 & $+0.042$ & $+0.007$ & $+0.013$ & $+0.036$ & $+0.042$ & $+0.050$ & $+0.027$ & $+0.001$ & $+0.003$ & $+0.036$ & $0.000$ & $0.000$ & $-0.046$ & $-0.069$ \\ 
     & s2 & $+0.022$ & $-0.094$ & $-0.108$ & $+0.018$ & $-0.062$ & $-0.075$ & $+0.023$ & $-0.037$ & $-0.043$ & $+0.036$ & $-0.042$ & $-0.050$ & $-0.023$ & $-0.052$ \\ 
     & rev & $-0.018$ & $+0.107$ & $+0.030$ & $-0.036$ & $+0.104$ & $+0.025$ & $-0.026$ & $+0.103$ & $+0.031$ & $-0.036$ & $+0.104$ & $+0.025$ & $-0.088$ & $-0.061$ \\ 
    \bottomrule
  \end{tabular}
\end{table*}

\paragraph{Open loop.}
\Cref{tab:E2} gives the cached ADE statistics of all runs (a) and the open-loop scores of the closed-loop panel against \Czero (b).
For \DiffP, cached \Dfin is negative in all 12 runs, and no group gains over \Czero in cached ADE or FDE at the final checkpoint or the learning endpoint (0 of 336 group cells; descriptive). 
Of \PlanTF's 72 open-loop panel intervals (\Dfin and \Dend of \ADEs, \FDEs and \PDMS in 12 runs), 15 lie above zero (5 \Dfin, 10 \Dend), all on displacement error; \DiffP has none of 72. 

\begin{table*}[!tp]
  \centering
  \caption{\textbf{Open loop, all 24 runs.}
    (a)~Cached ADE, 7-group macro over 640 development samples per group: final, stage-averaged (mean over \Tk{1}--\Tk{7}) and worst-stage error (stage in parentheses) and forgetting \Fgt, in m ($\downarrow$); \Dfin, \Dend, \Acq, \BWT and \BWTold in $Q=-$ADE, m ($\uparrow$). Group means only, hence no intervals; --: not applicable (\Czero is the reference).
    (b)~Open-loop scores of the closed-loop panel (\NR, 56 targets): \Dfin and \Dend of \ADEs ($Q$, m) and \PDMS ($\uparrow$) with 95\% intervals as in \cref{tab:E1}; \Czero rows give the levels (\ADEs as error, m).
    \FDEs and cached FDE will be in the released per-scenario results and stage matrices.}
  \label{tab:E2}
  \scriptsize
  \setlength{\tabcolsep}{2.5pt}
  \begin{tabular}{@{}ll cccc ccccc@{}}
    \multicolumn{11}{@{}l}{\textbf{(a)} Cached ADE} \\
    \toprule
    & & \multicolumn{4}{c}{Error (m) $\downarrow$} & \multicolumn{5}{c}{Change in $Q$ (m) $\uparrow$} \\
    \cmidrule(lr){3-6}\cmidrule(l){7-11}
    Regime & Run & Final & Stage-avg. & Worst (stage) & \Fgt & \Dfin & \Dend & \Acq & \BWT & \BWTold \\
    \midrule
    \multicolumn{11}{@{}l}{\textit{\PlanTF}} \\
    \Czero & & $3.561$ & -- & -- & -- & -- & -- & -- & -- & -- \\ 
    \FullFT & s1 & $5.489$ & $4.557$ & $5.489$ (7) & $2.279$ & $-1.928$ & $-0.143$ & $+0.413$ & $-2.083$ & $-1.919$ \\ 
     & s2 & $4.780$ & $4.503$ & $6.340$ (4) & $1.557$ & $-1.219$ & $-0.345$ & $+0.207$ & $-1.019$ & $-0.878$ \\ 
     & rev & $3.367$ & $4.496$ & $5.613$ (4) & $0.228$ & $+0.194$ & $+0.015$ & $+0.435$ & $+0.208$ & $+0.485$ \\ 
    \addlinespace[1pt]
    \AdaFT & s1 & $4.842$ & $3.829$ & $4.842$ (7) & $1.906$ & $-1.280$ & $+0.229$ & $+0.090$ & $-1.761$ & $-1.698$ \\ 
     & s2 & $4.432$ & $3.747$ & $4.432$ (7) & $1.445$ & $-0.870$ & $+0.257$ & $+0.104$ & $-1.316$ & $-1.304$ \\ 
     & rev & $3.248$ & $3.788$ & $4.355$ (2) & $0.251$ & $+0.313$ & $+0.278$ & $+0.553$ & $+0.041$ & $+0.082$ \\ 
    \addlinespace[1pt]
    \Replay & s1 & $3.034$ & $3.352$ & $3.558$ (3) & $0.052$ & $+0.527$ & $+0.513$ & $+0.320$ & $+0.017$ & $+0.047$ \\ 
     & s2 & $3.031$ & $3.338$ & $3.508$ (6) & $0.034$ & $+0.530$ & $+0.488$ & $+0.248$ & $+0.048$ & $+0.079$ \\ 
     & rev & $3.063$ & $3.486$ & $4.053$ (2) & $0.103$ & $+0.498$ & $+0.337$ & $+0.451$ & $+0.188$ & $+0.261$ \\ 
    \addlinespace[1pt]
    \PEGP & s1 & $4.208$ & $3.662$ & $4.208$ (7) & $1.207$ & $-0.647$ & $+0.352$ & $+0.197$ & $-1.165$ & $-1.179$ \\ 
     & s2 & $4.093$ & $3.646$ & $4.093$ (7) & $1.080$ & $-0.532$ & $+0.314$ & $+0.151$ & $-0.987$ & $-1.009$ \\ 
     & rev & $3.364$ & $3.825$ & $4.218$ (2) & $0.178$ & $+0.198$ & $+0.141$ & $+0.428$ & $+0.067$ & $+0.149$ \\ 
    \midrule
    \multicolumn{11}{@{}l}{\textit{\DiffP}} \\
    \Czero & & $2.978$ & -- & -- & -- & -- & -- & -- & -- & -- \\ 
    \FullFT & s1 & $8.372$ & $4.579$ & $8.372$ (7) & $5.324$ & $-5.394$ & $-0.841$ & $-0.091$ & $-5.312$ & $-5.088$ \\ 
     & s2 & $8.360$ & $4.610$ & $8.360$ (7) & $5.303$ & $-5.383$ & $-0.844$ & $-0.077$ & $-5.294$ & $-5.110$ \\ 
     & rev & $4.086$ & $4.441$ & $7.007$ (1) & $0.475$ & $-1.108$ & $-0.774$ & $+0.532$ & $-0.390$ & $-0.489$ \\ 
    \addlinespace[1pt]
    \AdaFT & s1 & $11.165$ & $4.777$ & $11.165$ (7) & $8.758$ & $-8.187$ & $-0.847$ & $-0.301$ & $-8.563$ & $-7.890$ \\ 
     & s2 & $10.776$ & $4.714$ & $10.776$ (7) & $8.326$ & $-7.798$ & $-0.833$ & $-0.283$ & $-8.126$ & $-7.506$ \\ 
     & rev & $3.762$ & $4.757$ & $11.137$ (1) & $0.336$ & $-0.784$ & $-0.851$ & $+0.662$ & $+0.077$ & $-0.103$ \\ 
    \addlinespace[1pt]
    \Replay & s1 & $3.842$ & $3.715$ & $3.904$ (4) & $0.319$ & $-0.864$ & $-0.696$ & $-0.149$ & $-0.196$ & $-0.132$ \\ 
     & s2 & $3.935$ & $3.748$ & $3.935$ (7) & $0.401$ & $-0.957$ & $-0.706$ & $-0.140$ & $-0.293$ & $-0.214$ \\ 
     & rev & $3.765$ & $4.782$ & $11.137$ (1) & $0.253$ & $-0.787$ & $-0.943$ & $+0.598$ & $+0.181$ & $-0.004$ \\ 
    \addlinespace[1pt]
    \PEGP & s1 & $3.907$ & $3.728$ & $3.907$ (7) & $0.402$ & $-0.929$ & $-0.711$ & $-0.138$ & $-0.254$ & $-0.225$ \\ 
     & s2 & $4.014$ & $3.786$ & $4.014$ (7) & $0.477$ & $-1.037$ & $-0.755$ & $-0.123$ & $-0.328$ & $-0.283$ \\ 
     & rev & $3.877$ & $4.957$ & $11.137$ (1) & $0.076$ & $-0.899$ & $-1.162$ & $+0.539$ & $+0.306$ & $+0.133$ \\ 
    \bottomrule
  \end{tabular}

  \vspace{6pt}
  \begin{tabular}{@{}ll cc cc@{}}
    \multicolumn{6}{@{}l}{\textbf{(b)} Closed-loop panel, open-loop scores (\NR)} \\
    \toprule
    & & \multicolumn{2}{c}{\ADEs, $Q$ (m) $\uparrow$} & \multicolumn{2}{c}{\PDMS $\uparrow$} \\
    \cmidrule(lr){3-4}\cmidrule(l){5-6}
    Regime & Run & \Dfin [95\% CI] & \Dend [95\% CI] & \Dfin [95\% CI] & \Dend [95\% CI] \\
    \midrule
    \multicolumn{6}{@{}l}{\textit{\PlanTF}} \\
    \Czero & & \multicolumn{2}{c}{$4.28$} & \multicolumn{2}{c}{$0.760$} \\ 
    \FullFT & s1 & $-1.50$ \ci{-2.88}{+0.11} & $+0.13$ \ci{-1.17}{+1.60} & $-0.085$ \ci{-0.199}{+0.029} & $+0.054$ \ci{-0.031}{+0.142} \\ 
     & s2 & $-0.76$ \ci{-2.07}{+0.81} & $+0.80$ \ci{-0.61}{+2.36} & $-0.066$ \ci{-0.192}{+0.059} & $+0.031$ \ci{-0.064}{+0.128} \\ 
     & rev & $-0.14$ \ci{-1.10}{+0.91} & $+0.51$ \ci{-0.86}{+2.15} & $-0.018$ \ci{-0.112}{+0.075} & $+0.067$ \ci{-0.028}{+0.165} \\ 
    \addlinespace[1pt]
    \AdaFT & s1 & $-0.69$ \ci{-2.30}{+1.04} & $+1.63$ \ci{+0.47}{+3.14} & $+0.012$ \ci{-0.067}{+0.092} & $+0.003$ \ci{-0.098}{+0.105} \\ 
     & s2 & $-0.44$ \ci{-2.00}{+1.24} & $+1.64$ \ci{+0.50}{+3.07} & $-0.009$ \ci{-0.082}{+0.068} & $+0.045$ \ci{-0.044}{+0.138} \\ 
     & rev & $+0.46$ \ci{+0.03}{+0.91} & $+0.90$ \ci{-0.16}{+2.24} & $+0.012$ \ci{-0.040}{+0.066} & $+0.007$ \ci{-0.079}{+0.092} \\ 
    \addlinespace[1pt]
    \Replay & s1 & $+1.53$ \ci{+0.30}{+3.06} & $+1.30$ \ci{+0.19}{+2.70} & $+0.028$ \ci{-0.052}{+0.109} & $+0.031$ \ci{-0.056}{+0.122} \\ 
     & s2 & $+1.13$ \ci{-0.06}{+2.50} & $+1.24$ \ci{+0.24}{+2.40} & $+0.038$ \ci{-0.033}{+0.110} & $+0.029$ \ci{-0.057}{+0.117} \\ 
     & rev & $+0.91$ \ci{-0.08}{+1.95} & $+1.18$ \ci{-0.11}{+2.66} & $-0.007$ \ci{-0.080}{+0.066} & $+0.060$ \ci{-0.014}{+0.142} \\ 
    \addlinespace[1pt]
    \PEGP & s1 & $-0.19$ \ci{-1.66}{+1.46} & $+1.26$ \ci{-0.19}{+2.93} & $+0.005$ \ci{-0.095}{+0.103} & $+0.034$ \ci{-0.055}{+0.124} \\ 
     & s2 & $+0.17$ \ci{-1.24}{+1.77} & $+1.52$ \ci{+0.43}{+2.90} & $-0.041$ \ci{-0.143}{+0.058} & $+0.026$ \ci{-0.061}{+0.116} \\ 
     & rev & $+0.01$ \ci{-0.40}{+0.44} & $+0.98$ \ci{-0.19}{+2.47} & $+0.019$ \ci{-0.038}{+0.078} & $+0.025$ \ci{-0.065}{+0.116} \\ 
    \midrule
    \multicolumn{6}{@{}l}{\textit{\DiffP}} \\
    \Czero & & \multicolumn{2}{c}{$3.10$} & \multicolumn{2}{c}{$0.831$} \\ 
    \FullFT & s1 & $-4.73$ \ci{-6.41}{-3.12} & $-1.23$ \ci{-2.06}{-0.47} & $-0.047$ \ci{-0.124}{+0.033} & $+0.036$ \ci{-0.017}{+0.097} \\ 
     & s2 & $-5.07$ \ci{-6.73}{-3.46} & $-1.25$ \ci{-2.11}{-0.51} & $-0.075$ \ci{-0.149}{+0.000} & $+0.034$ \ci{-0.023}{+0.096} \\ 
     & rev & $-1.00$ \ci{-1.56}{-0.47} & $-0.97$ \ci{-1.69}{-0.29} & $+0.029$ \ci{-0.029}{+0.091} & $+0.036$ \ci{-0.031}{+0.110} \\ 
    \addlinespace[1pt]
    \AdaFT & s1 & $-7.04$ \ci{-9.11}{-5.03} & $-1.29$ \ci{-2.48}{-0.29} & $-0.109$ \ci{-0.187}{-0.029} & $+0.030$ \ci{-0.013}{+0.086} \\ 
     & s2 & $-6.56$ \ci{-8.58}{-4.60} & $-1.26$ \ci{-2.39}{-0.32} & $-0.100$ \ci{-0.177}{-0.018} & $+0.030$ \ci{-0.013}{+0.084} \\ 
     & rev & $-0.73$ \ci{-1.26}{-0.23} & $-1.29$ \ci{-2.45}{-0.32} & $+0.020$ \ci{-0.008}{+0.062} & $+0.026$ \ci{-0.017}{+0.080} \\ 
    \addlinespace[1pt]
    \Replay & s1 & $-0.73$ \ci{-1.33}{-0.15} & $-0.60$ \ci{-1.09}{-0.14} & $+0.032$ \ci{-0.010}{+0.086} & $+0.020$ \ci{-0.007}{+0.063} \\ 
     & s2 & $-0.88$ \ci{-1.56}{-0.24} & $-0.66$ \ci{-1.20}{-0.18} & $+0.032$ \ci{-0.011}{+0.085} & $+0.020$ \ci{-0.007}{+0.059} \\ 
     & rev & $-0.76$ \ci{-1.33}{-0.22} & $-1.47$ \ci{-2.67}{-0.49} & $+0.016$ \ci{-0.012}{+0.057} & $+0.025$ \ci{-0.017}{+0.077} \\ 
    \addlinespace[1pt]
    \PEGP & s1 & $-0.45$ \ci{-1.09}{+0.19} & $-0.63$ \ci{-1.13}{-0.17} & $+0.045$ \ci{-0.006}{+0.108} & $+0.040$ \ci{-0.002}{+0.093} \\ 
     & s2 & $-0.60$ \ci{-1.27}{+0.03} & $-0.71$ \ci{-1.22}{-0.24} & $+0.044$ \ci{-0.008}{+0.106} & $+0.022$ \ci{-0.005}{+0.064} \\ 
     & rev & $-1.07$ \ci{-1.83}{-0.40} & $-1.79$ \ci{-3.02}{-0.74} & $+0.006$ \ci{-0.028}{+0.051} & $+0.007$ \ci{-0.023}{+0.046} \\ 
    \bottomrule
  \end{tabular}
\end{table*}

\paragraph{Scenario transitions.}
Summed over the 12 runs of a planner (56 targets each), targets that pass at \Czero fail at \Tk{7} 146 times and failed targets pass 56 times for \PlanTF under \NR (130 and 65 under \IDMR), against 107 and 23 for \DiffP (121 and 32); only 2 of the 48 run$\times$mode cells of both planners have no pass-to-fail transition. 
Under \NR, every main-run checkpoint fails at least one target that \Czero passes; over its seven checkpoints, a \PlanTF main run breaks at least once 32, 29, 23 and 27 of the 32 targets that \Czero passes (\FullFT, \AdaFT, \Replay, \PEGP) and fixes at least once 15, 15, 15 and 14 of the 24 it fails; for \DiffP these counts are 24, 32, 8 and 4 of 48, and 4, 3, 2 and 3 of 8 (descriptive). 
Per scenario, \DiffP \Replay's main run lowers \CLS-\NR on 32 targets and raises it on 7, but only 8 and 4 by more than 0.05 (17 ties; sign test $p<0.001$ over all changes, $p=0.39$ over those beyond 0.05; descriptive), whereas \FullFT and \AdaFT lower it on 39 and 40 targets and raise it on 7 and 5, all by more than 0.05. 
\Cref{fig:E3} shows the per-update view behind \cref{sec:reference}: the \SR passes lost and gained by the 48 bystander targets of every main-run update (the targets outside the trained group) and their net change, which macro scores net out. 

\begin{figure}[t]
  \centering
  \includegraphics[width=\columnwidth]{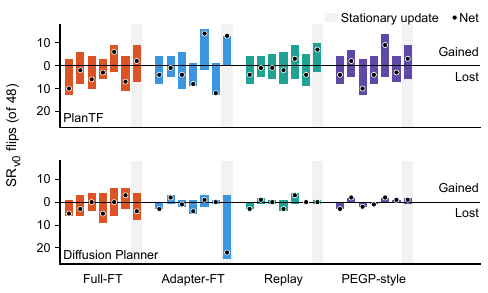}
  \caption{\textbf{Scenario churn per update.}
    \SR flips on the 48 bystander targets of each update (the targets outside the group it trains), main run, \NR: passes gained above zero, passes lost below, net change as a dot; shaded: the \gStationary update (stage~7).
    Counts of 48 per update, no intervals (descriptive).}
  \label{fig:E3}
\end{figure}

\subsection{Axis A: evaluation loop}
\label{app:loop}

\paragraph{Open-loop release gates.}
A gate approves a run when its open-loop \Dfin is positive (point gate) or its interval lies above zero (interval gate); the gates of \cref{sec:loop} and a \PDMS point gate can be read off \cref{tab:E2} for the 12 \PlanTF runs.
The three point gates approve 6 runs each, in three different sets: 7 distinct runs for the cached-ADE and \ADEs gates of \cref{sec:loop}, 9 with the \PDMS gate; every approved run is below \Czero in all four closed-loop point estimates, 3 of the 7 (5 of the 9) also have a closed-loop interval below zero, and the cached-ADE gate's approvals break 7--12 of the 32 targets that \Czero passes and fix 2--8 of the 24 it fails (\NR). 

\paragraph{Calibration of open against closed loop.}
For \DiffP, panel \ADEs gives the same pattern as cached ADE (\cref{sec:loop}): its Spearman $\rho$ with \CLS-\NR \Dfin is $+0.80$ over all 12 runs and $+0.33$ over the 8 runs without collapse. %
Over the 20 runs of both planners without collapse, cached ADE and \CLS-\NR give $\rho=+0.23$ \ci{-0.30}{+0.64} (run bootstrap). 
Update by update (56 main-run updates), the change on the six target groups other than the trained one has opposite signs in cached ADE and \CLS-\NR in 20 updates, against 8 for \CLS-\NR \vs \SR-\NR on the same rollouts (a zero \SR change counts as a disagreement) and 5 for cached \vs panel ADE. 
Within one update, the two loops can disagree with intervals: when \PlanTF \AdaFT learns \gStationary (main run), \ADEs on the 48 other targets worsens ($-1.56$\,m in $Q$, \ci{-2.88}{-0.46}\rob) while \CLS-\NR improves ($+0.226$ \ci{+0.131}{+0.327}\rob; \cref{tab:E9}), and \SR loses none and gains 13 of these scenarios. 
Per scenario (\NR, 12 runs $\times$ 56 targets), the share of targets that pass \SR at \Czero and fail at \Tk{7} does not detectably depend on the direction of the \ADEs change for \PlanTF (over all run--target pairs, 0.224 when \ADEs improves and 0.211 when it worsens; worse minus better $-0.013$ \ci{-0.110}{+0.085}; among \Czero-passed pairs, 0.344 \vs 0.424, $+0.080$ \ci{-0.066}{+0.227}; scenario-cluster bootstrap). 
For \DiffP, a worse \ADEs goes with more pass-to-fail transitions: 0.064 \vs 0.210 over all pairs ($+0.146$ \ci{+0.084}{+0.211}; \cref{fig:loopref}b) and 0.082 \vs 0.234 among \Czero-passed pairs ($+0.151$ \ci{+0.082}{+0.223}). %

\paragraph{Non-reactive against IDM-reactive closed loop.}
\NR and \IDMR give the same three-valued \Dfin verdict in 36 of the 48 run$\times$metric cells of \cref{tab:E1}, and the same best and the same worst \CLS regime in 5 of 6 planner$\times$setting groups; in the exception, the \PlanTF reverse-order run, \PEGP is best and \FullFT worst under \NR and the two swap places under \IDMR. 
Mean Kendall $\tau_b$ of final regime rankings (\CLS and \SR) is 1.00 between the seeds and 0.79 between the orders under \NR, but 0.77 and $-0.26$ under \IDMR for \PlanTF; for \DiffP, it is 0.67 and 0.67 under \NR against 0.79 and 0.37 under \IDMR (descriptive). %
The sign of \CLS backward transfer differs between the modes in 6 of 24 runs for \BWTold and in 7 of 24 for \BWT (\cref{tab:E1b}). 
The gap between the modes does not isolate interaction: IDM vehicles deviate from their logged motion by a median of 11.8\,m (mean per rollout; interquartile range 7.4--16.5\,m over the 1,616 \IDMR rollouts of the main runs with at least one IDM vehicle), and two of the six panel scenarios that start without any IDM vehicle, both \gFollow, differ in \CLS between the modes by 0.95 to 0.99 at \Czero for both planners. 
}{}
\IfFileExists{sec/app_e_axes.tex}{
\subsection{Axes B--D: reference, aggregation and stream}
\label{app:axes}

This subsection extends \cref{sec:reference,sec:aggregation,sec:order}; intervals and \rob{} are as in \cref{sec:axes}, and closed-loop values refer to the 56 targets.

\begin{table*}[t]
  \centering
  \caption{\textbf{Statistics anchored at \Tk{1}} (\PlanTF, main run, \NR).
    Macro changes from \Czero to \Tk{1}; from \Tk{1} to the learning endpoints $Q(j,j)$ (group~1 contributes 0) and to the final checkpoint; and from \Czero to the endpoints (\Dend); all seven groups.
    \ADEs in $Q=-$ADE (m; positive: lower error).
    Share: ($\Czero\to\Tk{1}$)\,/\,\Dend of point estimates (--: not applicable, as \Dend is a gain).
    The adapter regimes share \Tk{1}.}
  \label{tab:E7}
  \scriptsize
  \setlength{\tabcolsep}{4.5pt}
  \begin{tabular}{@{}llrrrrc@{}}
    \toprule
    Regime & Metric & \multicolumn{1}{c}{$\Czero\to\Tk{1}$} & \multicolumn{1}{c}{$\Tk{1}\to{}$endpoint} & \multicolumn{1}{c}{$\Tk{1}\to{}$final} & \multicolumn{1}{c}{$\Czero\to{}$endpoint (\Dend)} & Share (\%) \\
    \midrule
\FullFT & \CLS-\NR & $-0.217$ \ci{-0.334}{-0.099} & $-0.017$ \ci{-0.108}{+0.077} & $-0.053$ \ci{-0.178}{+0.073} & $-0.234$ \ci{-0.365}{-0.110} & 93 \\ 
 & \SR-\NR & $-0.214$ \ci{-0.339}{-0.089} & $-0.036$ \ci{-0.143}{+0.071} & $-0.054$ \ci{-0.196}{+0.089} & $-0.250$ \ci{-0.393}{-0.107} & 86 \\ 
 & \ADEs & $+0.47$ \ci{-0.03}{+0.97} & $-0.34$ \ci{-1.61}{+1.05} & $-1.97$ \ci{-3.31}{-0.40} & $+0.13$ \ci{-1.17}{+1.60} & -- \\ 
    \addlinespace[2pt]
\AdaFT & \CLS-\NR & $-0.075$ \ci{-0.185}{+0.033} & $-0.097$ \ci{-0.207}{+0.013} & $-0.075$ \ci{-0.193}{+0.041} & $-0.172$ \ci{-0.300}{-0.044} & 44 \\ 
 & \SR-\NR & $-0.071$ \ci{-0.196}{+0.054} & $-0.107$ \ci{-0.232}{+0.036} & $-0.071$ \ci{-0.214}{+0.071} & $-0.179$ \ci{-0.321}{-0.036} & 40 \\ 
 & \ADEs & $+0.40$ \ci{-0.00}{+0.86} & $+1.23$ \ci{+0.09}{+2.67} & $-1.09$ \ci{-2.84}{+0.72} & $+1.63$ \ci{+0.47}{+3.14} & -- \\ 
    \addlinespace[2pt]
\Replay & \CLS-\NR & $-0.075$ \ci{-0.185}{+0.033} & $-0.023$ \ci{-0.121}{+0.073} & $+0.001$ \ci{-0.107}{+0.106} & $-0.098$ \ci{-0.222}{+0.027} & 77 \\ 
 & \SR-\NR & $-0.071$ \ci{-0.196}{+0.054} & $-0.018$ \ci{-0.125}{+0.107} & $0.000$ \ci{-0.125}{+0.125} & $-0.089$ \ci{-0.232}{+0.054} & 80 \\ 
 & \ADEs & $+0.40$ \ci{-0.00}{+0.86} & $+0.90$ \ci{-0.15}{+2.28} & $+1.12$ \ci{-0.01}{+2.60} & $+1.30$ \ci{+0.19}{+2.70} & -- \\ 
    \addlinespace[2pt]
\PEGP & \CLS-\NR & $-0.075$ \ci{-0.185}{+0.033} & $-0.011$ \ci{-0.105}{+0.082} & $-0.047$ \ci{-0.166}{+0.074} & $-0.086$ \ci{-0.224}{+0.050} & 87 \\ 
 & \SR-\NR & $-0.071$ \ci{-0.196}{+0.054} & $-0.018$ \ci{-0.125}{+0.090} & $-0.036$ \ci{-0.179}{+0.089} & $-0.089$ \ci{-0.232}{+0.054} & 80 \\ 
 & \ADEs & $+0.40$ \ci{-0.00}{+0.86} & $+0.85$ \ci{-0.53}{+2.45} & $-0.59$ \ci{-2.16}{+1.09} & $+1.26$ \ci{-0.19}{+2.93} & -- \\ 
    \bottomrule
  \end{tabular}
\end{table*}

\paragraph{Reference: statistics anchored at \Tk{1}.}
\Cref{tab:E7} measures the \PlanTF statistics from \Tk{1} instead of \Czero (main run, the only run with full \NR stage matrices); none of the 32 closed-loop intervals measured from \Tk{1} excludes zero (four regimes; \CLS-\NR and \SR-\NR; learning endpoint and final checkpoint; all seven groups, as tabulated, or groups 2--7, not tabulated). 
The first update, on the \Czero pretraining group \gStraight, carries 77--93\% of the \Czero-to-endpoint \CLS-\NR loss of \FullFT, \Replay and \PEGP but 44\% of \AdaFT's, whose decline from \Tk{1} to the learning endpoint ($-0.097$ \ci{-0.207}{+0.013}) is larger than its first-update drop ($-0.075$ \ci{-0.185}{+0.033}), although neither interval excludes zero. 
Open loop moves the other way: after \Tk{1}, \AdaFT's \ADEs improves up to the learning endpoint ($+1.23$\,m \ci{+0.09}{+2.67} in $Q$), and \FullFT's worsens up to the final checkpoint ($-1.97$\,m \ci{-3.31}{-0.40}). 
For \DiffP (main run), the first update changes \CLS-\NR by $-0.089$ \ci{-0.167}{-0.019} (\FullFT) and $-0.048$ \ci{-0.103}{-0.005} (shared adapter \Tk{1}) and raises the 7-group cached ADE of the adapter regimes by 0.58\,m, confounded with the switch from EMA to raw weights (\cref{sec:deployment}). 

\paragraph{Reference: other families, acquisition and churn.}
Pooled over the 12 \PlanTF runs as in \cref{fig:loopref}a, \Dend and, except for \SR-\IDMR (upper bound $+0.003$), \Dfin lie below zero in all four closed-loop families while \BWT straddles zero; \PDMS shows no detectable change in any of the three statistics. 
Learning a group does not reliably register as a closed-loop gain over the previous checkpoint: \PlanTF's mean \CLS acquisition is negative in 8 of 12 runs under \NR and in 6 of 12 under \IDMR, although its open-loop acquisition is positive in all 12 (descriptive). 
For \DiffP, cached acquisition is positive in 38 of 84 group cases (12 runs $\times$ 7 groups), yet all 84 learning endpoints have a higher cached error than \Czero. 
The seed-to-seed floor of \SR churn (targets whose \SR at \Tk{7} differs between the two forward-order seeds) is 4, 8, 6 and 13 for \PlanTF \FullFT, \AdaFT, \Replay and \PEGP and 8, 5, 1 and 5 for \DiffP, against 23, 16, 14, 12 and 17, 30, 3, 4 targets whose \SR differs between \Czero and \Tk{7} in the main run; for \PEGP, this churn does not exceed the floor for either planner (descriptive). 

\begin{table*}[t]
  \centering
  \caption{\textbf{Sentinel retention} (\gSentinel, evaluation only), per run.
    Closed loop (\NR, 8 scenarios): mean \CLS at \Tk{7}; $p$: exact two-sided sign-flip test of the eight paired $\Tk{7}-\Czero$ differences; \SR passes at \Tk{7}.
    Cached ADE (640 samples, m; lower is better) at \Tk{7}.
    Closed-loop values are descriptive ($n=8$); cached values are task means without intervals.}
  \label{tab:E6}
  \scriptsize
  \setlength{\tabcolsep}{4pt}
  \begin{tabular}{@{}lccccccccc@{\hspace{10pt}}ccc@{}}
    \toprule
    & \multicolumn{3}{c}{Main run} & \multicolumn{3}{c}{Second seed} & \multicolumn{3}{c}{Reverse order} & \multicolumn{3}{c}{Cached ADE at \Tk{7} (m)} \\
    \cmidrule(lr){2-4}\cmidrule(lr){5-7}\cmidrule(lr){8-10}\cmidrule(l){11-13}
    Regime & \CLS & $p$ & \SR & \CLS & $p$ & \SR & \CLS & $p$ & \SR & s1 & s2 & rev \\
    \midrule
\multicolumn{13}{@{}l}{\textit{\PlanTF} (\Czero: \CLS-\NR 0.732, \SR 6/8, cached ADE 2.78\,m)} \\ 
\FullFT & 0.125 & 0.062 & 1/8 & 0.125 & 0.062 & 1/8 & 0.724 & 0.531 & 6/8 & 5.78 & 5.02 & 2.58 \\ 
\AdaFT & 0.691 & 0.719 & 6/8 & 0.436 & 0.156 & 4/8 & 0.669 & 0.562 & 6/8 & 3.40 & 3.05 & 2.62 \\ 
\Replay & 0.756 & 1.000 & 7/8 & 0.670 & 0.625 & 6/8 & 0.233 & 0.062 & 2/8 & 2.96 & 3.06 & 2.74 \\ 
\PEGP & 0.733 & 1.000 & 6/8 & 0.717 & 0.625 & 6/8 & 0.489 & 0.125 & 4/8 & 3.54 & 3.12 & 2.75 \\ 
\multicolumn{13}{@{}l}{\textit{\DiffP} (\Czero: \CLS-\NR 0.984, \SR 8/8, cached ADE 4.70\,m)} \\ 
\FullFT & 0.296 & 0.016 & 3/8 & 0.377 & 0.016 & 4/8 & 0.829 & 0.031 & 7/8 & 17.92 & 17.71 & 7.95 \\ 
\AdaFT & 0.125 & 0.016 & 1/8 & 0.285 & 0.016 & 3/8 & 0.940 & 0.031 & 8/8 & 23.24 & 22.41 & 7.42 \\ 
\Replay & 0.934 & 0.031 & 8/8 & 0.934 & 0.031 & 8/8 & 0.938 & 0.031 & 8/8 & 7.92 & 8.32 & 7.71 \\ 
\PEGP & 0.937 & 0.031 & 8/8 & 0.933 & 0.031 & 8/8 & 0.937 & 0.031 & 8/8 & 7.25 & 7.71 & 8.08 \\ 
    \bottomrule
  \end{tabular}
\end{table*}

\paragraph{Sentinel retention.}
\Cref{tab:E6} reports the evaluation-only sentinel \gSentinel at \Tk{7} in all 24 runs.
For \DiffP, all twelve runs end below \Czero on the sentinel in closed loop, and in none does a sentinel scenario improve; \Replay and \PEGP keep \SR at 8/8 in all three runs, whereas \FullFT and \AdaFT fall to a \CLS-\NR of 0.125--0.377 in both forward-order seeds. 
In cached sentinel ADE, the shared adapter \Tk{1} already reaches 6.470\,m in the main run (\Czero: 4.696\,m) and 23.31\,m in reverse order, where \gStationary opens the stream; there the four regimes end at 7.42--8.08\,m. 
For \PlanTF, whose \Czero was pretrained on the sentinel category, the closed-loop values are single-group values (descriptive): \Replay and \PEGP end the main run at 0.756 and 0.733 against 0.732 at \Czero, with 1 and 4 of the eight scenarios higher and 5 and 2 lower, but the reverse-order run at 0.233 and 0.489. 
Its cached sentinel error ends above \Czero (2.78\,m) in the forward-order runs (2.96--5.78\,m) and below it in the reverse-order run (2.58--2.75\,m), whose last update trains the \PlanTF-\Czero pretraining group \gStraight (descriptive). 

\begin{table}[t]
  \centering
  \caption{\textbf{Stream axis per metric family.}
    Kendall $\tau_b$ between the regime rankings (four regimes) of the main run and the second seed (S) or the reverse-order run (O), for final values and for \BWTold.
    Below the rule: means over the open-loop (5), closed-loop (4), \NR (2), \IDMR (2) and all (9) families, and the number of families in which both runs have the same worst regime.
    One seed pair and one order pair; descriptive.}
  \label{tab:E8}
  \scriptsize
  \setlength{\tabcolsep}{1.9pt}
  \begin{tabular}{@{}lcccccccc@{}}
    \toprule
    & \multicolumn{4}{c}{\PlanTF} & \multicolumn{4}{c}{\DiffP} \\
    \cmidrule(lr){2-5}\cmidrule(l){6-9}
    & \multicolumn{2}{c}{Final} & \multicolumn{2}{c}{\BWTold} & \multicolumn{2}{c}{Final} & \multicolumn{2}{c}{\BWTold} \\
    \cmidrule(lr){2-3}\cmidrule(lr){4-5}\cmidrule(lr){6-7}\cmidrule(l){8-9}
    Family & S & O & S & O & S & O & S & O \\
    \midrule
Cached ADE & $+1.00$ & $+0.67$ & $+0.33$ & $0.00$ & $+1.00$ & $0.00$ & $+1.00$ & $+0.33$ \\ 
Cached FDE & $+1.00$ & $+0.67$ & $0.00$ & $-0.33$ & $+0.67$ & $-0.33$ & $+1.00$ & $+0.67$ \\ 
\ADEs & $+1.00$ & $+0.67$ & $+1.00$ & $0.00$ & $+1.00$ & $-0.67$ & $+1.00$ & $+0.67$ \\ 
\FDEs & $+1.00$ & $+0.67$ & $+0.67$ & $+0.33$ & $+1.00$ & $-0.67$ & $+1.00$ & $+0.33$ \\ 
\PDMS & $+1.00$ & $0.00$ & $+0.67$ & $+0.33$ & $+1.00$ & $-0.67$ & $+0.67$ & $0.00$ \\ 
\CLS-\NR & $+1.00$ & $+0.67$ & $+0.33$ & $+0.33$ & $+0.67$ & $+0.67$ & $+0.67$ & $+0.67$ \\ 
\SR-\NR & $+1.00$ & $+0.91$ & $+0.80$ & $0.00$ & $+0.67$ & $+0.67$ & $+0.67$ & $+0.55$ \\ 
\CLS-\IDMR & $+1.00$ & $-0.33$ & $+0.67$ & $-0.33$ & $+0.67$ & $+0.33$ & $+1.00$ & $-0.67$ \\ 
\SR-\IDMR & $+0.55$ & $-0.18$ & $+0.18$ & $-0.33$ & $+0.91$ & $+0.40$ & $+1.00$ & $-0.55$ \\ 
\midrule
Open (5) & $+1.00$ & $+0.53$ & $+0.53$ & $+0.07$ & $+0.93$ & $-0.47$ & $+0.93$ & $+0.40$ \\ 
Closed (4) & $+0.89$ & $+0.27$ & $+0.50$ & $-0.08$ & $+0.73$ & $+0.52$ & $+0.83$ & $0.00$ \\ 
\NR (2) & $+1.00$ & $+0.79$ & $+0.57$ & $+0.17$ & $+0.67$ & $+0.67$ & $+0.67$ & $+0.61$ \\ 
\IDMR (2) & $+0.77$ & $-0.26$ & $+0.42$ & $-0.33$ & $+0.79$ & $+0.37$ & $+1.00$ & $-0.61$ \\ 
All (9) & $+0.95$ & $+0.41$ & $+0.52$ & $0.00$ & $+0.84$ & $-0.03$ & $+0.89$ & $+0.22$ \\ 
\midrule
Same worst & 9 & 7 & 4 & 1 & 9 & 0 & 9 & 0 \\ 
    \bottomrule
  \end{tabular}
\end{table}

\paragraph{Aggregation: stage-averaged and worst-stage readings.}
In the main run, no regime's stage-averaged or worst-stage \CLS-\NR lies above \Czero (\cref{fig:aggorder}a). 
Both lie robustly below \Czero (\rob) for \PlanTF \FullFT, \AdaFT and \PEGP and for \DiffP \FullFT and \AdaFT, and the worst stage also for \PlanTF \Replay, whose stage average, $-0.092$ \ci{-0.184}{-0.006}, excludes zero only under the task-stratified bootstrap (log-cluster upper bound $+0.002$, sign-flip $p=0.055$). 
Between regimes (six Holm-adjusted contrasts per planner and summary), stage averaging robustly separates every \PlanTF pair that contains \FullFT, and \AdaFT from \Replay, whereas at the final checkpoint only \FullFT minus \Replay is robust; for \DiffP, five contrasts are robust at the final checkpoint and four stage-averaged, and \Replay and \PEGP cannot be separated under either summary. 

\paragraph{Stream: rankings per metric family.}
\Cref{tab:E8} splits the mean Kendall $\tau_b$ of \cref{sec:order} by metric family (descriptive).
Final rankings agree perfectly across the two seeds in eight of nine families for \PlanTF (all but \SR-\IDMR) and in four for \DiffP. 
Across the two orders, \PlanTF loses agreement mainly in the two \IDMR families and in \PDMS, and \DiffP in the open-loop families ($\tau_b$ from 0.00 to $-0.67$), whereas its closed-loop families stay positive ($+0.33$ to $+0.67$). 
The worst regime is the same across seeds in all nine families for both planners, and across orders in seven of nine for \PlanTF and in none for \DiffP. 
Across seeds, \BWTold rankings agree less than final rankings for \PlanTF (mean $\tau_b$ 0.52 \vs 0.95) but not for \DiffP (0.89 \vs 0.84); across orders they agree little (0.00 and 0.22). 
With the third order, all three order pairs still agree far less in final rankings than the two seeds (\cref{app:third-order}).

\begin{table*}[t]
  \centering
  \caption{\textbf{The \gStationary update.}
    Changes in cached $Q=-$ADE (m; positive: lower error) from \Tk{k-1} to \Tk{k} at the \gStationary stage ($k=7$ forward, $k=1$ reverse): collateral (mean change on the six other target groups), sentinel (\gSentinel) and the 7-group macro; the adapter regimes share their reverse-order \Tk{1}.
    Recovery: share of the reverse-order \Tk{1} macro shock undone at \Tk{2}, computed for the \DiffP collapse (--: not computed for \PlanTF); task means without intervals (descriptive).
    Last column: change of \CLS-\NR on the 48 bystanders from \Tk{6} to \Tk{7} (main run); bold: its paired, task-stratified 95\% interval excludes zero (not adjusted for multiplicity).}
  \label{tab:E9}
  \scriptsize
  \setlength{\tabcolsep}{4pt}
  \begin{tabular}{@{}lccccccccccc@{}}
    \toprule
    & \multicolumn{3}{c}{Collateral (six groups)} & \multicolumn{3}{c}{Sentinel} & \multicolumn{3}{c}{Macro (seven groups)} & & \\
    \cmidrule(lr){2-4}\cmidrule(lr){5-7}\cmidrule(lr){8-10}
    Regime & s1 & s2 & rev & s1 & s2 & rev & s1 & s2 & rev & Recovery (\%) & \CLS-\NR \\
    \midrule
\multicolumn{12}{@{}l}{\textit{\PlanTF}} \\
\FullFT & $-0.87$ & $-0.49$ & $-1.63$ & $+0.14$ & $-0.19$ & $-6.71$ & $-0.66$ & $-0.31$ & $-1.41$ & -- & $+0.022$ \\ 
\AdaFT & $-0.87$ & $-0.43$ & $-0.50$ & $+0.36$ & $+0.32$ & $-0.31$ & $-0.69$ & $-0.30$ & $-0.43$ & -- & $\mathbf{+0.226}$ \\ 
\Replay & $+0.41$ & $+0.45$ & $-0.50$ & $+0.40$ & $+0.10$ & $-0.31$ & $+0.46$ & $+0.48$ & $-0.43$ & -- & $\mathbf{+0.129}$ \\ 
\PEGP & $-0.22$ & $-0.40$ & $-0.50$ & $+2.09$ & $+0.89$ & $-0.31$ & $-0.16$ & $-0.30$ & $-0.43$ & -- & $+0.067$ \\ 
\multicolumn{12}{@{}l}{\textit{\DiffP}} \\
\FullFT & $-4.70$ & $-4.50$ & $-4.62$ & $-8.96$ & $-8.09$ & $-8.56$ & $-4.10$ & $-3.92$ & $-4.03$ & 66.4 & $\mathbf{-0.157}$ \\ 
\AdaFT & $-8.22$ & $-7.86$ & $-9.32$ & $-14.80$ & $-14.21$ & $-18.61$ & $-7.22$ & $-6.89$ & $-8.16$ & 88.4 & $\mathbf{-0.477}$ \\ 
\Replay & $-0.10$ & $-0.19$ & $-9.32$ & $+0.11$ & $-0.25$ & $-18.61$ & $-0.09$ & $-0.16$ & $-8.16$ & 88.0 & $-0.001$ \\ 
\PEGP & $-0.25$ & $-0.31$ & $-9.32$ & $-0.01$ & $-0.17$ & $-18.61$ & $-0.21$ & $-0.26$ & $-8.16$ & 85.1 & $+0.007$ \\ 
    \bottomrule
  \end{tabular}
\end{table*}

\begin{figure*}[t]
  \centering
  \includegraphics[width=\textwidth]{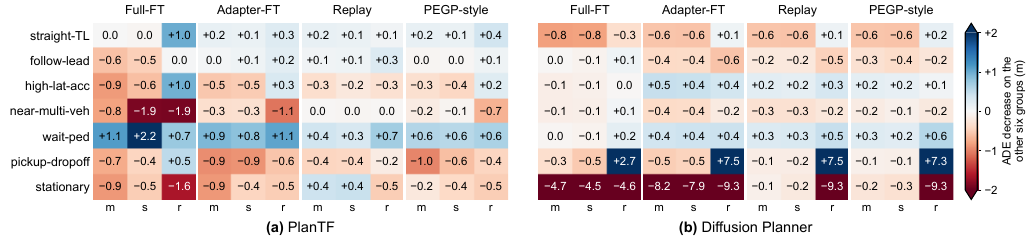}
  \caption{\textbf{Collateral of single updates} (cached ADE).
    Each cell: mean decrease in cached ADE (m) on the six target groups other than the trained one, caused by the update on the row's group, in each run of (a)~\PlanTF and (b)~\DiffP (m: main run; s: second seed; r: reverse order); positive (blue): the update lowered their error.
    Colours are clipped at $\pm2$\,m.
    A row collects the same group at any stage; the adapter regimes share update~1 (\gStraight in forward order, \gStationary in reverse order).
    Task means over 640 samples per group (descriptive).}
  \label{fig:E6}
\end{figure*}

\paragraph{Stream: signs, verdicts and order-to-seed ratios.}
Of the 54 pairwise final comparisons per planner (6 regime pairs~$\times$ 9 families), 36 (\PlanTF) and 22 (\DiffP) keep their sign in all three runs. 
Of the 32 \Czero-referenced \Dfin verdicts per stream setting (\cref{tab:E1}; planner$\times$regime$\times$mode$\times$metric), the three that change between the seeds all go from no detectable change to worse, eleven of the twelve that change between the orders go from worse in the main run to no detectable change in reverse order, and no cell changes on both axes (descriptive). 
The median order-to-seed ratio of final values over the nine families, $|$reverse$-$main$|\,/\,|$second seed$-$main$|$, is 13.1, 8.08, 1.73 and 0.775 for \DiffP \AdaFT, \FullFT, \PEGP and \Replay, but 2.99, 1.34, 2.52 and 2.70 for \PlanTF \FullFT, \AdaFT, \Replay and \PEGP (one seed pair each; descriptive). 
For \DiffP \AdaFT, final \CLS-\NR rises by $+0.458$ \ci{+0.326}{+0.584} from the main to the reverse-order run but by only 0.046 to the second seed. 
\DiffP \Replay's lead over \FullFT in final \CLS-\NR, which holds under both seed and order, is $+0.291$ \ci{+0.197}{+0.387} and $+0.336$ \ci{+0.237}{+0.438} in the forward-order seeds and $+0.107$ \ci{+0.028}{+0.198} in reverse order: smaller, but above zero in all three runs. 

\subsection{Interference dynamics}
\label{app:interference}

This subsection gives the single-update evidence behind \cref{sec:why}; the \emph{collateral} of the update at stage~$k$ is the mean change in cached $Q=-$ADE (m) from \Tk{k-1} to \Tk{k} on the six target groups other than the trained one (positive: lower error; task means over 640 samples per group, descriptive).

\paragraph{The \gStationary update.}
\Cref{tab:E9} lists its cached-ADE effects in all 24 runs and its closed-loop effect on the bystanders in the main runs.
Up to \Tk{6}, \DiffP \AdaFT shows no detectable change in \CLS-\NR from \Czero ($-0.038$ \ci{-0.104}{+0.028}); the \gStationary update then costs $-0.468$ \ci{-0.587}{-0.344}\rob. 
\FullFT loses on both sides of \Tk{6} ($-0.121$ \ci{-0.212}{-0.033}\rob{} and $-0.164$ \ci{-0.262}{-0.069}\rob), and \Replay and \PEGP change by less than 0.01 on either side, with intervals that include zero. 
On the 40 targets of groups 1--5, \AdaFT minus \FullFT in \BWT is $+0.035$ \ci{-0.087}{+0.158} measured at \Tk{6} but $-0.318$ \ci{-0.458}{-0.189} at \Tk{7}: their forgetting differs detectably only after the last update. 
Afterwards, the \DiffP \AdaFT model hardly depends on the path: its cached errors on the seven target groups and the sentinel (mean absolute log ratio) differ by 0.78\% between the main-run \Tk{7} and the reverse-order \Tk{1}, less than between the two forward-order seeds at \Tk{7} (3.35\%), although the six preceding main-run updates had moved it by 27.95\% from \Czero (\FullFT: 17.8\%, 1.54\% and 36.45\%; descriptive, as position, preceding state and random streams also differ). 
In the third order, where it comes fourth, it collapses \DiffP \AdaFT again, and the next update undoes the collapse (\cref{app:third-order}).
For \PlanTF the same update is mild (median collateral $-0.49$\,m, positive in 2 of 12 runs). 

\paragraph{The other updates.}
\Cref{fig:E6} extends the \gStationary rows of \cref{tab:E9} to all seven updates.
\gWaitPed is benign: its update lowers the error on the other groups in all 12 \PlanTF runs (median $+0.72$\,m) and in 10 of 12 \DiffP runs (median $+0.29$\,m). 
\gNearMulti raises it in all 12 \PlanTF runs (median $-0.33$\,m) and in 11 of 12 \DiffP runs (median $-0.19$\,m), and \gPickup, which trains on raw car-park scenes but is evaluated mostly on pickup/drop-off scenes (\cref{sec:stream}), in 11 of 12 \PlanTF runs (median $-0.49$\,m); its large positive \DiffP values in reverse order ($+2.73$ to $+7.50$\,m) are recovery from the stage-1 collapse, since it comes second there. 
No single \PlanTF update dominates: the largest single drop of an old group most often comes from \gPickup (30 of 72 cases; \gStationary: 15; \gNearMulti: 10), and the main-run and reverse-order collateral correlates at $r=0.53$ aligned by group identity but at $r=-0.05$ aligned by stream position (main run against second seed: $r=0.88$; 28 updates each). 
Budget alone does not set the damage: \gStraight and \gStationary receive the same 7,820 updates, yet their \PlanTF collateral has medians of $+0.14$ and $-0.49$\,m, and the within-run Spearman correlation between budget and collateral has a median of $+0.44$ for \PlanTF ($+0.89$ without \gStationary) and $-0.49$ for \DiffP ($-0.31$ without it). 
With 10,000 samples and 800 updates for every group, the \gStationary update still dominates the collateral of \DiffP \AdaFT (\cref{app:equal-data}).
In closed loop (main runs; intervals on the 48 bystanders, unadjusted for multiplicity), the \gWaitPed updates of \PlanTF \FullFT, \AdaFT and \PEGP raise \CLS-\NR detectably (\AdaFT: $+0.260$ \ci{+0.138}{+0.384}); \PlanTF intervals below zero belong to the \gPickup updates of \FullFT and \AdaFT, \AdaFT's \gNearMulti update, \PEGP's \gHighLat update and \FullFT's first update, and no \DiffP interval lies above zero (descriptive). 

\section{Extensions of the Core Suite}
\label{app:extensions}
\setcounter{table}{0}
\setcounter{figure}{0}
\setcounter{equation}{0}
\providecommand{\LoRAFT}{LoRA-FT\xspace}
\providecommand{\LoRAReplay}{LoRA+Replay\xspace}
\providecommand{\LoRAPEGP}{LoRA+PEGP\xspace}

The core suite (\cref{app:results}) varies the regime, the planner, the training seed and the direction of the stream.
Five extensions, defined in \cref{sec:extensions}, vary further parts of the setup, with 38 training runs in all (\cref{tab:extensions}): the order of the stream (third order, \cref{app:third-order}), sequential training itself (joint references, \cref{app:joint}), the size of the groups (equal-data runs, \cref{app:equal-data}), the size of the closed-loop panel (expanded panel, \cref{app:expanded-endpoints}) and the trainable parameters (LoRA branch, \cref{app:lora}).
Unless stated otherwise, they keep the deployed checkpoints \Czero, the groups, the development samples, the recipe and training seed 20261014 of the main run, and the core panel under \NR and \IDMR.
Intervals and three-valued verdicts refer to the core suite; we report the extensions as point estimates (descriptive).
Their checkpoints will be released with those of the core suite.

As point estimates, two extensions support the deployment verdict of the core suite (\cref{sec:deploy}): on the expanded panel, every main-run final checkpoint scores below \Czero on all four closed-loop readings, and no joint reference exceeds its \Czero in closed-loop \CLS, although joint training recovers much of the sequential loss.
The LoRA branch is the main exception: four of its runs end at or slightly above \Czero in \CLS-\NR (by at most $+0.022$), and some of its other closed-loop readings lie above \Czero as well (under \IDMR up to $+0.035$ in \CLS and 51 \vs 48 of 56 in \SR), all within the resolution of the core panel; we do not extend the invariant to it.
Two \IDMR readings of the equal-data runs also lie slightly above \Czero (\cref{app:equal-data}).
The third order and the equal-data runs probe the \gStationary update behind the collapse of \DiffP (\cref{sec:why}).
In the third order, where it comes fourth, it still collapses \AdaFT and the next update undoes the collapse, so the final-checkpoint margin of \Replay and \PEGP over \AdaFT shrinks while stage-averaged and worst-stage readings still separate them; training seeds still move rankings far less than orders.
With equal data, the shock halves but stays far larger than that of any other update.

\begin{table}[t]
  \centering
  \caption{\textbf{Extensions of the core suite.}
    Runs: training runs added for the two planners (--: none, as the expanded panel re-evaluates the main runs).
    Closed loop: checkpoints evaluated under \NR and \IDMR (b/a: each group before and after its own stage).
    All runs use training seed 20261014.}
  \label{tab:extensions}
  \scriptsize
  \setlength{\tabcolsep}{3pt}
  \begin{tabular}{@{}l r >{\raggedright\arraybackslash}p{1.32in} l@{}}
    \toprule
    Extension & Runs & Change from the core suite & Closed loop \\
    \midrule
    Third order & 8 & four regimes; groups in the order 3, 4, 2, 7, 5, 1, 6 & \Czero, b/a, \Tk{7} \\
    Joint references & 4 & \FullFT, \AdaFT on the union of the seven groups & \Czero, final \\
    Equal-data runs & 8 & four regimes; 10,000 samples per group & \Czero, b/a, \Tk{7} \\
    Expanded panel & -- & 720 scenarios (630 targets); main runs & \Czero, \Tk{7} \\
    LoRA branch & 18 & \LoRAFT, \LoRAReplay, \LoRAPEGP; three orders & \Czero, b/a, \Tk{7} \\
    \bottomrule
  \end{tabular}
\end{table}

\subsection{Third order}
\label{app:third-order}

\paragraph{Design.}
The third order is \gHighLat, \gNearMulti, \gFollow, \gStationary, \gWaitPed, \gStraight, \gPickup (groups 3, 4, 2, 7, 5, 1, 6 of the forward order), a fixed permutation that is not the mirror of the forward order: it moves \gStationary, last in the forward and first in the reverse order, to stage~4, and \gNearMulti, fourth in both, to stage~2.
Its eight runs (four regimes, two planners) keep the deployed checkpoints, full training pools, recipe and training seed 20261014 of the main and reverse-order runs, so each takes 32,420 updates and 8,278,780 current-sample visits.
Because replay starts at stage~2 and this order opens with the smallest group, \Replay adds 2,013,260 replay visits (24.3\% of current-sample visits; 19.0\% in the other two orders).
Closed loop covers \Czero, \Tk{7} and each group before and after its stage under both modes.
As with reversal, this order changes positions, preceding checkpoints and random streams together, so it measures order sensitivity, not the effect of one position.

\begin{table}[t]
  \centering
  \caption{\textbf{Final scores in three orders} (training seed 20261014; forward and reverse: the main and reverse-order runs of the core suite).
    Cached ADE and FDE: 7-group macro over 640 development samples per group (m); \PDMS, \CLS and \SR: 56 targets of the core panel, \SR as passes; \Czero: the deployed checkpoint.
    Point estimates (descriptive).}
  \label{tab:third-order-final}
  \scriptsize
  \setlength{\tabcolsep}{2pt}
  \begin{tabular}{@{}ll rrrrrrr@{}}
    \toprule
    & & \multicolumn{2}{c}{Cached, m $\downarrow$} & \multicolumn{3}{c}{\NR $\uparrow$} & \multicolumn{2}{c}{\IDMR $\uparrow$} \\
    \cmidrule(lr){3-4}\cmidrule(lr){5-7}\cmidrule(l){8-9}
    Regime & Order & ADE & FDE & \PDMS & \CLS & \SR & \CLS & \SR \\
    \midrule
    \multicolumn{9}{@{}l}{\textit{\PlanTF}} \\
    \Czero & & $3.561$ & $7.680$ & $0.760$ & $0.528$ & $32$ & $0.557$ & $34$ \\
    \addlinespace[1pt]
    \FullFT & Forward & $5.489$ & $10.980$ & $0.675$ & $0.258$ & $17$ & $0.310$ & $20$ \\
     & Reverse & $3.367$ & $7.830$ & $0.742$ & $0.364$ & $22$ & $0.506$ & $31$ \\
     & Third & $4.152$ & $9.299$ & $0.703$ & $0.391$ & $23$ & $0.239$ & $14$ \\
    \addlinespace[1pt]
    \AdaFT & Forward & $4.842$ & $9.431$ & $0.772$ & $0.378$ & $24$ & $0.501$ & $33$ \\
     & Reverse & $3.248$ & $7.258$ & $0.772$ & $0.454$ & $27$ & $0.471$ & $29$ \\
     & Third & $3.883$ & $8.882$ & $0.714$ & $0.351$ & $22$ & $0.379$ & $26$ \\
    \addlinespace[1pt]
    \Replay & Forward & $3.034$ & $7.010$ & $0.788$ & $0.454$ & $28$ & $0.549$ & $33$ \\
     & Reverse & $3.063$ & $6.965$ & $0.753$ & $0.482$ & $31$ & $0.486$ & $30$ \\
     & Third & $3.100$ & $7.175$ & $0.740$ & $0.391$ & $24$ & $0.482$ & $29$ \\
    \addlinespace[1pt]
    \PEGP & Forward & $4.208$ & $8.888$ & $0.765$ & $0.406$ & $26$ & $0.508$ & $32$ \\
     & Reverse & $3.364$ & $7.496$ & $0.779$ & $0.510$ & $31$ & $0.442$ & $27$ \\
     & Third & $3.464$ & $7.905$ & $0.744$ & $0.495$ & $31$ & $0.496$ & $32$ \\
    \midrule
    \multicolumn{9}{@{}l}{\textit{\DiffP}} \\
    \Czero & & $2.978$ & $6.721$ & $0.831$ & $0.823$ & $48$ & $0.829$ & $48$ \\
    \addlinespace[1pt]
    \FullFT & Forward & $8.372$ & $12.076$ & $0.784$ & $0.538$ & $37$ & $0.438$ & $29$ \\
     & Reverse & $4.086$ & $8.887$ & $0.860$ & $0.679$ & $41$ & $0.770$ & $47$ \\
     & Third & $4.059$ & $9.103$ & $0.834$ & $0.698$ & $41$ & $0.715$ & $42$ \\
    \addlinespace[1pt]
    \AdaFT & Forward & $11.165$ & $16.913$ & $0.722$ & $0.318$ & $22$ & $0.308$ & $20$ \\
     & Reverse & $3.762$ & $8.507$ & $0.851$ & $0.776$ & $45$ & $0.809$ & $48$ \\
     & Third & $3.732$ & $8.390$ & $0.901$ & $0.770$ & $46$ & $0.774$ & $47$ \\
    \addlinespace[1pt]
    \Replay & Forward & $3.842$ & $8.628$ & $0.864$ & $0.828$ & $49$ & $0.784$ & $48$ \\
     & Reverse & $3.765$ & $8.523$ & $0.847$ & $0.786$ & $46$ & $0.797$ & $48$ \\
     & Third & $3.728$ & $8.345$ & $0.868$ & $0.817$ & $48$ & $0.788$ & $47$ \\
    \addlinespace[1pt]
    \PEGP & Forward & $3.907$ & $8.597$ & $0.876$ & $0.832$ & $50$ & $0.798$ & $48$ \\
     & Reverse & $3.877$ & $8.775$ & $0.838$ & $0.821$ & $49$ & $0.839$ & $50$ \\
     & Third & $3.550$ & $8.233$ & $0.835$ & $0.780$ & $46$ & $0.797$ & $48$ \\
    \bottomrule
  \end{tabular}
\end{table}

\paragraph{Final scores.}
\Cref{tab:third-order-final} compares the three orders at the same training seed.
No third-order final checkpoint scores above \Czero in any closed-loop reading; two \SR readings tie with it (\DiffP \Replay under \NR and \PEGP under \IDMR, 48 of 56).
The third order ends on \gPickup, not on a \PlanTF pretraining group, yet the open-loop flip of \cref{sec:loop} recurs: \PlanTF \Replay and \PEGP improve on \Czero in cached ADE (3.100 and 3.464 \vs 3.561\,m) while scoring below it in all four closed-loop readings.
Cached ADE thus rates 8 of the 16 \PlanTF runs of the core suite and the third order better than \Czero, while every closed-loop reading places all 16 below it.
In this order, \PlanTF \PEGP is the highest-\CLS regime under both modes, and \Replay has the lowest cached ADE.
Final-checkpoint margins move with the order: \DiffP \Replay and \PEGP lead \AdaFT in \CLS-\NR by 0.046 and 0.009 (0.510 and 0.514 in the main run), whereas the lead of \PlanTF \PEGP over \AdaFT grows from 0.028 to 0.144.

\paragraph{The \gStationary update at stage~4.}
For \DiffP, the narrower margin is a final-checkpoint effect.
The \gStationary update collapses \AdaFT again: it raises cached ADE on the six other target groups by 9.90\,m (8.22\,m in the main run, where the group comes last, and 9.32\,m in reverse order, where it comes first; \cref{tab:E9}) and on the sentinel by 18.63\,m, and the seven-group macro from 3.92 to 12.59\,m.
\Replay and \PEGP limit the collateral to 0.26 and 0.25\,m and remove 97.4\% and 97.6\% of the macro shock (98.8\% and 97.1\% in the main run); \FullFT takes 3.98\,m.
The next update, on \gWaitPed, undoes the collapse: the macro of \AdaFT returns to 3.68\,m at \Tk{5}, and its final 3.732\,m is level with \Replay (3.728\,m).
Stage-averaged and worst-stage cached ADE still separate them (\AdaFT 4.952 and 12.593\,m, \Replay 3.740 and 4.094\,m, \PEGP 3.572 and 3.877\,m).
Whether the final checkpoint shows the protection of \Replay and \PEGP thus depends on where the damaging group sits in the stream; when it sits mid-stream, stage-averaged and worst-stage readings still register the protection, in line with the aggregation and stream axes (\cref{sec:aggregation,sec:order}).

\paragraph{Rankings across orders and seeds.}
Over the nine metric families of \cref{tab:E8}, the mean Kendall $\tau_b$ of final regime rankings is 0.65 (\PlanTF) and 0.29 (\DiffP) between the forward and third orders and 0.36 and 0.28 between the reverse and third orders, against 0.41 and $-0.03$ between the forward and reverse orders and 0.95 and 0.84 between the two forward-order seeds (descriptive).
All three order pairs thus move final rankings far more than the seeds; for the four closed-loop families of \DiffP, however, the reverse and third orders, neither of which ends on \gStationary, agree about as well (0.72) as the seeds (0.73).
Three orderings that hold in all three runs of the core suite hold in the third order too: \DiffP \Replay and \PEGP stay ahead of \FullFT in final \CLS under both modes, \DiffP \FullFT stays ahead of \AdaFT in stage-averaged cached ADE, and the cached ADE of \PlanTF ranks \Replay first and \FullFT last under all three summaries.

\paragraph{Acquisition and fixed-set backward transfer.}\phantomsection\label{app:third-order-stage}
The three orders end on different groups (\gStationary, \gStraight and \gPickup), so their six-group \BWT average different groups, and \BWTold covers only the two orders of the core suite.
\Cref{tab:third-order-stage} therefore measures backward transfer on the four groups that are old in every order, $G_{\cap}=\{\gFollow,\gHighLat,\gNearMulti,\gWaitPed\}$:
\begin{equation}
  \mathrm{BWT}_{\cap}^{(o)}=\frac{1}{4}\sum_{g\in G_{\cap}}\bigl[Q_o(7,g)-Q_o(p_o(g),g)\bigr],
  \label{eq:bwtcap}
\end{equation}
where $p_o(g)$ is the stage at which order $o$ learns group $g$ and $Q_o(k,g)$ the score of \Tk{k} on $g$, with displacement errors negated.
It fixes the groups, not the time since learning or the number of later updates, and differs from both the six-group \BWT and \BWTold.
For \DiffP \AdaFT, cached-ADE $\mathrm{BWT}_{\cap}$ is $-8.467$\,m in the forward and $-0.070$\,m in the third order, and \CLS-\NR $\mathrm{BWT}_{\cap}$ is $-0.487$ and $-0.024$: at the final checkpoint, these groups show almost no net forgetting in the third order, although cached ADE peaked at stage~4 (point estimates).

\begin{table}[t]
  \centering
  \caption{\textbf{Acquisition and fixed-set backward transfer in three orders} (training seed 20261014).
    \Acq: mean change of the seven groups across their own update; $\mathrm{BWT}_{\cap}$: final checkpoint minus learning endpoint on the four groups old in every order (\cref{eq:bwtcap}).
    $Q$ units (positive: better; cached $Q=-$ADE); cached ADE over 640 samples and \CLS over the 8 core-panel scenarios per group.
    Point estimates (descriptive).}
  \label{tab:third-order-stage}
  \scriptsize
  \setlength{\tabcolsep}{2pt}
  \begin{tabular}{@{}ll rrrrrr@{}}
    \toprule
    & & \multicolumn{2}{c}{Cached $Q$, m} & \multicolumn{2}{c}{\CLS-\NR} & \multicolumn{2}{c}{\CLS-\IDMR} \\
    \cmidrule(lr){3-4}\cmidrule(lr){5-6}\cmidrule(l){7-8}
    Regime & Order & \Acq & $\mathrm{BWT}_{\cap}$ & \Acq & $\mathrm{BWT}_{\cap}$ & \Acq & $\mathrm{BWT}_{\cap}$ \\
    \midrule
    \multicolumn{8}{@{}l}{\textit{\PlanTF}} \\
    \FullFT & Forward & $+0.413$ & $-2.079$ & $+0.075$ & $+0.004$ & $+0.024$ & $+0.049$ \\
     & Reverse & $+0.435$ & $+0.544$ & $-0.039$ & $+0.089$ & $-0.089$ & $+0.222$ \\
     & Third & $+0.933$ & $-0.481$ & $-0.034$ & $+0.014$ & $-0.023$ & $-0.177$ \\
    \addlinespace[1pt]
    \AdaFT & Forward & $+0.090$ & $-1.758$ & $-0.117$ & $+0.013$ & $-0.013$ & $+0.168$ \\
     & Reverse & $+0.553$ & $+0.026$ & $-0.112$ & $+0.009$ & $-0.028$ & $-0.112$ \\
     & Third & $+0.652$ & $-0.804$ & $+0.086$ & $+0.003$ & $-0.101$ & $-0.038$ \\
    \addlinespace[1pt]
    \Replay & Forward & $+0.320$ & $+0.051$ & $-0.041$ & $+0.013$ & $+0.043$ & $+0.089$ \\
     & Reverse & $+0.451$ & $+0.096$ & $-0.003$ & $-0.001$ & $-0.031$ & $+0.036$ \\
     & Third & $+0.561$ & $+0.048$ & $+0.008$ & $+0.015$ & $+0.035$ & $+0.056$ \\
    \addlinespace[1pt]
    \PEGP & Forward & $+0.197$ & $-1.088$ & $+0.022$ & $-0.043$ & $+0.051$ & $+0.028$ \\
     & Reverse & $+0.428$ & $+0.158$ & $-0.002$ & $-0.069$ & $-0.101$ & $-0.048$ \\
     & Third & $+0.575$ & $-0.395$ & $+0.056$ & $+0.127$ & $-0.010$ & $+0.080$ \\
    \midrule
    \multicolumn{8}{@{}l}{\textit{\DiffP}} \\
    \FullFT & Forward & $-0.091$ & $-5.659$ & $+0.029$ & $-0.159$ & $+0.012$ & $-0.282$ \\
     & Reverse & $+0.532$ & $-0.633$ & $-0.038$ & $-0.122$ & $+0.013$ & $+0.067$ \\
     & Third & $+0.285$ & $-0.625$ & $+0.073$ & $-0.121$ & $+0.018$ & $-0.078$ \\
    \addlinespace[1pt]
    \AdaFT & Forward & $-0.301$ & $-8.467$ & $+0.007$ & $-0.487$ & $-0.017$ & $-0.442$ \\
     & Reverse & $+0.662$ & $-0.184$ & $+0.003$ & $-0.023$ & $+0.015$ & $+0.086$ \\
     & Third & $+0.510$ & $-0.070$ & $+0.002$ & $-0.024$ & $+0.024$ & $-0.014$ \\
    \addlinespace[1pt]
    \Replay & Forward & $-0.149$ & $-0.119$ & $+0.047$ & $+0.024$ & $+0.022$ & $+0.033$ \\
     & Reverse & $+0.598$ & $-0.098$ & $-0.010$ & $-0.022$ & $-0.009$ & $+0.025$ \\
     & Third & $-0.254$ & $-0.078$ & $-0.040$ & $+0.034$ & $+0.007$ & $+0.031$ \\
    \addlinespace[1pt]
    \PEGP & Forward & $-0.138$ & $-0.213$ & $+0.042$ & $+0.019$ & $+0.027$ & $+0.004$ \\
     & Reverse & $+0.539$ & $-0.001$ & $-0.018$ & $+0.006$ & $-0.026$ & $+0.003$ \\
     & Third & $-0.142$ & $-0.060$ & $+0.015$ & $-0.014$ & $-0.029$ & $+0.008$ \\
    \bottomrule
  \end{tabular}
\end{table}

\subsection{Joint references}
\label{app:joint}

\paragraph{Design.}
Each planner is fine-tuned from its own \Czero with \FullFT and with \AdaFT on the union of the seven training groups (413,939 samples), shuffled anew in every epoch, without group-balanced sampling, replay or projection.
The optimizer, peak learning rate and effective batch are those of the sequential runs (\cref{sec:budget}), but one warm-up and cosine schedule spans the whole run, without restarts.
Twenty epochs over the union give the same 8,278,780 current-sample visits as sequential training, with 32,340 instead of 32,420 updates, and we keep the last epoch without validation-based selection.
As the schedule and the update count change along with the data order, the joint runs are non-sequential references, neither order-only controls nor upper bounds.
We evaluate \Czero and the final checkpoint on the cached development samples and on the core panel under both modes; without learning endpoints, \Acq and \BWT are undefined.

\begin{table}[t]
  \centering
  \caption{\textbf{Joint references and sequential main runs from the same \Czero.}
    Joint: final checkpoint of training on the shuffled union of the seven groups; main run: the sequential forward-order run.
    Columns as in \cref{tab:third-order-final}; point estimates (descriptive).}
  \label{tab:joint}
  \scriptsize
  \setlength{\tabcolsep}{2pt}
  \begin{tabular}{@{}ll rrrrrrr@{}}
    \toprule
    & & \multicolumn{2}{c}{Cached, m $\downarrow$} & \multicolumn{3}{c}{\NR $\uparrow$} & \multicolumn{2}{c}{\IDMR $\uparrow$} \\
    \cmidrule(lr){3-4}\cmidrule(lr){5-7}\cmidrule(l){8-9}
    Regime & Training & ADE & FDE & \PDMS & \CLS & \SR & \CLS & \SR \\
    \midrule
    \multicolumn{9}{@{}l}{\textit{\PlanTF}} \\
    \Czero & & $3.561$ & $7.680$ & $0.760$ & $0.528$ & $32$ & $0.557$ & $34$ \\
    \addlinespace[1pt]
    \FullFT & Main run & $5.489$ & $10.980$ & $0.675$ & $0.258$ & $17$ & $0.310$ & $20$ \\
     & Joint & $3.233$ & $7.550$ & $0.834$ & $0.480$ & $28$ & $0.506$ & $31$ \\
    \addlinespace[1pt]
    \AdaFT & Main run & $4.842$ & $9.431$ & $0.772$ & $0.378$ & $24$ & $0.501$ & $33$ \\
     & Joint & $2.946$ & $6.788$ & $0.815$ & $0.459$ & $28$ & $0.524$ & $33$ \\
    \midrule
    \multicolumn{9}{@{}l}{\textit{\DiffP}} \\
    \Czero & & $2.978$ & $6.721$ & $0.831$ & $0.823$ & $48$ & $0.829$ & $48$ \\
    \addlinespace[1pt]
    \FullFT & Main run & $8.372$ & $12.076$ & $0.784$ & $0.538$ & $37$ & $0.438$ & $29$ \\
     & Joint & $3.870$ & $8.451$ & $0.888$ & $0.742$ & $45$ & $0.817$ & $49$ \\
    \addlinespace[1pt]
    \AdaFT & Main run & $11.165$ & $16.913$ & $0.722$ & $0.318$ & $22$ & $0.308$ & $20$ \\
     & Joint & $3.814$ & $8.710$ & $0.847$ & $0.822$ & $48$ & $0.784$ & $47$ \\
    \bottomrule
  \end{tabular}
\end{table}

\paragraph{Results.}
\Cref{tab:joint} sets the joint references against \Czero and the forward-order main runs.
In final \CLS-\NR, joint training ends above the sequential main run in all four pairs, by 0.081 to 0.504, and recovers between about half (\PlanTF \AdaFT) and nearly all (\DiffP \AdaFT) of that run's loss against \Czero.
Yet no joint reference exceeds its \Czero in \CLS under either mode: \DiffP Joint \AdaFT comes closest (\CLS-\NR 0.822 \vs 0.823, \SR-\NR 48 \vs 48 of 56), and the only closed-loop reading above \Czero is the \SR-\IDMR of \DiffP Joint \FullFT, a net gain of one target (49 \vs 48; one lost, two gained).
Per mode, each joint reference still breaks 1 to 11 of the targets that \Czero passes and fixes 2 to 7 of those it fails.
Sequential training thus explains much of the regression but not the absence of a gain (point estimates).
Open loop again disagrees with closed loop: for \PlanTF, joint training improves cached ADE (3.233 and 2.946 \vs 3.561\,m) and \PDMS (0.834 and 0.815 \vs 0.760) over \Czero while every closed-loop reading stays below it, the loop flip of \cref{sec:loop} without a sequential stream; for \DiffP, it worsens cached ADE (3.870 and 3.814 \vs 2.978\,m) but improves \PDMS (0.888 and 0.847 \vs 0.831).
}{}
\IfFileExists{sec/app_e13_equal_data.tex}{
\subsection{Equal-data runs}
\label{app:equal-data}

\paragraph{Design.}
The equal-data runs repeat the four regimes for both planners in the forward order with 10,000 training samples per group (70,000 in total), drawn without replacement from each group's full pool with fixed per-group seeds derived from 20261110; both planners and all regimes train on the same samples.
\Czero, the development samples, the recipe (20 epochs per group, restarts at every group) and training seed 20261014 are those of the main run, so each group receives 800 updates (5,600 per run, instead of 900 to 7,820 per group and 32,420 per run), and \Replay adds 300,000 replay visits (21.4\% of the 1,400,000 current-sample visits).
The runs change the size of every group and the total budget together, so they are not a single-factor control.
Closed loop covers \Czero, \Tk{7} and each group before and after its stage under both modes.

\paragraph{Results.}
\Cref{tab:equal-data-final,tab:equal-data-stage} set the equal-data runs against the main runs.
With less data, the unprotected \DiffP regimes collapse less (\CLS-\NR: \AdaFT 0.528 \vs 0.318, \FullFT 0.570 \vs 0.538), the protected ones end lower (\Replay 0.723 \vs 0.828, \PEGP 0.795 \vs 0.832), and \PlanTF \FullFT ends higher (0.343 \vs 0.258).
All eight runs end below \Czero in \CLS-\NR; under \IDMR, \PlanTF \Replay ends slightly above \Czero in \CLS (0.567 \vs 0.557) and \DiffP \PEGP passes one more target (49 \vs 48 of 56).

\begin{table}[t]
  \centering
  \caption{\textbf{Final scores with full pools and with 10,000 samples per group} (forward order, training seed 20261014; full pool: the main run).
    Columns as in \cref{tab:third-order-final}; point estimates (descriptive).}
  \label{tab:equal-data-final}
  \scriptsize
  \setlength{\tabcolsep}{1.8pt}
  \begin{tabular}{@{}ll rrrrrrr@{}}
    \toprule
    & & \multicolumn{2}{c}{Cached, m $\downarrow$} & \multicolumn{3}{c}{\NR $\uparrow$} & \multicolumn{2}{c}{\IDMR $\uparrow$} \\
    \cmidrule(lr){3-4}\cmidrule(lr){5-7}\cmidrule(l){8-9}
    Regime & Data & ADE & FDE & \PDMS & \CLS & \SR & \CLS & \SR \\
    \midrule
    \multicolumn{9}{@{}l}{\textit{\PlanTF}} \\
    \Czero & & $3.561$ & $7.680$ & $0.760$ & $0.528$ & $32$ & $0.557$ & $34$ \\
    \addlinespace[1pt]
    \FullFT & Full pool & $5.489$ & $10.980$ & $0.675$ & $0.258$ & $17$ & $0.310$ & $20$ \\
     & 10k/group & $3.849$ & $8.057$ & $0.678$ & $0.343$ & $22$ & $0.381$ & $25$ \\
    \addlinespace[1pt]
    \AdaFT & Full pool & $4.842$ & $9.431$ & $0.772$ & $0.378$ & $24$ & $0.501$ & $33$ \\
     & 10k/group & $4.411$ & $8.985$ & $0.794$ & $0.370$ & $25$ & $0.398$ & $27$ \\
    \addlinespace[1pt]
    \Replay & Full pool & $3.034$ & $7.010$ & $0.788$ & $0.454$ & $28$ & $0.549$ & $33$ \\
     & 10k/group & $3.074$ & $7.047$ & $0.844$ & $0.426$ & $26$ & $0.567$ & $34$ \\
    \addlinespace[1pt]
    \PEGP & Full pool & $4.208$ & $8.888$ & $0.765$ & $0.406$ & $26$ & $0.508$ & $32$ \\
     & 10k/group & $3.395$ & $7.713$ & $0.782$ & $0.328$ & $21$ & $0.424$ & $27$ \\
    \midrule
    \multicolumn{9}{@{}l}{\textit{\DiffP}} \\
    \Czero & & $2.978$ & $6.721$ & $0.831$ & $0.823$ & $48$ & $0.829$ & $48$ \\
    \addlinespace[1pt]
    \FullFT & Full pool & $8.372$ & $12.076$ & $0.784$ & $0.538$ & $37$ & $0.438$ & $29$ \\
     & 10k/group & $6.772$ & $10.368$ & $0.792$ & $0.570$ & $38$ & $0.568$ & $37$ \\
    \addlinespace[1pt]
    \AdaFT & Full pool & $11.165$ & $16.913$ & $0.722$ & $0.318$ & $22$ & $0.308$ & $20$ \\
     & 10k/group & $7.598$ & $12.813$ & $0.812$ & $0.528$ & $36$ & $0.522$ & $34$ \\
    \addlinespace[1pt]
    \Replay & Full pool & $3.842$ & $8.628$ & $0.864$ & $0.828$ & $49$ & $0.784$ & $48$ \\
     & 10k/group & $3.971$ & $8.873$ & $0.880$ & $0.723$ & $44$ & $0.751$ & $46$ \\
    \addlinespace[1pt]
    \PEGP & Full pool & $3.907$ & $8.597$ & $0.876$ & $0.832$ & $50$ & $0.798$ & $48$ \\
     & 10k/group & $3.951$ & $8.772$ & $0.864$ & $0.795$ & $47$ & $0.808$ & $49$ \\
    \bottomrule
  \end{tabular}
\end{table}

\begin{table}[t]
  \centering
  \caption{\textbf{Acquisition and backward transfer with full pools and with 10,000 samples per group.}
    \Acq as in \cref{tab:third-order-stage}; \BWT over the six old groups of the forward order; $Q$ units as there.
    Point estimates (descriptive).}
  \label{tab:equal-data-stage}
  \scriptsize
  \setlength{\tabcolsep}{2pt}
  \begin{tabular}{@{}ll rrrrrr@{}}
    \toprule
    & & \multicolumn{2}{c}{Cached $Q$, m} & \multicolumn{2}{c}{\CLS-\NR} & \multicolumn{2}{c}{\CLS-\IDMR} \\
    \cmidrule(lr){3-4}\cmidrule(lr){5-6}\cmidrule(l){7-8}
    Regime & Data & \Acq & \BWT & \Acq & \BWT & \Acq & \BWT \\
    \midrule
    \multicolumn{8}{@{}l}{\textit{\PlanTF}} \\
    \FullFT & Full pool & $+0.413$ & $-2.083$ & $+0.075$ & $-0.042$ & $+0.024$ & $-0.047$ \\
     & 10k/group & $+0.320$ & $-0.447$ & $+0.107$ & $-0.081$ & $+0.068$ & $-0.060$ \\
    \addlinespace[1pt]
    \AdaFT & Full pool & $+0.090$ & $-1.761$ & $-0.117$ & $+0.025$ & $-0.013$ & $+0.091$ \\
     & 10k/group & $+0.235$ & $-1.425$ & $-0.014$ & $-0.084$ & $-0.058$ & $+0.011$ \\
    \addlinespace[1pt]
    \Replay & Full pool & $+0.320$ & $+0.017$ & $-0.041$ & $+0.028$ & $+0.043$ & $+0.098$ \\
     & 10k/group & $+0.320$ & $-0.065$ & $-0.041$ & $+0.006$ & $-0.054$ & $+0.115$ \\
    \addlinespace[1pt]
    \PEGP & Full pool & $+0.197$ & $-1.165$ & $+0.022$ & $-0.042$ & $+0.051$ & $+0.012$ \\
     & 10k/group & $+0.216$ & $-0.171$ & $+0.014$ & $-0.088$ & $-0.067$ & $-0.037$ \\
    \midrule
    \multicolumn{8}{@{}l}{\textit{\DiffP}} \\
    \FullFT & Full pool & $-0.091$ & $-5.312$ & $+0.029$ & $-0.226$ & $+0.012$ & $-0.320$ \\
     & 10k/group & $-0.050$ & $-3.502$ & $+0.023$ & $-0.239$ & $-0.063$ & $-0.205$ \\
    \addlinespace[1pt]
    \AdaFT & Full pool & $-0.301$ & $-8.563$ & $+0.007$ & $-0.517$ & $-0.017$ & $-0.489$ \\
     & 10k/group & $-0.337$ & $-4.333$ & $-0.088$ & $-0.208$ & $-0.063$ & $-0.264$ \\
    \addlinespace[1pt]
    \Replay & Full pool & $-0.149$ & $-0.196$ & $+0.047$ & $-0.004$ & $+0.022$ & $+0.012$ \\
     & 10k/group & $-0.195$ & $-0.227$ & $+0.004$ & $-0.070$ & $-0.051$ & $+0.006$ \\
    \addlinespace[1pt]
    \PEGP & Full pool & $-0.138$ & $-0.254$ & $+0.042$ & $+0.007$ & $+0.027$ & $+0.001$ \\
     & 10k/group & $-0.151$ & $-0.306$ & $-0.022$ & $-0.001$ & $-0.014$ & $+0.012$ \\
    \bottomrule
  \end{tabular}
\end{table}

\paragraph{The \gStationary update.}
With 800 instead of 7,820 updates, the \gStationary update still dominates the collateral of \DiffP \AdaFT: it raises cached ADE on the six other target groups by 3.938\,m, about half the full-pool 8.220\,m (\cref{tab:E9}) but about seven times the largest other update of that run (0.551\,m, \gStraight).
\FullFT takes 3.135\,m (full pool: 4.704\,m), and \Replay and \PEGP again neutralize the shock: their cached ADE on the other groups changes by $-0.041$ and $+0.049$\,m (full pool: $+0.104$ and $+0.246$\,m).
On \gStationary itself, the cached ADE of \AdaFT rises from 1.061 to 1.701\,m, whereas \Replay and \PEGP lower it (from 1.054 to 1.010\,m and from 1.059 to 1.031\,m).
As all groups have the same size here, the damage follows the content of the group, not only its size; the total budget, however, changes as well (point estimates).
}{}
\IfFileExists{sec/app_e3_panel.tex}{
\subsection{Expanded panel}
\label{app:expanded-endpoints}

\paragraph{Design.}
The expanded panel re-evaluates \Czero and the final checkpoints of the eight main runs on 720 twenty-second scenarios, 90 per group: the 64 core-panel scenarios and 82 more per group, drawn at random from the same development pool with selection seed 20261111 (271 logs; \cref{app:data-panel}).
Scores average the seven target groups equally (630 targets); the 90 sentinel scenarios are not included.
As both panels come from the development split, the expanded panel is larger than the core panel but not independent of it.
\ADEs, \FDEs and \PDMS score the initial plan of each scenario under \NR (\cref{sec:evaluation}), not the cached samples, and \CLS and \SR use the simulation settings of the core panel.
Intermediate checkpoints are not evaluated, so the panel supports final-checkpoint readings only.

\begin{table}[t]
  \centering
  \caption{\textbf{Final checkpoints of the main runs on the expanded panel} (630 targets; forward order, training seed 20261014).
    \ADEs and \FDEs (m) and \PDMS score the initial plan; \SR as passes of 630; the sentinel is not included.
    Point estimates (descriptive).}
  \label{tab:expanded-endpoints}
  \scriptsize
  \setlength{\tabcolsep}{2pt}
  \begin{tabular}{@{}l rrrrrrr@{}}
    \toprule
    & \multicolumn{3}{c}{Initial plan, \NR} & \multicolumn{2}{c}{\NR $\uparrow$} & \multicolumn{2}{c}{\IDMR $\uparrow$} \\
    \cmidrule(lr){2-4}\cmidrule(lr){5-6}\cmidrule(l){7-8}
    Regime & \ADEs $\downarrow$ & \FDEs $\downarrow$ & \PDMS $\uparrow$ & \CLS & \SR & \CLS & \SR \\
    \midrule
    \multicolumn{8}{@{}l}{\textit{\PlanTF}} \\
    \Czero & $3.735$ & $7.854$ & $0.811$ & $0.551$ & $384$ & $0.553$ & $383$ \\
    \FullFT & $5.227$ & $10.270$ & $0.735$ & $0.310$ & $239$ & $0.286$ & $211$ \\
    \AdaFT & $4.774$ & $9.281$ & $0.802$ & $0.434$ & $329$ & $0.401$ & $305$ \\
    \Replay & $2.904$ & $6.675$ & $0.824$ & $0.496$ & $349$ & $0.497$ & $350$ \\
    \PEGP & $4.233$ & $8.738$ & $0.822$ & $0.482$ & $348$ & $0.460$ & $327$ \\
    \midrule
    \multicolumn{8}{@{}l}{\textit{\DiffP}} \\
    \Czero & $3.078$ & $6.936$ & $0.886$ & $0.851$ & $558$ & $0.824$ & $551$ \\
    \FullFT & $8.301$ & $12.047$ & $0.792$ & $0.538$ & $424$ & $0.384$ & $296$ \\
    \AdaFT & $11.186$ & $17.088$ & $0.732$ & $0.330$ & $253$ & $0.294$ & $217$ \\
    \Replay & $3.902$ & $8.866$ & $0.874$ & $0.820$ & $547$ & $0.784$ & $540$ \\
    \PEGP & $3.923$ & $8.701$ & $0.881$ & $0.829$ & $555$ & $0.791$ & $542$ \\
    \bottomrule
  \end{tabular}
\end{table}

\paragraph{Results.}
On 630 targets, every main-run final checkpoint scores below \Czero in \CLS and \SR under both modes (\cref{tab:expanded-endpoints}), with \CLS-\NR changes from $-0.021$ (\DiffP \PEGP) to $-0.521$ (\DiffP \AdaFT).
The regime rankings of the core panel recur in \CLS under both modes and in \SR under \NR.
\DiffP \Replay and \PEGP, which show no detectable change on the core panel ($+0.005$ and $+0.009$ in \CLS-\NR; \cref{tab:E1}), end 0.031 and 0.021 below \Czero here.
The open-loop flip of \cref{sec:loop} recurs as well: \PlanTF \Replay improves on \Czero in \ADEs (2.904 \vs 3.735\,m) and \PDMS (0.824 \vs 0.811) but not in \CLS-\NR (0.496 \vs 0.551).
All values are point estimates.
}{}
\IfFileExists{sec/app_e5_lora.tex}{
\subsection{LoRA branch}
\label{app:lora}

\paragraph{Design.}
The LoRA branch replaces the adapters by low-rank updates of attention projections~\cite{hu2022lora} and trains three update regimes, \LoRAFT, \LoRAReplay and \LoRAPEGP, for both planners in the forward, reverse and third orders: 18 runs with the deployed checkpoints, full training pools, development samples, recipe and training seed 20261014 of the core suite.
Seventeen runs take 32,420 updates at the global batch of 256.
The \PlanTF \LoRAReplay run in reverse order uses a global batch of 2,048 and takes 4,100 updates, with the same 20 epochs per group, learning-rate schedule and 8,278,780 current-sample visits; as it is not update-matched, we list it separately.
Cached evaluation covers \Czero--\Tk{7}; closed loop covers \Czero, \Tk{7} and each group before and after its stage under both modes.

\paragraph{Placement and trainable parameters.}
Each query and value projection of the selected attention modules receives $\Delta W=(\alpha/r)BA$ with rank $r=16$, scale $\alpha=16$ and dropout 0; $A$ is initialized Kaiming-uniform and $B$ with zeros.
One set of factors is trained through the whole stream, without task-specific modules.
The backbone, the key and output projections, biases and BatchNorm statistics stay frozen, and inference uses merged weights.
\PlanTF carries 22 such projections in 11 attention modules (six history-attention blocks, one ego-state attention module and four encoder blocks; 69,632 trainable parameters), \DiffP 18 in nine modules (three encoder-fusion, three decoder self-attention and three decoder cross-attention modules; 110,592 parameters), against 32,768 and 36,864 adapter parameters.
The two branches thus differ in placement and size and are not parameter-matched.

\paragraph{Regimes.}
\LoRAFT trains the factors without replay or projection.
\LoRAReplay follows the replay rule of \Replay (512 stored samples per completed group; from stage~2, replay batches of a quarter of the current micro-batch with loss weight 0.25) and adds the same replay visits as \Replay in the same order (19.0\% of current-sample visits in the forward and reverse orders, 24.3\% in the third).
\LoRAPEGP applies the projection of \PEGP (\cref{sec:regimes}) to the gradients of both factors from stage~2, before gradient clipping and AdamW.
It collects the inputs $x$ of $A$ and $Ax$ of $B$ on up to 1,024 current-group samples (centered float64 covariance, energy threshold 0.9, cumulative rank cap 16 per factor) and merges each new basis with the stored directions; a full-rank protected input space gives an exactly zero projected gradient.
Like \PEGP, it projects gradients rather than optimizer updates, stores no \Czero basis and omits the official frozen-adapter warm-up, feature alignment and weight fusion; it does not reuse the adapter bases.
As in the core suite, no EMA is maintained.

\begin{table}[t]
  \centering
  \caption{\textbf{Final scores of the LoRA branch} (training seed 20261014).
    Columns as in \cref{tab:third-order-final}; \Czero as there.
    The \PlanTF \LoRAReplay run in reverse order uses a global batch of 2,048 and is listed separately.
    Point estimates (descriptive).}
  \label{tab:lora-final}
  \scriptsize
  \setlength{\tabcolsep}{2pt}
  \begin{tabular}{@{}ll rrrrrrr@{}}
    \toprule
    & & \multicolumn{2}{c}{Cached, m $\downarrow$} & \multicolumn{3}{c}{\NR $\uparrow$} & \multicolumn{2}{c}{\IDMR $\uparrow$} \\
    \cmidrule(lr){3-4}\cmidrule(lr){5-7}\cmidrule(l){8-9}
    Regime & Order & ADE & FDE & \PDMS & \CLS & \SR & \CLS & \SR \\
    \midrule
    \multicolumn{9}{@{}l}{\textit{\PlanTF}} \\
    \Czero & & $3.561$ & $7.680$ & $0.760$ & $0.528$ & $32$ & $0.557$ & $34$ \\
    \addlinespace[1pt]
    \LoRAFT & Forward & $4.241$ & $8.725$ & $0.826$ & $0.415$ & $27$ & $0.474$ & $31$ \\
     & Reverse & $3.283$ & $7.270$ & $0.797$ & $0.550$ & $33$ & $0.510$ & $32$ \\
     & Third & $3.609$ & $8.915$ & $0.778$ & $0.532$ & $33$ & $0.493$ & $32$ \\
    \addlinespace[1pt]
    \LoRAReplay & Forward & $3.091$ & $7.099$ & $0.812$ & $0.472$ & $29$ & $0.542$ & $34$ \\
     & Third & $3.229$ & $7.772$ & $0.796$ & $0.505$ & $30$ & $0.593$ & $36$ \\
    \addlinespace[1pt]
    \LoRAPEGP & Forward & $4.787$ & $9.927$ & $0.798$ & $0.416$ & $28$ & $0.486$ & $32$ \\
     & Reverse & $3.437$ & $7.508$ & $0.759$ & $0.541$ & $33$ & $0.542$ & $34$ \\
     & Third & $3.378$ & $7.726$ & $0.790$ & $0.449$ & $27$ & $0.474$ & $29$ \\
    \midrule
    \multicolumn{9}{@{}l}{\textit{\DiffP}} \\
    \Czero & & $2.978$ & $6.721$ & $0.831$ & $0.823$ & $48$ & $0.829$ & $48$ \\
    \addlinespace[1pt]
    \LoRAFT & Forward & $6.980$ & $10.778$ & $0.839$ & $0.527$ & $35$ & $0.531$ & $35$ \\
     & Reverse & $3.658$ & $8.263$ & $0.887$ & $0.799$ & $47$ & $0.848$ & $51$ \\
     & Third & $3.754$ & $8.474$ & $0.852$ & $0.824$ & $48$ & $0.800$ & $48$ \\
    \addlinespace[1pt]
    \LoRAReplay & Forward & $4.036$ & $8.973$ & $0.833$ & $0.762$ & $45$ & $0.743$ & $45$ \\
     & Reverse & $3.740$ & $8.402$ & $0.886$ & $0.812$ & $48$ & $0.829$ & $50$ \\
     & Third & $3.747$ & $8.510$ & $0.850$ & $0.806$ & $47$ & $0.813$ & $49$ \\
    \addlinespace[1pt]
    \LoRAPEGP & Forward & $4.113$ & $8.627$ & $0.875$ & $0.762$ & $46$ & $0.768$ & $48$ \\
     & Reverse & $3.651$ & $8.269$ & $0.870$ & $0.780$ & $46$ & $0.767$ & $46$ \\
     & Third & $3.836$ & $8.649$ & $0.869$ & $0.803$ & $47$ & $0.804$ & $48$ \\
    \midrule
    \multicolumn{9}{@{}l}{\textit{\PlanTF, global batch 2,048 (4,100 updates)}} \\
    \LoRAReplay & Reverse & $2.951$ & $6.820$ & $0.783$ & $0.506$ & $31$ & $0.547$ & $33$ \\
    \bottomrule
  \end{tabular}
\end{table}

\begin{table}[t]
  \centering
  \caption{\textbf{Acquisition and backward transfer of the LoRA branch.}
    \Acq as in \cref{tab:third-order-stage}; \BWT over the six old groups of each order, so its groups change with the order; $Q$ units as there.
    Point estimates (descriptive).}
  \label{tab:lora-stage}
  \scriptsize
  \setlength{\tabcolsep}{1.6pt}
  \begin{tabular}{@{}ll rrrrrr@{}}
    \toprule
    & & \multicolumn{2}{c}{Cached $Q$, m} & \multicolumn{2}{c}{\CLS-\NR} & \multicolumn{2}{c}{\CLS-\IDMR} \\
    \cmidrule(lr){3-4}\cmidrule(lr){5-6}\cmidrule(l){7-8}
    Regime & Order & \Acq & \BWT & \Acq & \BWT & \Acq & \BWT \\
    \midrule
    \multicolumn{8}{@{}l}{\textit{\PlanTF}} \\
    \LoRAFT & Forward & $+0.278$ & $-1.268$ & $-0.024$ & $-0.027$ & $-0.062$ & $+0.085$ \\
     & Reverse & $+0.322$ & $-0.135$ & $-0.050$ & $+0.140$ & $-0.058$ & $+0.076$ \\
     & Third & $+0.413$ & $-0.384$ & $+0.006$ & $+0.169$ & $-0.093$ & $+0.114$ \\
    \addlinespace[1pt]
    \LoRAReplay & Forward & $+0.382$ & $-0.069$ & $-0.040$ & $+0.041$ & $-0.020$ & $+0.022$ \\
     & Third & $+0.448$ & $-0.002$ & $+0.015$ & $+0.076$ & $+0.016$ & $+0.096$ \\
    \addlinespace[1pt]
    \LoRAPEGP & Forward & $+0.274$ & $-1.961$ & $-0.070$ & $+0.036$ & $-0.029$ & $-0.044$ \\
     & Reverse & $+0.078$ & $-0.182$ & $-0.009$ & $+0.093$ & $+0.008$ & $+0.058$ \\
     & Third & $+0.395$ & $-0.303$ & $-0.003$ & $-0.025$ & $+0.029$ & $-0.062$ \\
    \midrule
    \multicolumn{8}{@{}l}{\textit{\DiffP}} \\
    \LoRAFT & Forward & $-0.152$ & $-3.882$ & $+0.006$ & $-0.273$ & $-0.059$ & $-0.260$ \\
     & Reverse & $+0.327$ & $-0.037$ & $-0.001$ & $+0.038$ & $-0.006$ & $+0.107$ \\
     & Third & $+0.126$ & $-0.055$ & $+0.015$ & $+0.075$ & $-0.014$ & $+0.043$ \\
    \addlinespace[1pt]
    \LoRAReplay & Forward & $-0.084$ & $-0.481$ & $+0.053$ & $-0.065$ & $-0.041$ & $-0.003$ \\
     & Reverse & $+0.338$ & $-0.105$ & $+0.036$ & $+0.023$ & $-0.002$ & $+0.067$ \\
     & Third & $-0.160$ & $-0.100$ & $+0.007$ & $+0.006$ & $-0.046$ & $+0.070$ \\
    \addlinespace[1pt]
    \LoRAPEGP & Forward & $-0.056$ & $-0.699$ & $+0.015$ & $-0.024$ & $-0.043$ & $-0.036$ \\
     & Reverse & $+0.351$ & $+0.212$ & $-0.009$ & $+0.036$ & $+0.016$ & $-0.036$ \\
     & Third & $-0.101$ & $-0.309$ & $-0.041$ & $+0.042$ & $-0.016$ & $+0.008$ \\
    \midrule
    \multicolumn{8}{@{}l}{\textit{\PlanTF, global batch 2,048 (4,100 updates)}} \\
    \LoRAReplay & Reverse & $+0.307$ & $+0.107$ & $-0.017$ & $+0.028$ & $+0.024$ & $-0.003$ \\
    \bottomrule
  \end{tabular}
\end{table}

\paragraph{Forward order.}
For \DiffP, LoRA reproduces the adapter pattern with a smaller collapse (\cref{tab:lora-final,tab:lora-stage}).
\LoRAFT ends at \CLS-\NR 0.527 and cached ADE 6.980\,m (\AdaFT: 0.318 and 11.165\,m; \Czero: 0.823 and 2.978\,m), and \LoRAReplay and \LoRAPEGP both end at 0.762: 0.235 above \LoRAFT, about half the adapter margins (0.510 and 0.514), and below \Replay and \PEGP (0.828 and 0.832).
\CLS-\NR \BWT rises from \LoRAFT to \LoRAReplay and \LoRAPEGP, from $-0.273$ to $-0.065$ and $-0.024$ for \DiffP and from $-0.027$ to $+0.041$ and $+0.036$ for \PlanTF.
For \PlanTF, the LoRA and adapter runs end within 0.04 of each other in \CLS-\NR, and cached ADE does not follow this \BWT: \LoRAPEGP forgets more than \LoRAFT ($-1.961$ \vs $-1.268$\,m) and ends with a higher error (4.787 \vs 4.241\,m).

\paragraph{The \gStationary update.}
For \DiffP, the \gStationary update raises the cached ADE of \LoRAFT on the six other target groups by 3.46, 4.46 and 4.27\,m in the forward, reverse and third orders, 0.74 to 1.07 times the collateral of \FullFT in the same order, against 1.75 to 2.49 times for \AdaFT.
LoRA thus collapses less than the adapters, but parameter efficiency alone protects neither branch.
Protection removes less of the shock in the LoRA branch: \LoRAReplay and \LoRAPEGP remove 95\% and 82\% of the macro shock of \LoRAFT when \gStationary comes last and 93\% and 67\% when it comes fourth, against 97\% to 99\% for \Replay and \PEGP in the main run and the third order.

\paragraph{Reverse and third orders.}
When \gStationary does not come last, the protected regimes show no consistent closed-loop gain over \LoRAFT: only \DiffP \LoRAReplay in reverse order ends above \LoRAFT in \CLS-\NR (0.812 \vs 0.799), and the final \CLS-\NR of all these runs lies between $-0.079$ and $+0.022$ of \Czero.

\paragraph{Against \Czero.}
Several final checkpoints of the LoRA branch end slightly above \Czero in at least one closed-loop reading, among them four \PlanTF runs, whereas every \PlanTF run of the core suite ends below \Czero in all closed-loop readings.
In \CLS-\NR, four runs end at or above \Czero, by at most $+0.022$ (\PlanTF \LoRAFT, reverse order); under \IDMR, \CLS reaches 0.593 against 0.557 (\PlanTF \LoRAReplay, third order) and \SR 51 against 48 of 56 targets (\DiffP \LoRAFT, reverse order).
These differences lie within the resolution of the core panel (\cref{sec:deploy}), and we do not extend the deployment invariant of the core suite to the LoRA branch.
All values are point estimates.
}{}
\IfFileExists{sec/app_f_regimes.tex}{
\section{Update-Regime Mechanisms}
\label{app:regimes}
\providecommand*{\theHtable}{}\providecommand*{\theHfigure}{}%
\setcounter{table}{0}\setcounter{figure}{0}
\renewcommand{\thetable}{\thesection\arabic{table}}\renewcommand{\thefigure}{\thesection\arabic{figure}}
\renewcommand{\theHtable}{\thesection.\arabic{table}}\renewcommand{\theHfigure}{\thesection.\arabic{figure}}

This section details the four core update regimes \FullFT, \AdaFT, \Replay and \PEGP of \cref{sec:regimes} (\cref{tab:F1}), shows how \PEGP's projection saturates and leaks, and summarizes the \DiffP ablations that test where its protection comes from (\cref{tab:F2,fig:F2}).
The LoRA branch (LoRA-FT, LoRA+Replay and LoRA+PEGP) is specified in \cref{app:lora}; the saturation, leak and ablation results below concern the adapter regimes only.
Statements are descriptive unless an interval is given.

\begin{table}[t]
  \centering
  \caption{\textbf{Update-regime implementation.}
    The three adapter regimes of a planner share its adapter parameter space; \FullFT inserts no adapters and updates all planner weights.
    Percentages: share of the \FullFT parameters.
    \PlanTF's neighbor-prediction loss is zero for a batch without a valid neighbor future target.}
  \label{tab:F1}
  \scriptsize
  \setlength{\tabcolsep}{2.5pt}
  \begin{tabular}{@{}>{\raggedright\arraybackslash}p{0.80in}>{\raggedright\arraybackslash}p{1.15in}>{\raggedright\arraybackslash}p{1.15in}@{}}
    \toprule
    & \PlanTF & \DiffP \\
    \midrule
    Adapter sites & 8: after the attention and MLP residual sums of 4 encoder blocks & 6: after the outputs of 3 fusion MLPs and 3 DiT \texttt{mlp2} blocks \\ 
    Down, up shapes & $128\to16$, $16\to128$ & $192\to16$, $16\to192$ \\ 
    Adapter parameters & 32,768 (1.62\%) & 36,864 (0.61\%) \\ 
    BatchNorm & frozen statistics (adapter regimes); training mode (\FullFT) & none \\ 
    Objective & native planning regression and classification plus neighbor prediction & official ego and neighbor diffusion loss \\
    \midrule
    Adapter & \multicolumn{2}{>{\raggedright\arraybackslash}p{2.37in}@{}}{$x+\mathrm{up}(\mathrm{GELU}(\mathrm{down}(x)))$ in series after the existing output; no biases; up-projection zero-initialized; backbone frozen} \\
    \Replay & \multicolumn{2}{>{\raggedright\arraybackslash}p{2.37in}@{}}{512 stored samples per completed group; about $b/4$ replay samples per current micro-batch of $b$; loss weight 0.25; from stage~2} \\ 
    \PEGP & \multicolumn{2}{>{\raggedright\arraybackslash}p{2.37in}@{}}{inputs of the down-projection and post-GELU inputs of the up-projection; centered float64 covariance over valid tokens of at most 1,024 current-group samples; energy threshold 0.9; cumulative rank cap 16 per layer and side; gradients projected before AdamW from stage~2; no \Czero basis} \\ 
    \bottomrule
  \end{tabular}
\end{table}

\begin{table*}[t]
  \centering
  \caption{\textbf{\DiffP adapter ablations} (forward order, main recipe; cached ADE in m).
    Design: adapter width, per-layer rank caps of the down/up bases, basis rule (cand.: candidate-first, the main rule; res.: residual-first) and training seed (s1: 20261014, s2: 20261020). 
    Final: 7-group ADE at \Tk{7} (\Czero: 2.978); Forg.: forgetting \Fgt over the six old groups; Acq.: mean acquisition ($Q=-$ADE); Prot.: up-protected fraction (rank of the up basis in use divided by width, averaged over layers and stages 2--7). 
    --: no separate ablation run. Caps and basis rule only shape the projection basis, so \AdaFT was ablated over width only (seed s1), and up-only projection, which leaves the down cap unused, was not run with up cap 32. 
    \textsuperscript{a}Main run. \textsuperscript{b}Second-seed run of the main results.}
  \label{tab:F2}
  \scriptsize
  \setlength{\tabcolsep}{3.4pt}
  \begin{tabular}{@{}rccc rrr rrrr rrrr@{}}
    \toprule
    \multicolumn{4}{c}{Design} & \multicolumn{3}{c}{No projection (\AdaFT)} & \multicolumn{4}{c}{\PEGP, up-only} & \multicolumn{4}{c}{\PEGP, two-sided} \\
    \cmidrule(r){1-4}\cmidrule(lr){5-7}\cmidrule(lr){8-11}\cmidrule(l){12-15}
    Width & Caps & Basis & Seed & Final & Forg. & Acq. & Final & Forg. & Acq. & Prot. & Final & Forg. & Acq. & Prot. \\
    \midrule
    16 & 16/16 & cand. & s1 & 11.165\textsuperscript{a} & 8.758 & $-$0.301 & 7.256 & 4.258 & $-$0.217 & 0.90 & 3.907\textsuperscript{a} & 0.402 & $-$0.138 & 0.89 \\ 
    16 & 16/16 & cand. & s2 & -- & -- & -- & 8.421 & 5.544 & $-$0.244 & 0.90 & 4.014\textsuperscript{b} & 0.477 & $-$0.123 & 0.89 \\ 
    16 & 16/16 & res. & s1 & -- & -- & -- & 10.252 & 7.765 & $-$0.293 & 0.72 & 4.450 & 0.908 & $-$0.192 & 0.61 \\ 
    16 & 32/16 & cand. & s1 & -- & -- & -- & -- & -- & -- & -- & 3.767 & 0.261 & $-$0.105 & 0.89 \\ 
    16 & 64/16 & cand. & s1 & -- & -- & -- & -- & -- & -- & -- & 3.861 & 0.287 & $-$0.072 & 0.89 \\ 
    \midrule
    32 & 16/16 & cand. & s1 & 11.864 & 9.496 & $-$0.343 & 10.599 & 8.012 & $-$0.297 & 0.48 & 5.274 & 1.776 & $-$0.221 & 0.48 \\ 
    32 & 16/16 & cand. & s2 & -- & -- & -- & 11.079 & 8.534 & $-$0.353 & 0.48 & 5.534 & 2.049 & $-$0.185 & 0.48 \\ 
    32 & 16/16 & res. & s1 & -- & -- & -- & 10.009 & 7.327 & $-$0.315 & 0.48 & 5.276 & 1.849 & $-$0.193 & 0.47 \\ 
    \midrule
    64 & 16/16 & cand. & s1 & 11.185 & 8.721 & $-$0.316 & 11.789 & 9.407 & $-$0.319 & 0.25 & 5.747 & 2.431 & $-$0.122 & 0.25 \\ 
    64 & 16/16 & cand. & s2 & -- & -- & -- & 11.662 & 9.267 & $-$0.338 & 0.25 & 5.634 & 2.397 & $-$0.100 & 0.25 \\ 
    64 & 64/16 & cand. & s1 & -- & -- & -- & -- & -- & -- & -- & 5.960 & 2.677 & $-$0.135 & 0.25 \\ 
    64 & 64/32 & cand. & s1 & -- & -- & -- & -- & -- & -- & -- & 5.412 & 2.038 & $-$0.113 & 0.47 \\ 
    \bottomrule
  \end{tabular}
\end{table*}

\begin{figure*}[t]
  \centering
  \includegraphics[width=\textwidth]{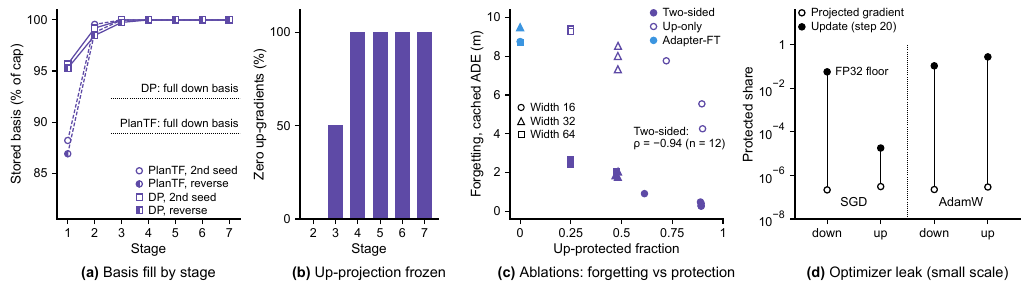}
  \caption{\textbf{Where \PEGP's protection acts, and how it leaks} (descriptive).
    (a)~Stored projection basis after each stage as a share of its fp32 maximum (rank 16 on both sides), second-seed and reverse-order runs (DP: \DiffP); dotted: a full down basis alone. 
    (b)~\DiffP, traced main-run configuration: share of up-projection observations whose projected gradient is exactly zero.
    (c)~The 23 ablation outcomes of \cref{tab:F2}: forgetting against the up-protected fraction. 
    (d)~Small-scale diagnostic (earlier code version, two-sided projection, seed means): protected share of the projected gradient and of the actual update (SGD, down: an update of norm about $10^{-6}$, whose protected share is FP32 rounding); Adam equals AdamW without weight decay.} 
  \label{fig:F2}
\end{figure*}

\subsection{Implementation}
\label{app:regimes-impl}

Each adapter sits in series after an existing output (\cref{tab:F1}), not in parallel with an MLP.
Every regime keeps each planner's native objective and native training-time dropout and adds no input augmentation. 
Every sequential run uses AdamW with peak learning rate $10^{-4}$, weight decay 0.01, gradient clipping at norm 5 and FP32, with a linear warm-up over each group's first epoch and a per-epoch cosine decay toward $10^{-6}$; optimizer and schedule restart at every group (\cref{sec:budget}).
The replay memory is a uniform random subset of each completed group's training samples.
For a current micro-batch of actual size $b$, \Replay draws $\max(1,\lfloor b/4\rfloor)$ replay samples with replacement from the memory of the completed groups and adds their separately averaged loss with weight 0.25 (\cref{tab:F1}); since replay starts at stage~2, the number of replay visits depends on the size of the first group (\cref{app:compute}). 
\PEGP merges the leading covariance directions of each new group with the stored basis by orthogonalization, stored directions first, up to the cumulative cap (not 16 directions per group); for \DiffP the activations come from the official inference-sampling path. 
It projects the gradient of each down- and up-projection weight onto the orthogonal complement of its stored input basis, so it protects an input-direction subspace, not a set of stored gradients, and a projected gradient vanishes exactly only when the basis spans the whole input space: possible on the 16-dimensional up side, never on the 128- or 192-dimensional down side. 

\subsection{Saturation and the optimizer leak}
\label{app:regimes-saturation}

\paragraph{Saturation.}
In a traced re-run of the main-run \DiffP \PEGP configuration (identical stage matrices), the down basis is full after stage~1 (\cref{sec:regimes}) and covers, averaged over the six layers, 65--68\% of the down-input energy of each later group. 
The up basis holds 5--9 directions per layer after stage~1, 11--16 after stage~2 and 16 from stage~3 on, so the projected up-gradient is exactly zero in 84 of 168 observations at stage~3 and in all from stage~4 on (\cref{fig:F2}b); the second-seed and reverse-order runs fill their bases in the same way, up to the cap after stage~3 and stage~4 (\cref{fig:F2}a). 
\PlanTF's bases also fill both sides by stage~3, although its down basis still grows after stage~1. 
Once both sides are capped, the projected gradient moves only the down-projections, within the complements of their input bases: 14,336 of \PlanTF's 32,768 adapter parameters (43.8\%) and 16,896 of \DiffP's 36,864 (45.8\%). 
Saturation does not lower \PEGP's acquisition below that of \AdaFT: cached acquisition at learning positions 4--7 (means over the three runs) is $-0.081$ \vs $-0.269$\,m for \DiffP and $+0.194$ \vs $+0.139$\,m for \PlanTF. 

\paragraph{Optimizer leak.}
Because projection acts on the gradient that enters AdamW, not on AdamW's update (\cref{sec:regimes}), 10.4\% of the down-projection update in the traced run lies in the protected subspace (mean over 2,172 layer--step observations), although the protected share of the projected gradient is numerically zero ($2.9\times10^{-7}$). 
A small-scale diagnostic on an earlier code version (12 DiT adapters; 20 steps on four repeated \gPickup samples; fixed basis; three seeds) traces the leak to AdamW's per-coordinate scaling. 
With two-sided projection, the protected share of the projected gradient stays below $5\times10^{-7}$ for SGD, Adam and AdamW, and SGD keeps the update outside the protected subspace up to FP32 rounding, whereas under Adam and AdamW 0.27 of the up- and 0.11 of the down-projection update lie inside it at step~20 (\cref{fig:F2}d); weight decay 0.01 changes this share by at most $1.9\times10^{-5}$. 
Momentum cannot cause the leak: with a basis fixed within a stage and optimizer states reset at every stage, Adam's first moment stays in the allowed subspace, whereas the division by the per-coordinate second-moment estimate does not commute with the projection.
Accordingly, two-sided projection reduces the drift of the earlier group's adapter outputs to 0.24 of the unprojected drift under SGD but only to 0.55 under Adam and AdamW; in our runs, the leak is the traced 10.4\%. 

\subsection{Ablations: where the protection comes from}
\label{app:regimes-ablations}

All ablations use \DiffP, the forward order, the main recipe and the cached-set evaluation, and vary the adapter width, the per-side rank caps, the basis rule and the training seed (\cref{tab:F2}). 
The 23 ablations comprise 3 \AdaFT runs without projection and 20 \PEGP variants, 12 with two-sided and 8 with up-only projection (no down-only arm). 
Three of them are runs of the main results (\cref{tab:F2}, notes a and b); none of the others informed the main configuration. 
Two-sided projection forgets less than up-only projection in all 8 matched pairs (same width, caps, basis rule and seed), by 3.86 to 6.98\,m of cached ADE. 
At caps 16/16 with the candidate-first rule, wider adapters protect a smaller share of the up-input space (about 0.89, 0.48 and 0.25 at widths 16, 32 and 64) and forget more, with either projection side and in both seeds, whereas \AdaFT shows no width trend. 
Across the 23 outcomes, forgetting falls as the protected fraction rises (Spearman $\rho=-0.94$, 12 two-sided runs) and as own-group ADE losses, the negative of acquisition, shrink ($\rho=+0.84$ over all 23, $+0.79$ over the 20 projection variants): protection partly reflects smaller updates (\cref{fig:F2}c). 
The best final 7-group ADE, 3.767\,m, stays above the 2.978\,m of \Czero, and mean acquisition is negative in all 23 outcomes ($-0.353$ to $-0.072$\,m). 
The outcomes separate only at the \gStationary stage (7-group ADE 3.43--4.24\,m up to stage~6, 3.77--11.86\,m at \Tk{7}). 

\paragraph{A ``no update after stage $k$'' reference.}
Stopping a run after stage~$k$, the extreme form of update suppression, scores as row~$k$ of its stage matrix. 
In the forward order, the final \DiffP \Replay and \PEGP checkpoints of both seeds differ by $-0.11$ to $+0.13$\,m in cached 7-group ADE from stopping \AdaFT after stage~6, and in the main run's closed loop neither differs detectably from that reference (\CLS-\NR, their \Tk{7} minus \AdaFT's \Tk{6}: $+0.043$ \ci{-0.008}{+0.101} and $+0.046$ \ci{-0.020}{+0.115}). 
As no regime improves its own \gStationary ADE by more than 0.02\,m in its stage-7 update, this reference cannot separate selective protection from suppression for \DiffP. 
For \PlanTF, \PEGP does not behave like this reference: its final checkpoint is 1.08 and 0.88\,m worse in cached ADE than its own \Tk{5} in the two forward-order seeds. 
}{}
\IfFileExists{sec/app_g_diffusion.tex}{
\section{Diffusion Planner Recipe Diagnostics}
\label{app:diffusion}
\providecommand*{\theHtable}{}\providecommand*{\theHfigure}{}%
\setcounter{table}{0}\setcounter{figure}{0}
\renewcommand{\thetable}{\thesection\arabic{table}}\renewcommand{\thefigure}{\thesection\arabic{figure}}
\renewcommand{\theHtable}{\thesection.\arabic{table}}\renewcommand{\theHfigure}{\thesection.\arabic{figure}}

This section reports a recipe diagnostic of \DiffP that complements the discussion of training without EMA in \cref{sec:why}.
It uses width-32 adapters (73,728 trainable parameters; width 16 in the main runs) trained only on the first group, \gStraight, with one training seed: the \Tk{1} of the width-32 ablation runs with seed s1 (\cref{tab:F2}), not a main-run checkpoint. 
As in the main runs (\cref{sec:deployment}), updates train and evaluate raw weights without EMA, and each evaluation samples one trajectory with the official 10-step DPM solver unless stated otherwise.
The diagnostics use the 640 \gStraight development samples (validation) and a 640-sample training subset, each with three sampling seeds (paired repeats, not training seeds). 

\begin{table}[t]
  \centering
  \caption{\textbf{Width-32 \Tk{1} diagnostic of \DiffP} (\gStraight only, one training seed). 
    Changes from \Czero after 20 epochs at each peak learning rate (LR). Loss: denoising (ego) loss, \%; ADE: sampled ADE, m (mean of three sampling seeds, 10-step DPM); Steps: share of the 10-step ADE gap to \Czero closed by 20/50 DPM steps, \%; --: not applicable (\Czero is the reference).} 
  \label{tab:G1}
  \scriptsize
  \setlength{\tabcolsep}{3.5pt}
  \begin{tabular}{@{}lrrrrrr@{}}
    \toprule
    & \multicolumn{3}{c}{Validation (640)} & \multicolumn{3}{c}{Training subset (640)} \\ 
    \cmidrule(lr){2-4}\cmidrule(l){5-7}
    Peak LR & Loss & ADE & Steps & Loss & ADE & Steps \\
    \midrule
    \Czero (level) & 0.0403 & 3.609 & -- & 0.0235 & 2.898 & -- \\ 
    $10^{-4}$ & $-$16.8 & $+$0.79 & 3.0/4.0 & $-$22.3 & $+$0.92 & 2.6/3.5 \\ 
    $3\times10^{-5}$ & $-$14.7 & $+$0.85 & 3.5/4.7 & $-$15.1 & $+$1.02 & 2.2/3.1 \\ 
    $10^{-5}$ & $-$12.8 & $+$0.97 & 2.8/3.7 & $-$11.4 & $+$1.06 & 1.4/2.0 \\ 
    \bottomrule
  \end{tabular}
\end{table}

\begin{figure*}[t]
  \centering
  \includegraphics[width=\textwidth]{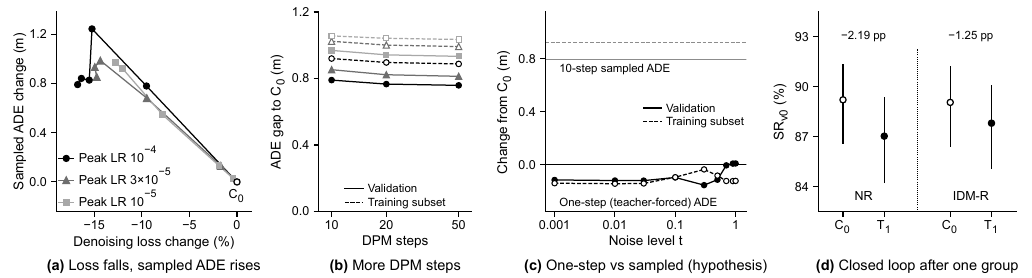}
  \caption{\textbf{\DiffP recipe diagnostics: width-32 adapter, \Tk{1} only, one training seed} (descriptive). 
    (a)~Change from \Czero of the validation denoising loss (\%) against that of the sampled ADE (m) after epochs 1, 2, 5, 10 and 20 at three peak learning rates. 
    (b)~ADE gap to \Czero at the same number of DPM steps for the epoch-20 checkpoints; solid: validation, dashed: training subset. 
    (c)~Change from \Czero to the epoch-20 checkpoint at peak $10^{-4}$ of the one-step (teacher-forced) ADE at each noise level $t$ (black) and of the 10-step sampled ADE (grey); a hypothesis, not a decomposition. 
    (d)~\SR on the 640 \gStraight scenarios at \Czero and \Tk{1} under \NR and \IDMR, with Wilson 95\% intervals. 
    Sampled values average three sampling seeds.}
  \label{fig:G1}
\end{figure*}

\paragraph{Learning rate.}
After 20 epochs at each of the three peak learning rates, the validation denoising loss falls by 13--17\% while the sampled ADE rises by 0.79--0.97\,m; the training subset shows the same, so the rise is not a generalization gap (\cref{tab:G1}, \cref{fig:G1}a). 
All 90 seed-paired comparisons (3 learning rates $\times$ epochs 1, 2, 5, 10 and 20 $\times$ 3 sampling seeds $\times$ 2 splits) are worse than \Czero, by at least 0.025\,m; a lower learning rate delays the rise (after epoch~1: $+0.78$, $+0.13$ and $+0.03$\,m) but does not prevent it. 

\paragraph{Sampler steps.}
Moving from 10 to 20 or 50 DPM steps closes only 1.4--4.7\% of the 10-step ADE gap to \Czero over the three learning rates and both splits (\cref{fig:G1}b), while \Czero's own ADE changes by at most 0.015\,m. 

\paragraph{One-step versus sampled error (a hypothesis).}
From \Czero to the epoch-20 checkpoint at $10^{-4}$, the one-step (teacher-forced) denoising loss falls at all 9 noise levels on both splits, and the one-step ADE improves at 7 of 9 levels on the validation set and at all 9 on the training subset (\cref{fig:G1}c). 
At the noisiest level ($t=0.999$), the one-step ADE changes by only $+0.011$\,m (validation) and $-0.120$\,m (training subset), so the excess of the sampled over the one-step ADE carries 98.6\% of the validation rise. 
We read this as damage in the sampling chain or in the spread of samples rather than in the denoiser's mean prediction; sample dispersion was not measured.

\paragraph{Closed loop after one group.}
In closed loop on 640 \gStraight scenarios, \SR under \NR goes from 571/640 (89.2\%, Wilson 95\% interval \ci{86.6}{91.4}) at \Czero to 557/640 (87.0\%, \ci{84.2}{89.4}) at this checkpoint, $-2.19$ percentage points, and under \IDMR from 570 to 562 ($-1.25$ points), while \ADEs rises by 0.83\,m under \NR (\cref{fig:G1}d). 
A large open-loop change thus maps to a small \SR change on the group just learned.

\paragraph{Scope.}
These diagnostics cover \Tk{1} only, whereas \DiffP's final results are driven by its \gStationary update (\cref{sec:why}).
Two observations are compatible with a recipe effect: the shared adapter \Tk{1} already raises the main run's 7-group cached ADE from 2.978 to 3.554\,m and carries 55--69\% of the final sentinel loss of \Replay and \PEGP (\cref{sec:reference}), and both coincide with the switch from EMA to raw weights. 
We did not test whether maintaining EMA during the updates removes them.
}{}
\IfFileExists{sec/app_h_compute.tex}{
\section{Training Cost}
\label{app:compute}

\makeatletter
\edef\cdBHsavedtable{\the\c@table}%
\let\cdBHsavedthetable\thetable
\providecommand*{\theHtable}{\arabic{table}}%
\let\cdBHsavedtheHtable\theHtable
\makeatother
\setcounter{table}{0}
\renewcommand{\thetable}{\thesection\arabic{table}}
\renewcommand{\theHtable}{\thesection.\arabic{table}}

\Cref{tab:H1} lists the budget of every run family: the sequential runs on the full training pools are matched in optimizer updates (\cref{sec:budget}), except one LoRA run with a global batch of 2,048, and the joint references are matched to them in sample visits; equal updates do not mean equal compute.
Replay starts at stage~2, so its overhead depends on the size of the first group: on the full pools, \Replay and LoRA+Replay add 19.0\% to the current-sample visits in the main and reverse orders, which open with a 100,000-sample group, and 24.3\% in the third order, which opens with the 11,282-sample \gHighLat group; in the equal-data runs, \Replay adds 21.4\%. 
Across groups, \Replay carries 512 stored samples per completed group (3,584 at the end), and \PEGP carries bases of rank at most 16 per layer and side (at most 73,728\,B for \PlanTF and 79,872\,B for \DiffP in FP32), whose collection costs compute that the visit counts omit; LoRA+PEGP caps its bases at rank 16 per factor (\cref{app:lora}).
LoRA trains more parameters than the adapters (\cref{tab:H2}), so the two branches are not matched in parameter count.
The wall-clock times of the core suite, 122.3\,h for its 24 runs, are descriptive only: the runs shared one server, and the times include evaluation, basis collection and checkpoint I/O, so they support no claim about efficiency. 

\begin{table}[!htb]
  \centering
  \caption{\textbf{Training budget per run}, by run family.
    Visits: current-sample visits; Replay: replay visits added by each \Replay or LoRA+Replay run (other regimes add none; --: the family has no such run). Neither counts \PEGP's basis collection. 
    \textsuperscript{a}Main run, second seed and reverse order.
    \textsuperscript{b}One run, \PlanTF LoRA+Replay in the reverse order, uses a global batch of 2,048: 4,100 updates and 1,569,640 replay visits.}
  \label{tab:H1}
  \scriptsize
  \setlength{\tabcolsep}{3pt}
  \begin{tabular}{@{}lrrrr@{}}
    \toprule
    Run family & Runs & Updates & Visits & Replay \\
    \midrule
    Core suite\textsuperscript{a}        & 24 & 32,420 & 8,278,780 & 1,569,660 \\ 
    Third order                          &  8 & 32,420 & 8,278,780 & 2,013,260 \\ 
    LoRA branch, main and reverse orders & 12 & 32,420\textsuperscript{b} & 8,278,780 & 1,569,660\textsuperscript{b} \\ 
    LoRA branch, third order             &  6 & 32,420 & 8,278,780 & 2,013,260 \\ 
    Joint references                     &  4 & 32,340 & 8,278,780 & -- \\ 
    Equal-data runs                      &  8 & 5,600 & 1,400,000 & 300,000 \\ 
    \bottomrule
  \end{tabular}
\end{table}

\begin{table}[!htb]
  \centering
  \caption{\textbf{Trainable parameters and wall-clock time per run.}
    LoRA branch: LoRA-FT, LoRA+Replay and LoRA+PEGP under \AdaFT, \Replay and \PEGP; --: no \FullFT counterpart.
    Wall-clock hours (core suite only): mean over the three stream settings, with the ratio to \AdaFT in parentheses (descriptive).}
  \label{tab:H2}
  \scriptsize
  \setlength{\tabcolsep}{3pt}
  \begin{tabular}{@{}l rrrr@{}}
    \toprule
    & \FullFT & \AdaFT & \Replay & \PEGP \\
    \midrule
    \multicolumn{5}{@{}l}{\emph{Trainable parameters}} \\
    \quad\PlanTF & 2,023,073 & 32,768 & 32,768 & 32,768 \\ %
    \quad\DiffP & 6,042,628 & 36,864 & 36,864 & 36,864 \\ %
    \multicolumn{5}{@{}l}{\emph{Trainable parameters, LoRA branch}} \\
    \quad\PlanTF & -- & 69,632 & 69,632 & 69,632 \\ 
    \quad\DiffP & -- & 110,592 & 110,592 & 110,592 \\ 
    \multicolumn{5}{@{}l}{\emph{Wall-clock hours per run, core suite}} \\
    \quad\PlanTF & 8.52 (2.07) & 4.11 (1.00) & 6.03 (1.47) & 4.12 (1.00) \\ 
    \quad\DiffP & 5.60 (1.61) & 3.48 (1.00) & 5.38 (1.54) & 3.51 (1.01) \\ 
    \bottomrule
  \end{tabular}
\end{table}

\setcounter{table}{\cdBHsavedtable}
\let\thetable\cdBHsavedthetable
\let\theHtable\cdBHsavedtheHtable
}{}
\IfFileExists{sec/app_i_related.tex}{
\section{Extended Related Work}
\label{app:related}
\makeatletter
\edef\cdIsavedtable{\the\c@table}%
\let\cdIsavedthetable\thetable
\providecommand*{\theHtable}{\arabic{table}}%
\let\cdIsavedtheHtable\theHtable
\makeatother
\setcounter{table}{0}%
\renewcommand{\thetable}{\thesection\arabic{table}}%
\renewcommand{\theHtable}{\thesection.\arabic{table}}%

\begin{figure*}[t]
  \captionof{table}{\textbf{Contemporaneous work} (first public on or after September 15, 2026), by date of arXiv v1.
    Relation: to \CD (``we'').}
  \label{tab:I1}
  \centering
  \scriptsize
  \setlength{\tabcolsep}{4pt}
  \setlength{\extrarowheight}{2pt}
  \begin{tabular}{@{}>{\raggedright\arraybackslash}p{0.18\textwidth}>{\raggedright\arraybackslash}p{\dimexpr0.51\textwidth-4\tabcolsep\relax}>{\raggedright\arraybackslash}p{0.31\textwidth}@{}}
    \toprule
    Work, v1 & Setting & Relation \\
    \midrule
    OPTED \cite{dacol2026opted} \newline Sep 17 
      & Closed-loop post-training of camera policies by a privileged RL teacher (3DGS scenes of real logs)
      & Post-trains in closed loop; we update offline on logs \\
    CounterPlay \cite{wei2026counterplay} \newline Sep 18 
      & Self-play driving policies post-trained on verified counterfactual retries (BehaviorBench)
      & Learns from its own retries; we from expert logs \\
    INTERACT \cite{distelzweig2026interact} \newline Sep 25 
      & Interactive planning: anchor-conditioned prediction, trust-region refinement (nuPlan, interPlan)
      & A planner, not an update protocol \\
    EvoAD \cite{yang2026evoad} \newline Sep 26 
      & Experience memory and latent world-model reflection; base policy not retrained (NAVSIM)
      & No weight updates; we update weights \\
    SAMBAR \cite{shrivastava2026sambar} \newline Sep 26 
      & Replay-free, constrained continual fine-tuning of vision-language-action models (LIBERO, hardware)
      & Acquisition \vs retention, in manipulation \\
    RefineDrive \cite{sun2026refinedrive} \newline Sep 28 
      & Vision-language-action driver post-trained on its own failures (retrieved corrections, GRPO; NAVSIM)
      & Learns from its own failures; we from expert logs \\
    RoXDrive \cite{lin2026roxdrive} \newline Sep 29 
      & Closed-loop RL post-training on action-faithful world-model rollouts (nuScenes, in-house data)
      & RL in a learned simulator; we score imitation updates \\
    Odyssey \cite{kim2026odyssey} \newline Oct 5 
      & Long-horizon closed-loop benchmark from nuPlan logs with SD-map routes
      & Scores fixed planners; we score updates \\
    Plans Are Not Executions \cite{zhao2026plansnotexec} \newline Oct 15 
      & Frozen world-model planners adapted to controller-induced execution shift (NAVSIM)
      & Plan scores can misjudge execution (cf.\ Axis~A) \\
    CL3DOD \cite{khan2026cl3dod} \newline Nov 2 
      & Class-incremental 3D object detection (KITTI, nuScenes)
      & Perception, not planning \\
    \bottomrule
  \end{tabular}
\end{figure*}

This section extends \cref{sec:related} with contemporaneous work (\cref{app:related-contemporaneous}) and with continual learning for trajectory prediction (\cref{app:related-prediction}).
None of the driving benchmarks and protocols compared with \CD in \cref{tab:positioning} (CLAD~\cite{verwimp2023clad}, $\infty$-Driver~\cite{lai2024infdriver} and DeLL~\cite{du2026dell}) scores updates against its start checkpoint, counts regressing cases, reports a worst-stage summary or repeats training with another seed, so none separates order effects from training variance.

\subsection{Contemporaneous work}
\label{app:related-contemporaneous}

\Cref{tab:I1} lists the contemporaneous works of the footnote in \cref{sec:related} (public on or after September 15, 2026) by the date of their first arXiv version, with their settings and their relation to \CD.
To our knowledge, none of these works evaluates a sequence of weight updates of a deployed driving planner against that planner.

\subsection{Continual learning for trajectory prediction}
\label{app:related-prediction}

Forecasting work replays generated samples~\cite{ma2021continual,bao2021lifelong}, adds classes over time~\cite{schischka2026cimf} or expands the network to avoid forgetting~\cite{li2026deitp}.
These works update predictors, which are scored against logged futures, whereas \CD updates planners whose plans are executed, so open- and closed-loop verdicts can differ (\cref{sec:loop}).
Of their mechanisms, \CD probes replay, with stored rather than generated samples (\Replay, \cref{sec:regimes}).

\setcounter{table}{\cdIsavedtable}%
\let\thetable\cdIsavedthetable
\let\theHtable\cdIsavedtheHtable
}{}

\fi

\end{document}